\documentclass[letterpaper]{article}
\usepackage{authblk}

\usepackage[footnotes,definitionLists,hashEnumerators,smartEllipses,hybrid]{markdown}
\usepackage{adjustbox}
\usepackage[super,comma,sort&compress]{natbib} \usepackage{alifeconf} 
\usepackage[table,xcdraw]{xcolor}
\usepackage{subcaption}
\usepackage{longtable}
\usepackage{listings}
\usepackage{amsmath}
\usepackage{hyperref,cleveref}
\usepackage[moderate]{savetrees}
\usepackage[parfill]{parskip}
\usepackage{rotating}
\usepackage{tabularx}
\usepackage{rotating}
\usepackage{pdflscape,lipsum}
\usepackage{eso-pic,zref-user}
\usepackage{fancyheadings}
\usepackage{tikz}
\usetikzlibrary{calc}
\usepackage{orcidlink}

\usepackage{floatpag}
\floatpagestyle{plain}
\fancypagestyle{mylandscape}{%
  \fancyhf{}
  \fancyfoot{
  \tikz[remember picture,overlay]
        \node[outer sep=0.5in,above,rotate=90] at (current page.east) {\thepage};}  
}

\usepackage{eso-pic,zref-user}
\newcounter{cntsideways}
\makeatletter
\AddToShipoutPictureBG{%
  \PLS@RemoveRotate
}
\AddToShipoutPictureBG{%
 \ifnum\zref@extractdefault{rotate\number\value{page}}{page}{0}=0
  \PLS@RemoveRotate
 \else
  \PLS@AddRotate{90}%
 \fi}

\newcommand\rotatesidewayslabel{\stepcounter{cntsideways}%
 \zlabel{tmp\thecntsideways}\zlabel{rotate\zref@extractdefault{tmp\thecntsideways}{page}{0}}}

\renewenvironment{sidewaysfigure}{%
  \begin{originalsidewaysfigure}%
  \thisfloatpagestyle{mylandscape}%
  \rotatesidewayslabel%
}{%
  \end{originalsidewaysfigure}%
}

\renewenvironment{sidewaystable}{%
  \begin{originalsidewaystable}%
  \thisfloatpagestyle{mylandscape}%
  \rotatesidewayslabel%
}{%
  \end{originalsidewaystable}%
}
\renewenvironment{sidewaystable*}{%
  \begin{originalsidewaystable}%
  \thisfloatpagestyle{mylandscape}%
  \rotatesidewayslabel%
}{%
  \end{originalsidewaystable}%
}

\renewenvironment{landscape}{%
  \begin{originallandscape}%
  \thispagestyle{mylandscape}%
  \rotatesidewayslabel%
}{%
  \end{originallandscape}%
}

\makeatother

\makeatletter
\let\pragma@iinput=\@iinput
\def\@iinput#1{\xdef\@pragmafile{#1}\pragma@iinput{#1}}
\def\@pragmafile{default}
\def\pragmaonce{%
   \csname pragma@\@pragmafile\endcsname
   \global\expandafter\let \csname pragma@\@pragmafile\endcsname = \endinput
}
\makeatother

\ifdefined\mydraft
\mydraft
\fi

\begin{document}


\title{Case Study of Novelty, Complexity, and Adaptation in a Multicellular System}
\author{
    Matthew Andres Moreno\orcidlink{0000-0003-4726-4479}$^{1,2,3,4,\dagger}$,\ %
    Santiago Rodriguez Papa\orcidlink{0000-0002-6028-2105}$^{5,7}$,\ %
    Charles Ofria\orcidlink{0000-0003-2924-1732}$^{5,6,7}$ \\
    \mbox{}\\
    $^1$Department of Ecology and Evolutionary Biology
    $^2$Center for the Study of Complex Systems \\
    $^3$Michigan Institute for Data and AI in Society
    $^4$University of Michigan, Ann Arbor, USA \\
    $^5$Department of Computer Science and Engineering
    $^6$Program in Ecology, Evolution, and Behavior \\
    $^7$Michigan State University, East Lansing, USA \\
    $^\dagger$\texttt{morenoma@umich.edu}
}

\maketitle

%
%


\begin{abstract}

Continuing generation of novelty, complexity, and adaptation are well-established as core aspects of open-ended evolution.
However, it has yet to be firmly established to what extent these phenomena are coupled and by what means they interact.
In this work, we track the co-evolution of novelty, complexity, and adaptation in a case study from the DISHTINY simulation system, which is designed to study the evolution of digital multicellularity.
In this case study, we describe ten qualitatively distinct multicellular morphologies, several of which exhibit asymmetrical growth and distinct life stages.
We contextualize the evolutionary history of these morphologies with measurements of complexity and adaptation.
Our case study suggests a loose --- sometimes divergent --- relationship can exist among novelty, complexity, and adaptation.
\end{abstract}


\providecommand{\dissertationexclude}[1]{%
\ifdefined\DISSERTATION
\else
#1
\fi
}

\section{Introduction}

The challenge and promise of open-ended evolution has animated decades of inquiry and discussion within the artificial life community \citep{packard2019overview}.
The difficulty of devising models that produce continuing open-ended evolution suggests profound philosophical or scientific blind spots in our understanding of the natural processes that gave rise to contemporary organisms and ecosystems.
Already, pursuit of open-ended evolution has yielded paradigm-shifting insights.
For example, novelty search demonstrated how processes promoting non-adaptive diversification can ultimately yield adaptive outcomes that were previously unattainable \citep{lehman2011abandoning}.
Such work lends insight into fundamental questions in evolutionary biology, such as the relevance --- or irrelevance --- of natural selection with respect to increases in complexity \citep{lehman2012evolution, Lynch8597} and the origins of evolvability \citep{lehman2013evolvability,Kirschner8420}.
Evolutionary algorithms devised in support of open-ended evolution models also promise to deliver tangible, broader impacts for society.
Possibilities include the generative design of engineering solutions, consumer products, art, video games, and AI systems \citep{nguyen2015,stanley2019open}.

The preceding decades have witnessed advances in defining, both quantitatively and philosophically, the concept of open-ended evolution \citep{lehman2012beyond,dolson2019modes,bedau1998classification}.
Notable investigations have also been made into causal phenomena that promote open-ended dynamics, such as ecological dynamics, selection, and evolvability \citep{dolson2019constructive,soros2014identifying,huizinga2018emergence}.
The concept of open-endedness is fundamentally characterized by intertwined generation of novelty, functional complexity, and adaptation \citep{taylor2016open}.
The strength and nature of relationships among these three phenomena, however, are yet to be fully understood.
Here, we aim to complement ongoing work to develop a firmer theoretical understanding of the relationship between novelty, complexity, and adaptation by exploring the evolution of these phenomena through a case study using the DISHTINY digital multicellularity framework \dissertationexclude{\citep{moreno2019toward}}.
We apply a suite of qualitative and quantitative measures to assess how these qualities can change over evolutionary time and in relation to one another.


\providecommand{\dissertationelse}[2]{%
\ifdefined\DISSERTATION
#1
\else
#2
\fi
}


\providecommand{\dissertationexclude}[1]{%
\ifdefined\DISSERTATION
\else
#1
\fi
}


\providecommand{\dissertationonly}[1]{%
\ifdefined\DISSERTATION%
#1%
\else%
\fi
}

\section{Methods} \label{sec:methods;ch:measuring-cna}

\subsection{Simulation}

\begin{figure*}

\includegraphics[width=\linewidth]{{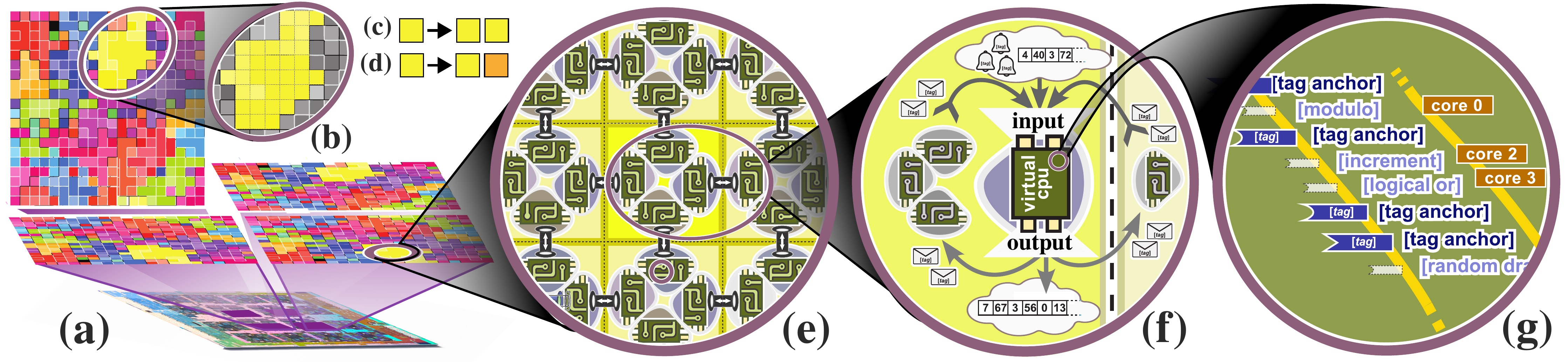}}

\caption{
\textbf{Overview of digital multicell model.}
\footnotesize
Population comprises a toroidal grid of cells occupied by replicating computer programs.
In the surveyed case study, grid size was 14,400 cells ($120\times120$).
To accelerate the feasible evolutionary depth of experiments, population subgrids are evaluated across available hardware threads and processes --- in this work, 4 threads (panel $a$).
At subgrid boundaries where inter-cell interactions reach between different threads or processes, communication is handled on a best-effort basis via the underlying Conduit library.
On the population grid, cells may form local multicell groups --- shown as colored patches in visualizations (panel $b$).
Cells replicate by copying program content into a chosen neighbor cell, and may choose to grow their existing multicell group ($c$) --- or splinter their offspring off to found a new group ($d$).
Within each cell, behavior is directed by a collection of four virtual CPUs (N, S, E, and W ``cardinal processors''), each independently managing the cell's interactions with a one neighbor cell (panel $c$).
Each cardinal processor can interact with other cardinal processors in its own cell via message passing.
A cardinal processor can also exchange messages with its counterpart cardinal processor in the abutting neighbor cell, allowing arbitrary inter-cell communication.
Alongside message-passing I/O, cardinal processors accept input to sense local simulation state and create output to perform cell actions (panel $f$).
Sensory input occurs via a combination of special input registers (e.g., for own- and neighbor-cell resource availability, cell age, etc.) and special messages triggered by simulation conditions (``events,'' depicted as bells in panel $f$).
Behavior outputs are dispatched via special output registers (e.g., for resource sharing, replication, apoptosis, etc.).
At implementation level, replicator program code is evaluated via an event-driven execution model.
That is, tag-matching mechanisms trigger program submodules in response to stimuli (panel $g$).
In addition to external stimuli (i.e., inter-cellular messages, intra-cellular messages, and simulation events), program code can also trigger arbitrary self-stimuli.
Virtual hardware supports concurrent interaction handling via independent pseudo-cores, as well as dynamic plasticity through regulation of tag-matching affinities within each cardinal processor.
}
\label{fig:overview}
\end{figure*}

The DISHTINY simulation environment tracks cells occupying tiles on a toroidal grid (size $120\times120$ by default).
Cells collect a uniform inflow of continuous-valued resource.
This resource can be spent in increments of $1.0$ to attempt asexual reproduction into any of a cell's four adjacent cells.
A cell can only be replaced if it commands less than $1.0$ resource.
If a cell rebuffs a reproduction attempt, its resource stockpile decrements by $1.0$ down to a minimum of $0.0$.

To facilitate the formation of coherent multicellular groups, the DISHTINY framework provides a mechanism for cells to form groups and detect group membership \dissertationexclude{\citep{moreno2019toward}}.
Groups arise through cellular reproduction.
When a cell proliferates, it may choose to initiate its offspring as a member of its kin group, thereby growing it, or induce the offspring to found a new kin group.
This process is similar to the growth of biological multicellular tissues, where cell offspring can be retained as members of the tissue or permanently expelled.

We incentivize group formation by providing an additional resource inflow bonus based on group size.
Per-cell resource collection rate increases linearly with group size up to a cap of 12 members.
Beyond 12 members, the decay rate of cells' resource stockpiles begins to increase exponentially.
These mechanisms select for medium-sized groups; in particular, the harsh penalization of oversized groups prevents any single group from consuming the entire population.
Undersized groups, on the other hand, do not receive resource bonuses.
Groups that are too large receive a penalty.
To ensure group turnover, we force groups to fragment into unicells after 8,192 ($2^{13}$) updates.


In \dissertationelse{Chapter \ref{ch:case-studies}}{previous work}, we established that this framework can select for traits characteristic of multicellularity, such as cooperation, coordination, and reproductive division of labor \dissertationexclude{\citep{moreno2021exploring}}.
We also found that more case studies of interest arose when two nested levels of group membership were tracked
as opposed to a single, unnested level of group membership \dissertationexclude{\citep{moreno2021exploring}}.
With nested group membership, group growth still occurs by cellular reproduction.
Cells are given the choice to retain offspring within both groups, to expel offspring from both groups, or to expel offspring from the innermost group only.
\dissertationonly{Section \ref{sec:hierarchical_nesting} provides greater detail on group membership and hierarchical group membership in DISHTINY.}
In this work, we allow for nested kin groups.

In addition to controlling reproduction behavior, evolving genomes can also share resources with adjacent cells, perform apoptosis (recovering a small amount of resource that may be shared with neighboring cells), and pass arbitrary messages to neighboring cells.
Cell behaviors are controlled by event-driven genetic programs in which linear GP modules are activated in response to cues from the environment or neighboring agents; signals are handled in quasi-parallel on up to 32 virtual cores (Figure \ref{fig:overview}g) \citep{lalejini2018evolving}.
Each cell contains four independent virtual CPUs, all of which execute the same genetic program (Figure \ref{fig:overview}e).
Each CPU manages interactions with a single neighboring cell.
We refer to a CPU managing interactions with a particular neighbor as a ``cardinal processor'' (in analogy to ``cardinal directions'').
These CPUs may communicate via intracellular message passing.
To support plasticity and differentiation, each CPU may also dynamically regulate tag-matching priorities of program modules \citep{lalejini2021tag}.

Full details on the instruction set and event library used, as well as simulation logic and parameter settings, appear in supplementary material.


\subsection{Evolution}
\label{sec:evolution;ch:measuring-cna}

\begin{figure}
\includegraphics[width=\linewidth]{{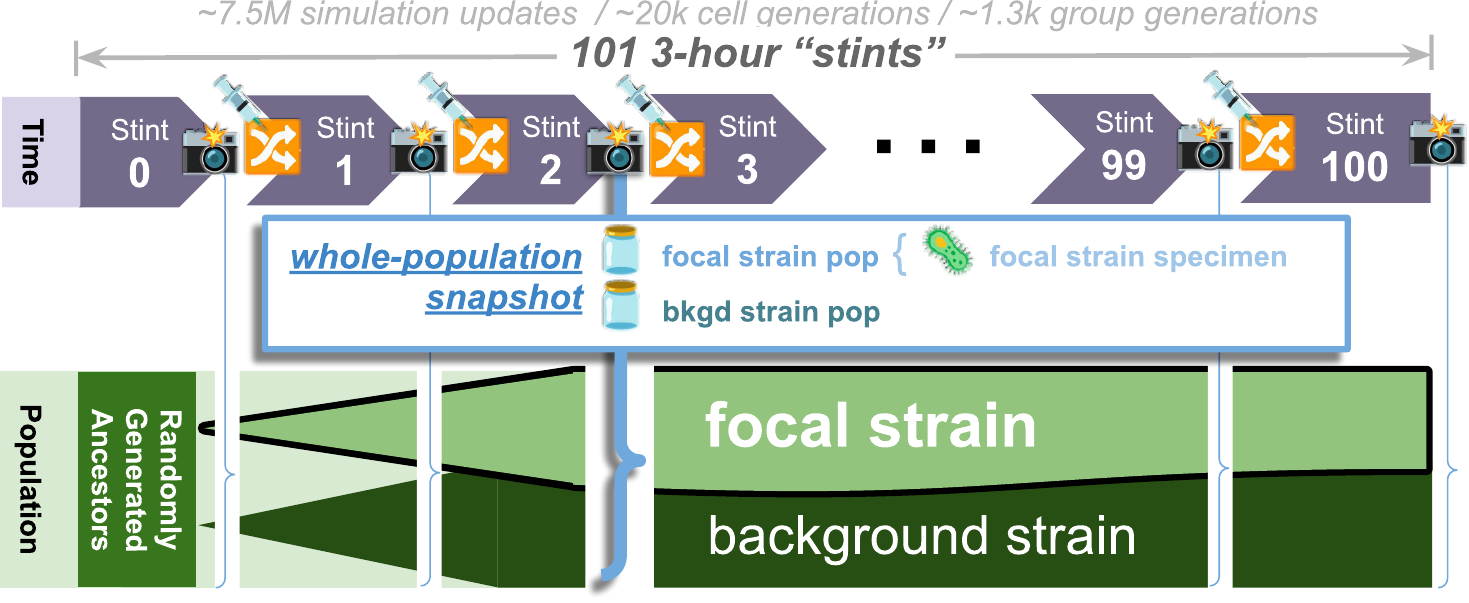}}
\caption{%
\textbf{Evolutionary regimen.}
\footnotesize
Experiment was initialized with a founder population of randomly generated genomes.
Among founder lineages, an arbitrary clade was designated as the ``focal strain'' subject of case study.
To ensure genomes maintained viability as simulation seeds (e.g., for competition and monoculture trials) and provide checkpoints in the case of cluster downtime, population was subjected to serial passage between 3-hour simulation windows --- referred to as ``stints.''
Between stints, a whole-population snapshot of genome content was recorded --- allowing later isolation of focal and background strain subpopulations, as well as a sample focal strain genome (``focal specimen'').
To continue evolution in the following stint, a fresh simulation instance was initialized by shuffled injection of snapshot genomes.
Note that, due to early extinctions, an initially-secondary strain gained focal designation at stint 2, which was maintained onwards.
A diversity-maintenance procedure was used to ensure stable long-term coexistence between at least two founder lineages, hence ensuring long-term preservation of at least one independent ``background'' strain.
}
\label{fig:stints}
\end{figure}

We performed evolution in three-hour windows for compatibility with our compute cluster's scheduling system.
We refer to these windows as ``stints.''
We randomly generated 100-instruction genomes at the outset of the initial stint, stint 0.
At the end of each three-hour window, the system harvested and stored genomes in a population file.
We then seeded subsequent stints with the previous stint's population.
No simulation state besides genome content was preserved between stints.
In addition to simplifying implementation concerns, re-seeding each stint ensured that strains retained the capability to grow from a well-mixed inoculum.
This facilitated later competition experiments between strains.
Figure \ref{fig:stints} summarizes this procedure.

In order to ensure heterogeneity of biotic environmental factors experienced by evolving cells, we imposed a diversity maintenance scheme.
In this scheme, descendants of a single progenitor cell from stint 0 that proliferated to constitute more than half of the population were penalized with resource loss.
The severity of the penalty increased with increasing prevalence beyond half of the population.
Thus, we ensured that descendants from at least two distinct stint 0 progenitors remained over the course of the simulation.
We arbitrarily chose a strain for primary study --- we refer to this strain as the ``focal'' strain and others as ``background'' strains.
In our case study, there was only one background strain in addition to this focal strain.

In our screen for case studies, we evolved 40 independent populations for 101 stints.
We selected population 16005 from among these 40 to profile as a case study due to its distinct asymmetrical group morphology.

At the conclusion of each stint, we selected the most abundant genome within the population as a representative specimen.
We performed a suite of follow-up analyses on each representative specimen to characterize aspects of complexity, detailed in the following subsections.
To ensure that specimens were consistently sampled from descendants of the same stint 0 progenitor, we only considered genomes with the lowest available stint 0 progenitor ID.

\subsection{Phenotype-neutral Nopout}
\label{sec:phenotype_neutral_nopout}

After harvesting representative specimens from each stint, we filtered out genome instructions that had no impact on the simulation.

To accomplish this, we performed sequential single-site ``nopouts'' where individual genome instructions were disabled by replacing them with a \texttt{Nop} instruction.%
\footnote{
This \texttt{Nop} instruction was chosen to perform the same number of random number generator touches as the original instruction to control for arbitrary effects of advancing the generator.
}
We reverted nopouts that altered a strain's phenotype and kept those that did not.
To determine whether phenotypic alteration occurred, we seeded an independent, mutation-disabled simulation with the strain in question and ran it side-by-side with an independent, mutation-disabled simulation of the wildtype strain.
If any divergence in resource concentration was detected between the two strains within a 2,048-update window, the single-site nopout was reverted.
We continued this process until no single-site nopouts were possible without altering the genome's phenotype.
To speed up evaluation, we performed step-by-step, side-by-side comparisons using a smaller toroidal grid size of just 100 tiles.

This process left us with a ``Phenotype-neutral Nopout'' variant of the wildtype genome where all remaining instructions contributed to the phenotype.

However, in further analyses we discovered that 21 phenotype-neutral nopouts from our case study were \textit{not} actually neutral --- competition experiments revealed they were significantly less fit than the wildtype strain.
This might be due to insufficient spatial or temporal scope to observe the expression of particular genome sites in our test for phenotypic divergence.

\subsection{Estimating Critical Fitness Complexity}

Next, we sought to detect genome instructions that contributed to strain fitness.

For each remaining op instruction in the Phenotype-neutral Nopout variant, we took the wildtype strain and applied a nopout at the corresponding site.
We then competed this variant against the wildtype strain.
Evaluating only remaining op instructions in the Phenotype-neutral Nopout variant allowed us to decrease the number of fitness competitions we had to perform.

We initialized fitness competitions by seeding a population half-and-half with two strains.
We ran these competitions for 10 minutes (about 4,200 updates) on a $60\times60$ toroidal grid, after which we assessed the relative abundances of descendants of both seeded strains.

To determine whether fitness differed significantly between a wildtype and variant strain, we compared the relative abundance of the strains observed at the end of competitions against outcomes from 20 control wildtype-vs-wildtype competitions.
We fit a $T$-distribution to the abundance outcomes observed under the control wildtype-vs-wildtype competitions and deemed outcomes that fell outside the central 98\% probability density of that distribution a significant difference in fitness.
This allowed us to screen for fitness effects of single-site nopouts while only performing a single competition per site.

This process left us with a ``Fitness-noncritical Nopout'' variant of the wildtype genome where all remaining instructions contributed to the phenotype.
We called the number of remaining instructions its ``critical fitness complexity.''
We adjusted this figure downward for the expected 1\% rate of false-positive fitness differences among tested genome sites.
This metric mirrors the MODES complexity metric described by \citet{dolson2019modes} and the approximation of sequence complexity advanced by \citet{adami2000evolution}.

\subsection{Estimating State Interface Complexity}
\label{sec:estimating-state-interface-complexity;ch:measuring-cna}

In addition to estimating the number of genome sites that contribute to fitness, we measured the number of different environmental cues and the number of different output mechanisms that cells adaptively incorporated into behavior.

One possible way to take this measure would be to disable event cues, sensor instructions, and output registers one by one and test for changes in fitness.
However, this approach would fail to distinguish context-dependent input/output from merely contingent input/output.
For example, a cell might arbitrarily depend on a sensor being set at a certain frequency, but not make use of underlying information the sensor provides.

To isolate context-dependent input/output state interactions, we tested the fitness effect of swapping particular input/output states between CPUs rather than completely disabling them.
That is, for example, CPU $b$ would be forced to perform the output generated by CPU $a$ or CPU $b$ would be shown the input meant for CPU $a$.
We performed this manipulation on half the population in a fitness competition for each individual component ofitnessf the simulation's introspective state (44 sensor states relating to the status of a CPU's own cell), extrospective state (61 sensor states relating to the status of a neighboring cell), and writable state (18 output states, 10 of which control cell behavior and 8 of which act as global memory for the CPU).%
\footnote{
A full description of each piece of introspective, extrospective, and writable state is listed in supplementary material.
}
We deemed a state as fitness-critical if this manipulation resulted in decreased fitness at significance $p < 0.01$ using a $T$-test parameterized by 20 control wildtype vs wildtype competitions.

We describe the number of states that cells interact with to contribute to fitness as ``State Interface Complexity.''

\subsection{Estimating Messaging Interface Complexity}
\label{sec:estimating-messaging-interface-complexity;ch:measuring-cna}

In addition to estimating the number of input/output states cells use to interact with the environment, we also estimated the number of distinct intracellular messages cardinal processors within a cell use to coordinate, as well as the intercellular messages that cells use to coordinate.
As with state interface complexity, distinguishing context-dependent behavior from contingent behavior is critical to attaining a meaningful measurement.
For example, a cardinal processor might depend on always receiving an intercellular message from a neighbor or an intracellular message from another cardinal processor.
Although meaningless, if that message were blocked, fitness would decrease.
So, instead of simply discarding messages to test for a fitness effect, we reroute messages back to the sending cardinal processor instead of their intended recipient.
We deemed messages as fitness-critical if this manipulation resulted in decreased fitness at significance $p < 0.01$ using a $T$-test parameterized by 20 control wildtype vs wildtype competitions.

We refer to the number of distinct messages that cells send to contribute to fitness as ``Messaging Interface Complexity.''

We refer to the sum of State Interface Complexity, Intra-messaging Interface Complexity, and Inter-messaging Interface Complexity as ``Cardinal Interface Complexity.''

\subsection{Estimating Adaptation}
\label{sec:measuring-adaptation;ch:measuring-cna}

\begin{figure}

\begin{minipage}[c]{0.2\linewidth}
~
\end{minipage}
~
\begin{minipage}[b]{0.8\linewidth}
        \includegraphics[width=\linewidth]{{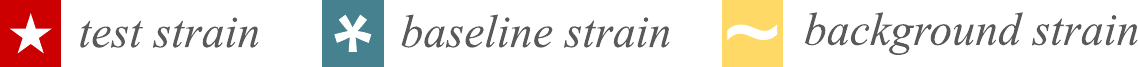}}
\end{minipage}

\vspace{1em}

\begin{subfigure}{\linewidth}
\begin{minipage}{0.2\linewidth}
\caption{\footnotesize no biotic background}
\label{fig:fit-nobbg}
\end{minipage}
~
\begin{minipage}{0.78\linewidth}
    \includegraphics[width=\linewidth]{{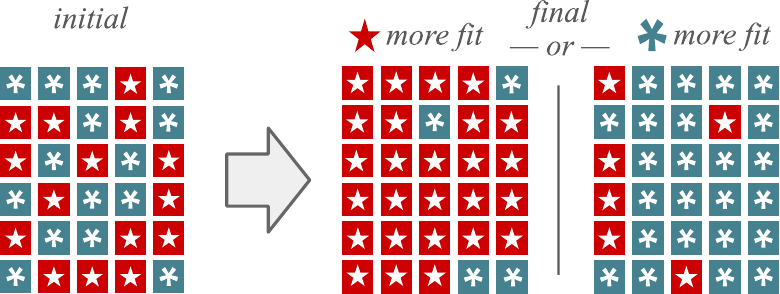}}
\end{minipage}
\end{subfigure}

\vspace{1em}

\begin{subfigure}{\linewidth}
\begin{minipage}{0.2\linewidth}
\caption{\footnotesize with biotic background}
\label{fig:fit-bbg}
\end{minipage}
~
\begin{minipage}{0.78\linewidth}
    \includegraphics[width=\linewidth]{{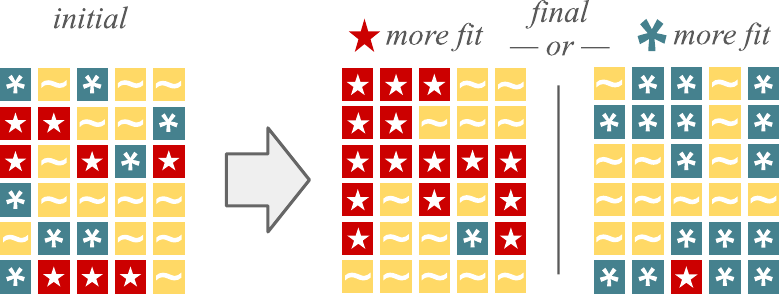}}
\end{minipage}
\end{subfigure}

\caption{%
\textbf{Adaptation assay design.}
\footnotesize
Due to the implicit nature of fitness in our study system, adaptation was primarily assessed on a relative basis, using competition experiments, rather than on an absolute basis.
Panel \ref{fig:fit-nobbg} overviews competition experiment design.
For these experiments, population cells are seeded with genomes from two strains in equal proportion,
After a fixed simulation duration, strain abundances are assessed.
In stint-to-stint adaptation assays, ``winning'' (higher abundance) strain consistency across replicates determined outcome significance.
For genotype/phenotype complexity screens --- which required assessment of numerous knockout variants --- significance was instead assessed by comparing end-state abundance ratio against a sampled distribution of outcomes from control WT vs. WT competition replicates.
In some cases, to assess context-sensitivity of fitness, competitions were performed under ``biotic background'' conditions (panel \ref{fig:fit-bbg}).
For these competition trials, half the initial population was seeded with a background strain.
Although the background strain participated fully during competition runtime, it was excluded from end-state abundance ratio calculations --- which considered only the test and baseline strains.
}
\label{fig:fit}

\end{figure}

In order to assess ongoing changes in fitness, we performed fitness competitions between the representative focal strain specimen sampled at each stint and the focal strain population from the preceding stint.%
\footnote{Recall from Section \ref{sec:evolution;ch:measuring-cna} that, due to a diversity maintenance procedure, two completely independent strains coexisted over the course of the experiment --- the ``focal'' strain selected for analysis and a ``background'' strain.}
Using the population from the preceding stint as the competitive baseline (rather than the representative specimen) ensured more focused, consistent measurement of the fitness properties of the specimen at the current stint (e.g., preventing skewed results from a sampled ``dud'' at the preceding stint).
Figure \ref{fig:fit} overviews fitness competition design.

We performed 20 independent replicates of each competition.
Competing strains were well-mixed within the full-sized toroidal grid at the outset of each competition, which lasted for 10 minutes of wall time.
This was sufficient to simulate about 8,000 updates at stint 0 and 2,000 updates at stint 100 (Supplementary Figures \ref{fig:num_updates_elapsed_heatmap}, \ref{fig:num_updates_elapsed_barplot}, and \ref{fig:num_updates_elapsed_boxplot}).
We determined that a gain of fitness had occurred if the current stint specimen constituted a population majority at the conclusion of more than 17 of those competitions, corresponding to a significance level of $p < 0.005$ under the two-tailed binomial null hypothesis.
Likewise, we deemed winning fewer than 3 competitions a significant fitness loss.

\subsection{Implementation}

We employed multithreading to speed up execution.
We split the simulation into four $60\times60$ subgrids.
Each subgrid executed asynchronously, using the Conduit C++ Library \dissertationonly{presented in Chapter \ref{ch:conduit}} to orchestrate best-effort, real-time interactions between simulation elements on different threads \dissertationexclude{\citep{moreno2021conduit}}.
This approach is in part informed by Ackley's principles of indefinite scalability, which highlight fault tolerance, soft synchronization, decentralization, and spatial locality as key engineering principles for large-scale \textit{in silico} artificial life systems \citep{ackley2014indefinitely,ackley2023robust}.
In particular, the localized grid topology and best-effort communication model of DISHTINY are designed with an eye towards compatibility with the dataflow architecture of next-generation AI/ML hardware accelerator devices, which field up to hundreds of thousands of independent processor cores but impose restrictions on communication locality and memory availability \citep{lauterbach2021path,moreno2024trackable}.
In other work benchmarking the system, we have demonstrated that this approach improves scalability.
The simulation scales to 4 threads with 80\% efficiency, up to 64 threads with 40\% efficiency and up to 64 nodes with 80\% efficiency \dissertationelse{(Chapter \ref{ch:conduit})}{\citep{moreno2021conduit}}.

Over the 101 three-hour evolutionary stints performed to evolve the case study, 7,565,309 simulation updates elapsed.
This translates to 74,904 updates elapsed per stint or about 6.9 updates per second.
However, the update processing rate was not uniform across stints: the simulation slowed about 77\% as stints progressed.
Supplementary Figure \ref{fig:simulation} shows elapsed updates for each stint.
During stint 0, 176,816 updates elapsed (about 16.3 updates per second).
During stint 100, only 41,920 updates elapsed (about 3.8 updates per second).

Although working asynchronously, threads processed similar number of updates during each stint.
The mean standard deviation of update-processing rate between threads was 2\%.
The mean difference of the update-processing rate between the fastest and slowest threads was 5\%.
The maximum value of these statistics observed during a stint was 9\% and 20\%, respectively, at stint 44.
Supplementary Figure \ref{fig:simulation:thread_updates} shows the distribution of elapsed updates across threads for each stint evolved during the case study.

This project benefited significantly from open-source scientific software and data standards \citep{2020SciPy-NMeth,harris2020array,reback2020pandas,mckinney-proc-scipy-2010,waskom2021seaborn,hunter2007matplotlib,moreno2023teeplot,seabold2010statsmodels,vostinar2024empirical,moreno2021signalgp,moreno2021conduit,cock2009biopython,lalejini2019data}.


\providecommand{\dissertationelse}[2]{%
\ifdefined\DISSERTATION
#1
\else
#2
\fi
}


\providecommand{\dissertationexclude}[1]{%
\ifdefined\DISSERTATION
\else
#1
\fi
}


\providecommand{\dissertationonly}[1]{%
\ifdefined\DISSERTATION%
#1%
\else%
\fi
}

\section{Results}

\subsection{Evolutionary History}

Due to the parallel, asynchronous nature of the experimental framework, we did not perform perfect phylogeny tracking \citep{moreno2024analysis}.
\dissertationonly{Chapter \ref{ch:distributed-phylogeny} discusses challenges parallelizing perfect phylogeny tracking in depth.}

Instead, we opted to track the total number of ancestors seeded into stint 0 with extant descendants.
At the end of stints 0 and 1, three distinct original phylogenetic roots were present in the population.
From stint 2 onward, only two distinct original phylogenetic roots were present.

We performed follow-up analyses on specimens sampled from the lowest original phylogenetic root ID present in the population.%
\footnote{
This approach was designed to choose an arbitrary strain as focal.
Barring extinction, that same strain will then be identified as focal consistently across subsequent stints.
Phylogenetic root ID had no functional consequences; it is simply an arbitrary basis for focal strain selection.
}
For the first two stints, the focal strain was root ID 2,378.
During stint 2, original phylogenetic root 2,378 went extinct.
So, all further follow-up analyses were sampled from descendants of ancestor 12,634.

We also tracked the number of genomes reconstituted at the outset of each stint that had extant descendants at the end of that stint.
This count grows from approximately 10 around stint 15 to upwards of 30 around stint 40 (Supplementary Figure \ref{fig:phylogeny:stint_roots}).
Among descendants of the lowest original phylogenetic root, the number of independent lineages spanning a stint also increases from around 5 to around 15
(Supplementary Figure \ref{fig:phylogeny:lowestroot_stint_roots}).
This decrease in phylogenetic consolidation on a stint-by-stint basis correlates with the waning number of simulation updates performed per stint (Supplementary Figures \ref{fig:phylogeny:updates_vs_stint_roots} and \ref{fig:phylogeny:log_updates_vs_stint_roots}).
More complete phylogenetic data will be necessary in future experiments to address questions about the possibility of long-term stable coexistence beyond the two strains supported under the explicit diversity maintenance scheme.

On the specimen from stint 100 used in the final case study, an evolutionary history of 20,212 cell generations had elapsed.
Of these cellular reproductions, 11,713 (58\%) had full kin group commonality, 7,174 had partial kin group commonality (35\%), and 1,325 had no kin group commonality (7\%).
On this specimen, 1,672 mutation events had elapsed.
During these events, 7,240 insertion-deletion alterations had occurred and 26,153 point mutations had occurred.
This strain experienced a selection pressure of 18\% over its evolutionary history, meaning that only 82\% of the mutations that would be expected given the number of cellular reproductions that had elapsed were present.

\begin{figure*}
\centering
\includegraphics[width=0.8\linewidth]{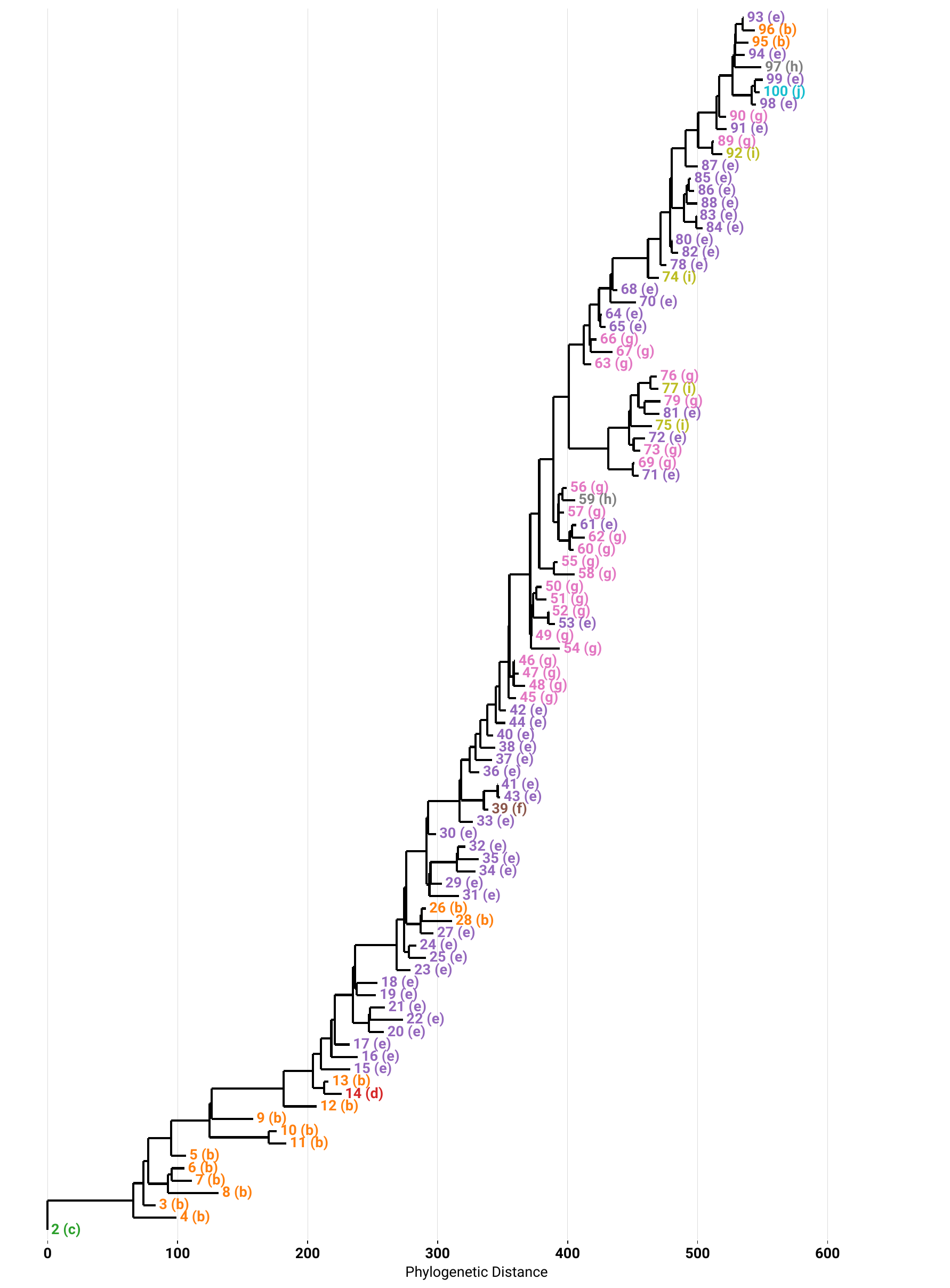}

\caption{
\textbf{Estimated phylogeny of sampled focal strain representatives.}
\footnotesize
Phylogeny constructed using parsimony algorithm based on genome tag bitstring content \citep{cock2009biopython}.
Each leaf node corresponds to a sampled representative.
Representatives from stints 0 and 1, which share no common ancestry with representatives from other stints, are excluded.
Numbers refer to stint that each representative was sampled from.
Color coding and parentheticals of stint labels correspond to qualitative morph codes described in Table \ref{tab:morph_descriptions}.
}
\label{fig:phylo_parsimony_tree}
\end{figure*}

In order to characterize the evolutionary history of the experiment in greater detail, we performed a parsimony-based phylogenetic reconstruction on the sampled representative specimens from each stint, shown in Figure \ref{fig:phylo_parsimony_tree}.
We used genomes' fixed-length blocks of 35 64-bit tags that mediate environmental interactions as the basis for this reconstruction.
These tag blocks underwent bitwise mutation over the course of the experiment.%
\footnote{
In future experiments, we plan to incorporate new methodology for ``hereditary stratigraph'' genome annotations expressly designed to facilitate phylogenetic reconstruction \dissertationelse{(Chapter \ref{ch:distributed-phylogeny})}{\citep{moreno2022hereditary}}.
}
Supplementary Figure \ref{fig:phylo_distance_matrix_heatmap} shows Hamming distance between all pairs of tag blocks.
We additionally tried several other tree inference methods, discussed in supplementary material; however, these yielded lower-quality reconstructions.

Although the phylogeny of stint representatives includes many instances that do not constitute a strict sequential lineage (i.e., each stint's representative descending directly from the preceding stint's representative), we did not observe evidence of long-term coexistence of clades over more than ten stints.

\subsection{Qualitative Morphological Categorizations}


\providecommand{\dissertationelse}[2]{%
\ifdefined\DISSERTATION
#1
\else
#2
\fi
}


\providecommand{\dissertationexclude}[1]{%
\ifdefined\DISSERTATION
\else
#1
\fi
}


\providecommand{\dissertationonly}[1]{%
\ifdefined\DISSERTATION%
#1%
\else%
\fi
}

\newcommand{\includesnapshot}[1] {%
\adjustbox{trim={0.05\width} {0.35\width} {0.05\width} {0.35\width},clip}%
    {\includegraphics[height=\dissertationexclude{0.3}\dissertationonly{0.25}\textheight]{#1}}
}
\newcommand{\morphtext}[1] {%
\color[HTML]{FFFFFF} \huge \raisebox{1.2em}{\textbf{#1}}%
}
\newcommand{\videolink}[1] {%
\raisebox{
\dissertationexclude{2.8em}
\dissertationonly{2.2em}
}{\begin{minipage}{\dissertationelse{2.4cm}{1.3 cm}} \dissertationelse{\fontsize{6}{7}\selectfont}{\tiny}\url{#1} \end{minipage}}
}
\newcommand{\descript}[1] {%
\raisebox{
  \dissertationexclude{2.8em}
  \dissertationonly{2.2em}
}{\begin{minipage}{\linewidth}\dissertationonly{\fontsize{8}{8}\selectfont} #1 \end{minipage}}
}

{
\catcode`\%=12
\begin{table*}
\begin{tabular}{cp{0.4\textwidth}ll}
\multicolumn{1}{l}{\textbf{ID}}               & \textbf{Morphology} & \textbf{Snapshot} & \textbf{Video} \\
\cellcolor[HTML]{4C72B0}{ \morphtext{a} } & \descript{Individual cells, no multicellular kin groups. Resource use is low---most cells simply hoard resource until their stockpile is beyond sufficient to reproduce. Only a handful of cells intermittently expend resource.} & \includesnapshot{\detokenize{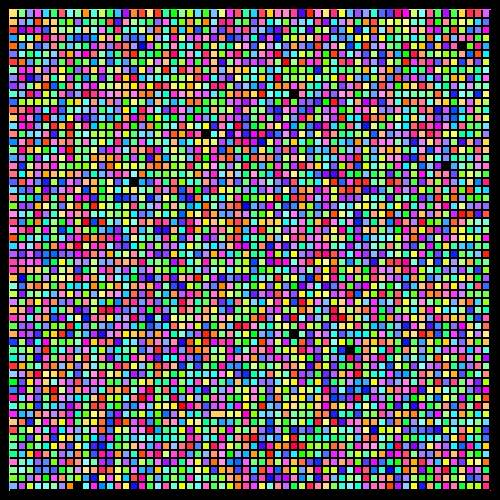}}             &  \videolink{https://hopth.ru/21/b=prq49+s=16005+t=0+v=video+w=specimen}          \\
\cellcolor[HTML]{DD8452}{\morphtext{b}} & \descript{Mostly individual cells, with some two-, three-, and four-cell groups evenly spread out. Resource usage occurs in short spurts in one or two adjacent cells. } & \includesnapshot{\detokenize{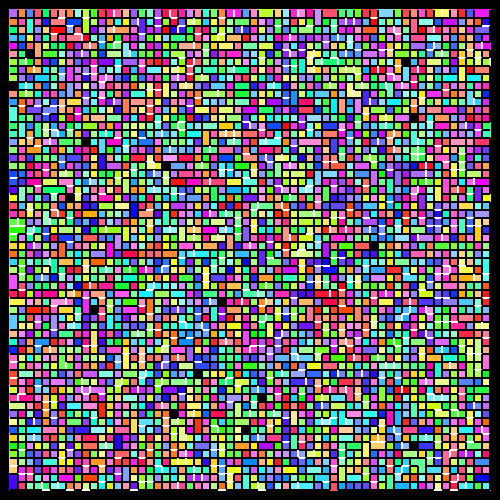}}                 & \videolink{https://hopth.ru/21/b=prq49+s=16005+t=1+v=video+w=specimen}              \\
\cellcolor[HTML]{55A868}{\morphtext{c}} & \descript{Large multicellular groups dominate, consisting of hundreds of cells. Group growth is unchecked and continues until cells' resource stockpiles are entirely depleted by the excess group size penalty.} & \includesnapshot{\detokenize{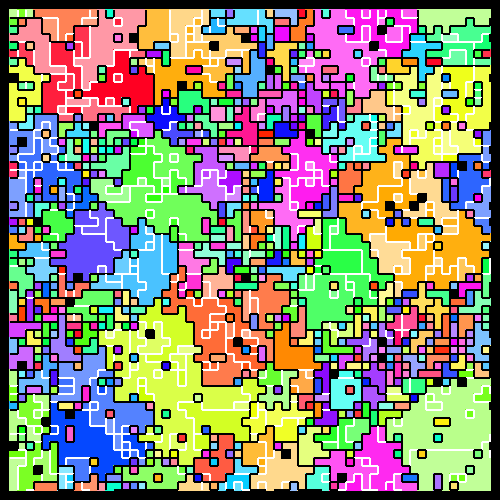}}                  & \videolink{https://hopth.ru/21/b=prq49+s=16005+t=2+v=video+w=specimen}             \\
\cellcolor[HTML]{C44E52}{\morphtext{d}} & \descript{Clear groups of 10 to 15 cells form. Cell proliferation appears somewhat more active at the periphery of groups compared to the interior.} & \includesnapshot{\detokenize{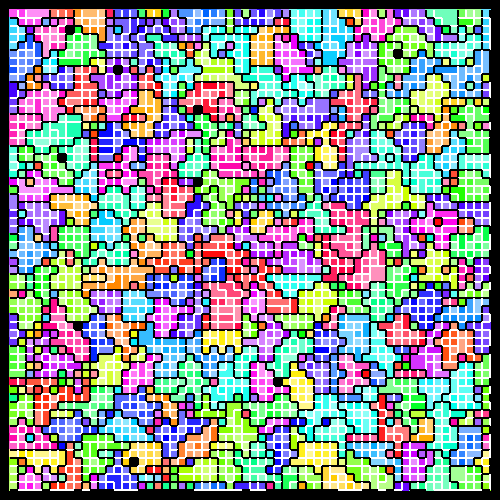}}                  & \videolink{https://hopth.ru/21/b=prq49+s=16005+t=14+v=video+w=specimen}               \\
\cellcolor[HTML]{8172B3}{\morphtext{e}} & \descript{Groups are visibly elongated along the horizontal axis. After initial development, some gradual, irregular growth occurs along the vertical axis.} & \includesnapshot{\detokenize{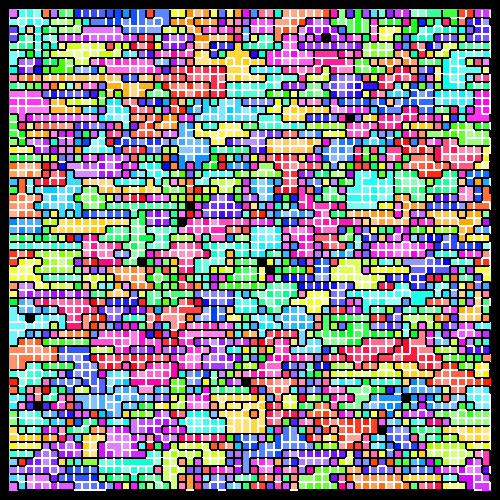}}                  & \videolink{https://hopth.ru/21/b=prq49+s=16005+t=15+v=video+w=specimen}              \\
\cellcolor[HTML]{937860}{\morphtext{f}} & \descript{Groups are horizontally elongated similarly to morphology $e$, but have a larger consistent vertical thickness of three or four cells.} & \includesnapshot{\detokenize{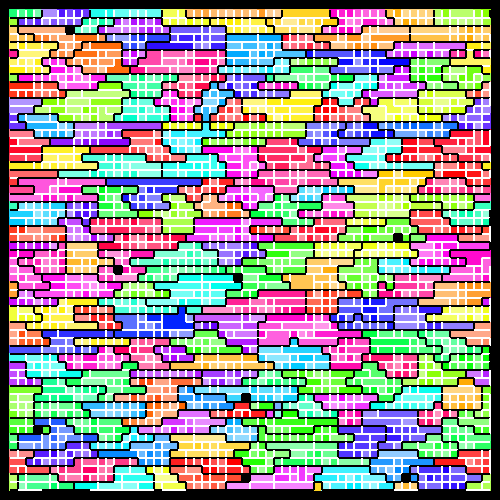}}                  & \videolink{https://hopth.ru/21/b=prq49+s=16005+t=39+v=video+w=specimen}             \\
\cellcolor[HTML]{DA8BC3}{\morphtext{g}} & \descript{Initial group growth is almost entirely horizontal, with groups usually taking up only one row of cells. However, after an apparent timing cue groups perform a brief bout of aggressive vertical growth.} & \includesnapshot{\detokenize{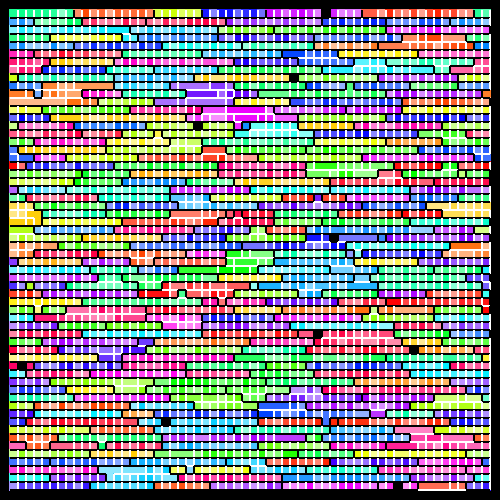}}                  & \videolink{https://hopth.ru/21/b=prq49+s=16005+t=45+v=video+w=specimen}              \\
\cellcolor[HTML]{8C8C8C}{\morphtext{h}} & \descript{Groups grow horizontally and then proliferate vertically on a timing cue like morph $e$. However, after that timing cue cell proliferation is incessant with almost no resource retention.} & \includesnapshot{\detokenize{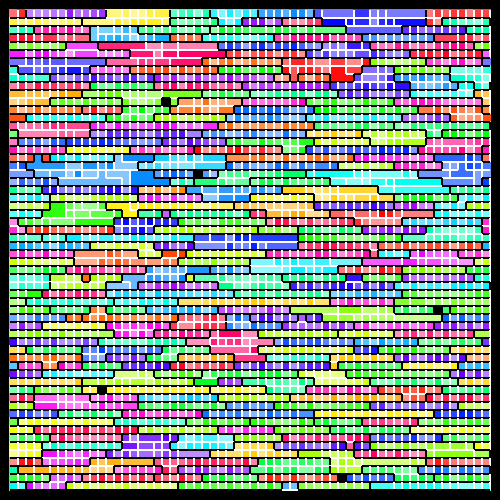}}                  & \videolink{https://hopth.ru/21/b=prq49+s=16005+t=59+v=video+w=specimen}              \\
\cellcolor[HTML]{CCB974}{\morphtext{i}} & \descript{Irregular groups of mostly less than ten cells. Incessant proliferation with almost no resource retention leads to rapid group turnover.} & \includesnapshot{\detokenize{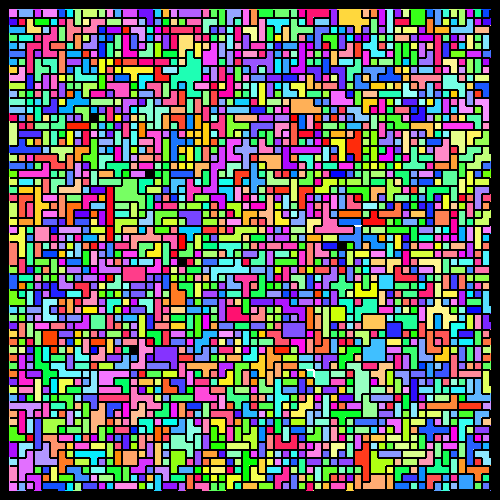}}                  & \videolink{https://hopth.ru/21/b=prq49+s=16005+t=74+v=video+w=specimen}              \\
\cellcolor[HTML]{64B5CD}{\morphtext{j}} & \descript{
Groups grow horizontally and then proliferate vertically on a timing cue like morph $e$. However, several viable horizontal-bar offspring groups form before forced fragmentation.} & \includesnapshot{\detokenize{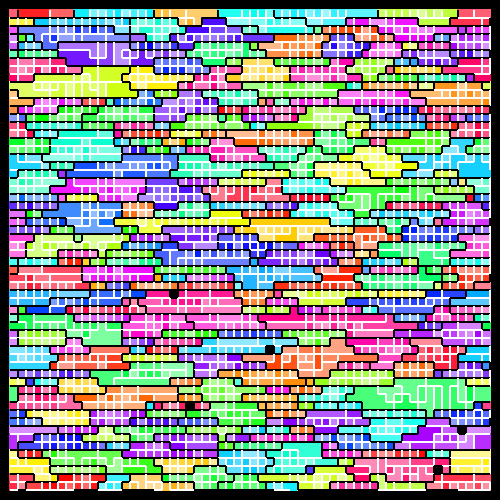}}                  & \videolink{https://hopth.ru/21/b=prq49+s=16005+t=100+v=video+w=specimen}
\end{tabular}

\caption{
\textbf{Qualitative morph phenotype categorizations.}
\footnotesize
Color coding of morph IDs has no significance beyond guiding the eye in scatter plots where points are labeled by morph.
Snapshot visualizes spatial layout of kin groups on toroidal grid at a fixed point in time.
Each cell corresponds to a small square tile.
Color hue denotes, and black borders divide, outermost kin groups; color saturation denotes, and white borders divide, innermost kin groups.
}
\label{tab:morph_descriptions}

\end{table*}
}

We performed a qualitative survey of the evolved life histories along the evolutionary timeline by reviewing video animations of the representative specimen from each stint in monoculture.

Table \ref{tab:morph_descriptions} summarizes the ten morphological categories we grouped specimens into.
In brief, specimens from early stints largely grew as unicellular or small multicellular groups (morphs $a$, $b$).
Then, the specimen from stint 14 grew as larger, symmetrical groups (morph $d$).
At stint 15, a distinct, asymmetrical horizontal bar morphology evolved (morph $e$).
At stint 45, a delayed secondary spurt of group growth in the vertical direction arose (morph $g$).
This morphology was sampled frequently until stint 60, when morph $e$ began to be sampled primarily again.
However, morph $g$ was observed as late as stint 90.

Phylogenetic analysis (Figure \ref{fig:phylo_parsimony_tree}) indicates that observations of morph $e$ at stint 53 and onward are instances of secondary loss rather than retention of trait $e$ by a separate lineage coexisting with the lineage expressing morph $g$.
Three separate reversion events from morph $g$ to morph $e$ appear likely.
Interestingly, morph $g$ individuals at stints 89 and 90 appear to represent subsequent trait re-gain after reversion from morph $g$ to morph $e$.

Table \ref{tab:morph_descriptions} provides more detailed descriptions of each qualitative morph category, as well as a video and still image example of each.

Supplementary Table \ref{tab:morph_by_stint} provides morph categorization for each stint as well as links to view the stint's specimen in a video or in-browser web simulation.

\subsection{Fitness}

\begin{figure*}

\includegraphics[width=\linewidth]{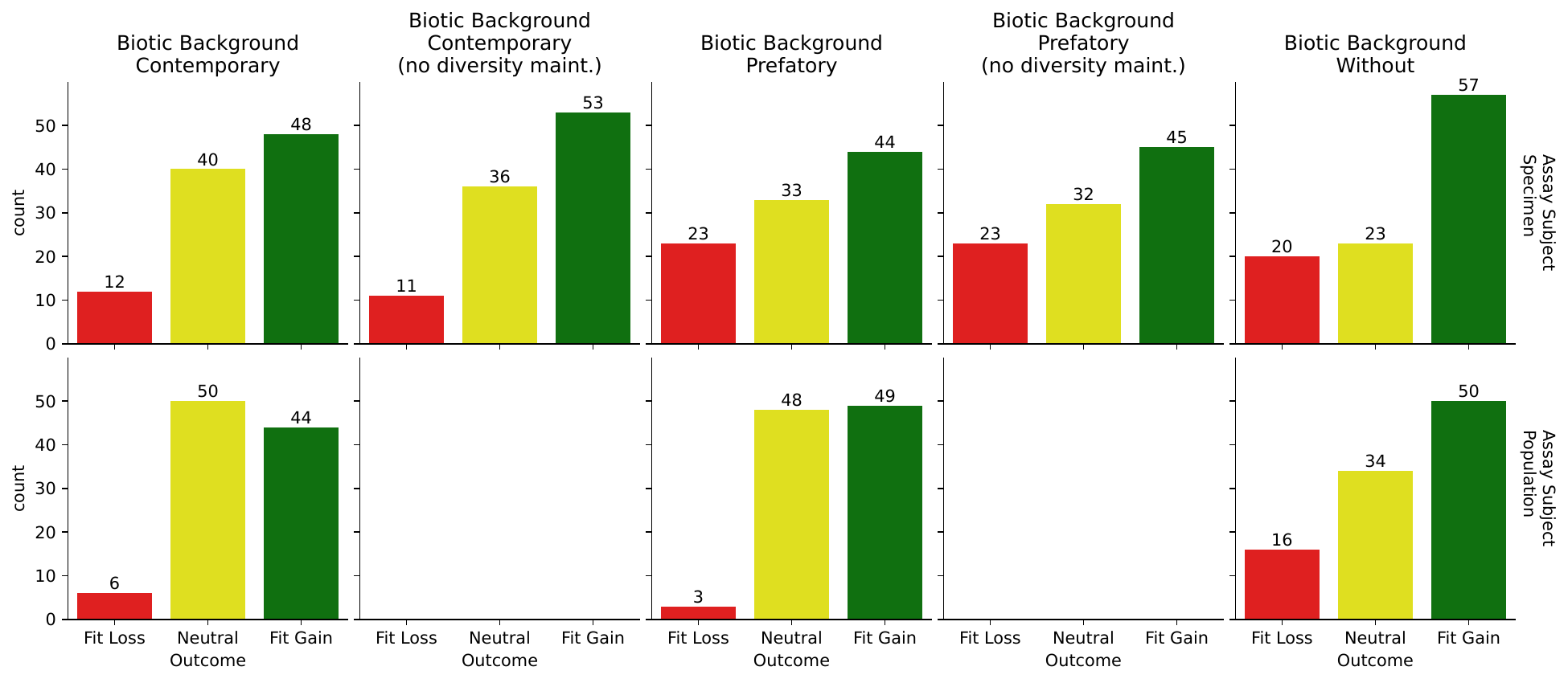}

\caption{
\textbf{Distribution of adaptation assay outcomes.}
\footnotesize
For each adaptation assay, three outcomes were possible: significant fitness gain, significant fitness loss, or no significant fitness change (``neutral'').
Significance cutoff $p < 0.005$ was used.
A fitness loss --- color-coded red --- corresponds to winning 2 or fewer competitions out of 20 against the preceding stint's focal strain population.
A fitness gain --- color-coded green --- corresponds to winning 18 or more competitions out of 20 against the preceding stint's focal strain population.
Neutral fitness outcomes are color-coded yellow.
Outcome counts are accumulated over experiments from stint 1 through stint 100.
Upper row shows results for sampled focal strain genome, lower row shows results for entire focal strain population.
See Figure \ref{fig:adaptation_assay_cartoon} for explanation of competition biotic backgrounds.
See Figure \ref{fig:outcome_count_joint_distns} for joint distributions of fitness outcomes across biotic backgrounds.
}
\label{fig:outcome_count_distns}
\end{figure*}

Of the 100 competition assays performed, 57 indicated significant fitness gain, 23 were neutral, and 20 indicated significant fitness loss (shown in upper right of Figure \ref{fig:outcome_count_distns}, at the intersection of the ``Biotic Background, Without'' column and ``Assay Subject, Specimen'' row.)

We were surprised by the frequency of deleterious outcomes, leading us to perform a second set of experiments to investigate whether these outcomes could be explained as sampling of ``dud'' representatives.
In these competition assays, we competed the entire focal strain population against the focal strain population from the preceding stint.
However, we observed a similar result: 50 assays indicated significant fitness gain, 34 were neutral, and 16 indicated significant fitness loss (shown in lower right of Figure \ref{fig:outcome_count_distns}, at the intersection of the ``Biotic Background, Without'' column and ``Assay Subject, Population'' row.)

\begin{figure*}
\centering
\includegraphics[width=0.5\linewidth]{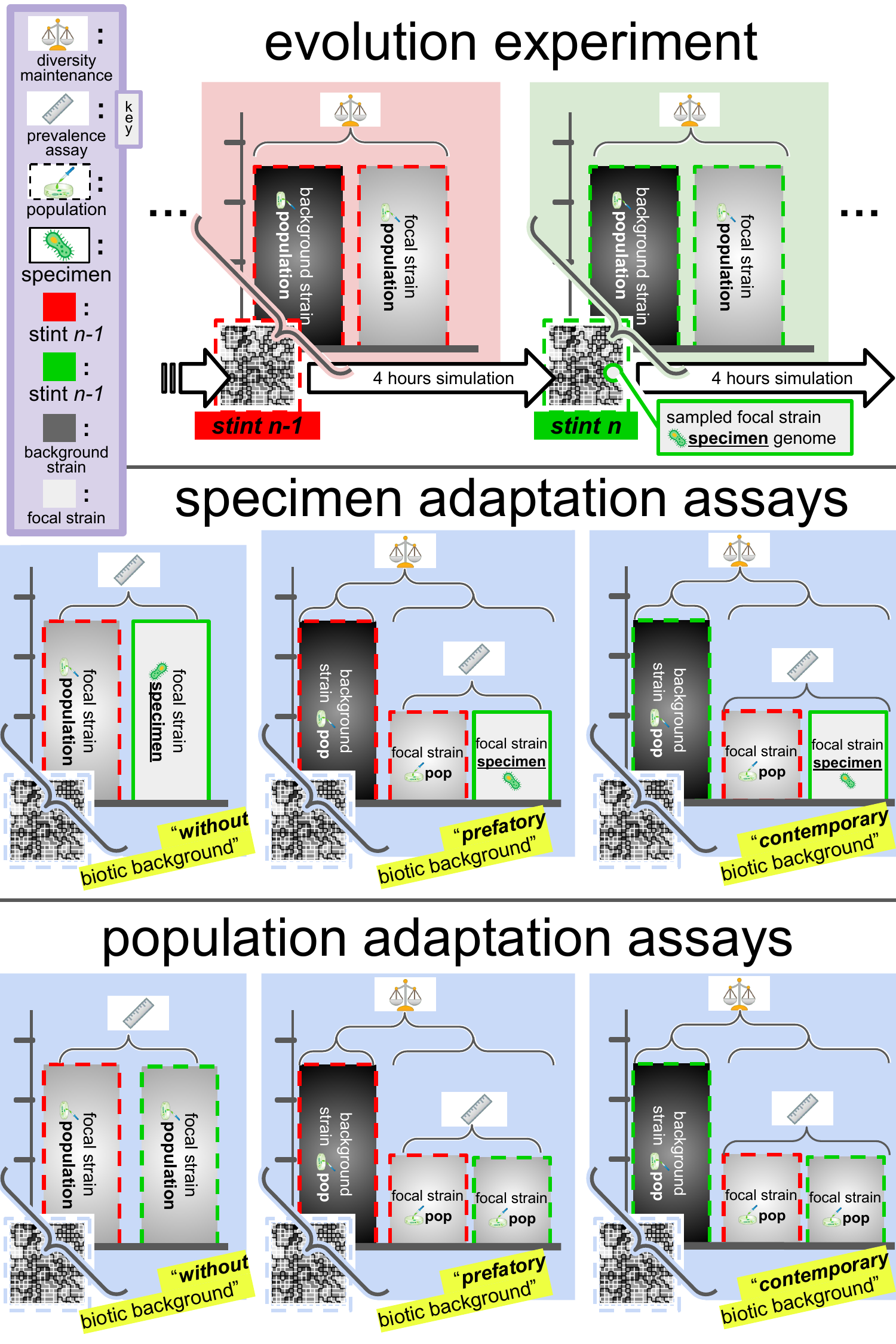}

\caption{
\textbf{Adaptation assay design.}
\footnotesize
Top panel shows progress of original evolutionary experiment over one stint.
A diversity maintenance procedure was used to ensure long-term coexistence of at least two strains over the course of the experiment by penalizing any strain that occupied more than half of thread-local population space.
A ``focal strain'' was arbitrarily chosen for study; we refer to the other strain as the ``background strain.''
Adaptation assays in lower panels measure fitness change over the course of that stint through competition against the population from the preceding stint.
The middle panel shows measurement of adaptation of the representative specimen that was sampled for analysis at each stint.
The bottom panel shows measurement of the adaptation of the entire focal strain population at each stint.
Competitors were mixed in even proportion into the environment.
Bar heights represent initial relative proportions of assay participants at the beginning of the competition.
Adaptation was measured by measuring change in population composition over a 10 minute competition window.
We call this measurement of population composition change a ``prevalence assay''.
Competition experiments were performed absent the background strain, with the background strain population from the preceding stint, or with the background strain population from the current stint --- shown separately in each panel.
}
\label{fig:adaptation_assay_cartoon}
\end{figure*}

Next, we investigated whether the presence of the background strain as a ``biotic background'' influenced fitness.
We repeated the two experiments described above (specimen and population competition assays), but inserted the background strain as half of the initial well-mixed population.
In one assay setup, we used the background strain population from the current stint.
We refer to this as ``contemporary biotic background.''
In another, which we call ``prefatory biotic background,'' we used the background strain population from the previous stint.
We refer to the original competition assays absent the background strain as ``without biotic background.''
Figure \ref{fig:adaptation_assay_cartoon} summarizes these competition assay designs.

\begin{sidewaysfigure*}
\thisfloatpagestyle{mylandscape}%
\rotatesidewayslabel%
\centering

\begin{subfigure}[t]{0.32\textwidth}
\includegraphics[width=\linewidth]{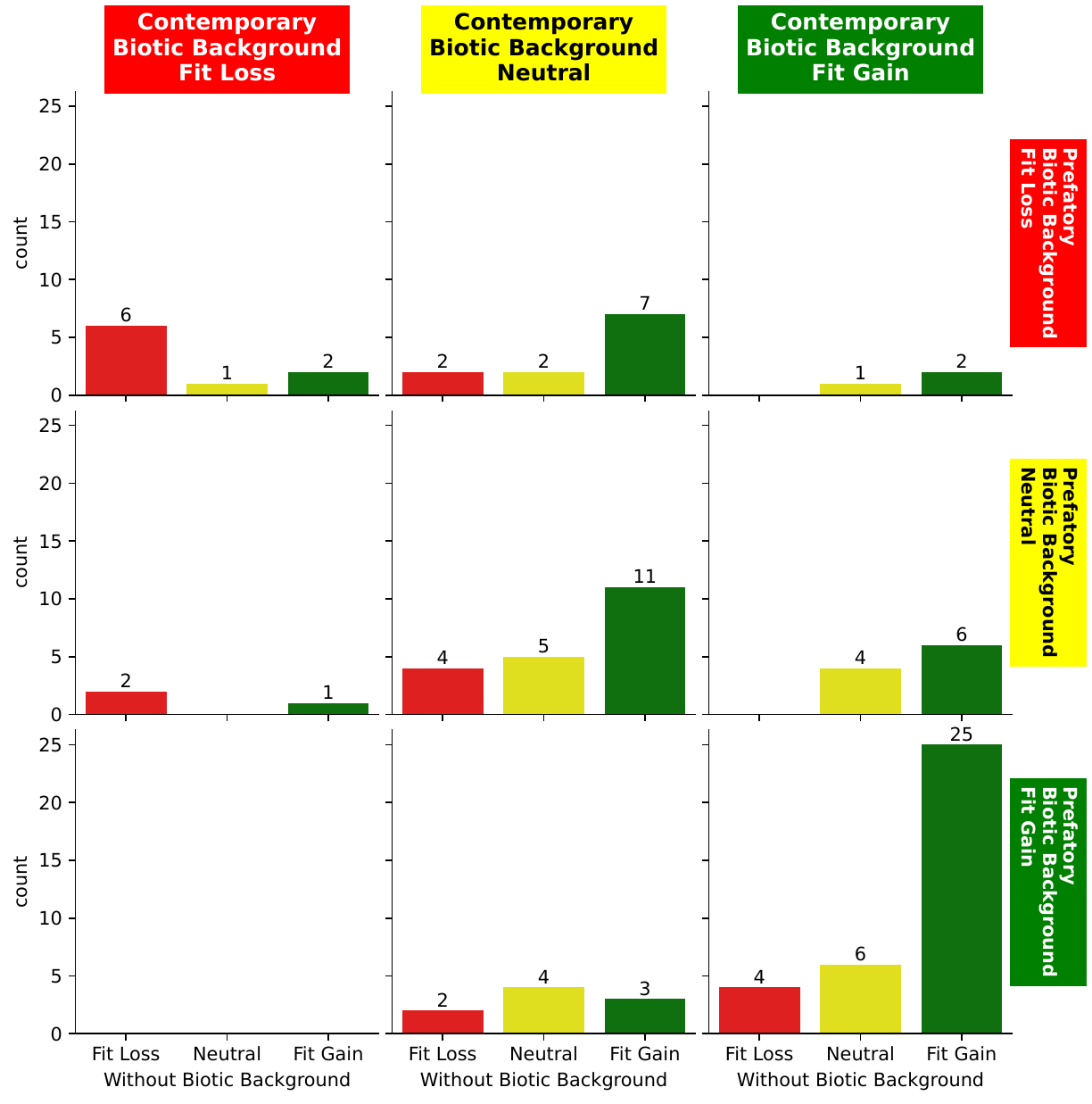}
\caption{Joint distribution of adaptation assay on representative specimen from focal strain over biotic backgrounds, with diversity maintenance during competition.}
\label{fig:outcome_count_joint_distns:specimen_with_dm}
\end{subfigure}%
\hfill%
\begin{subfigure}[t]{0.32\textwidth}
\includegraphics[width=\linewidth]{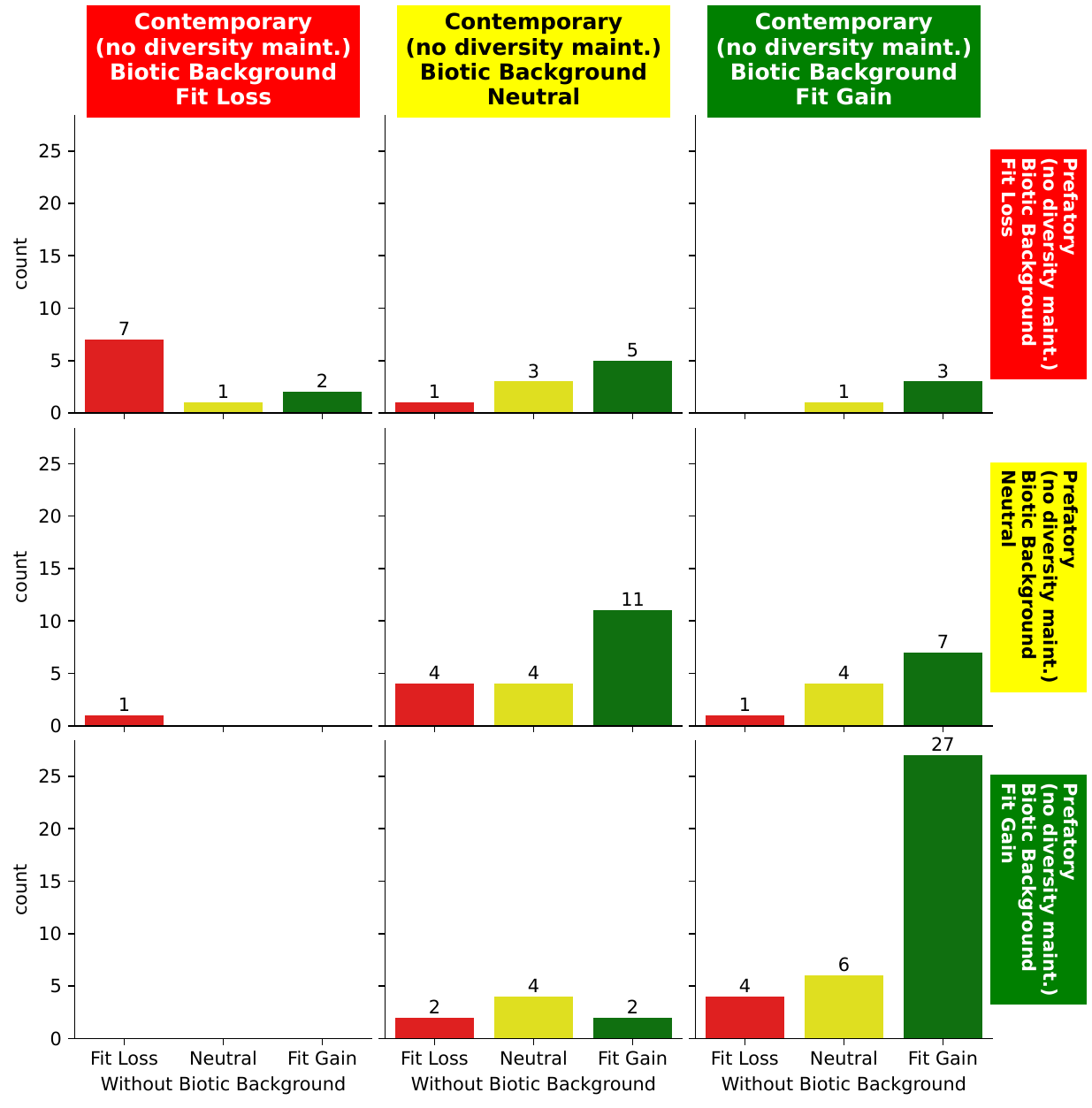}
\caption{Joint distribution of adaptation assay on representative specimen from focal strain over biotic backgrounds, with diversity maintenance disabled during competition.}
\label{fig:outcome_count_joint_distns:specimen_no_dm}
\end{subfigure}%
\hfill%
\begin{subfigure}[t]{0.32\textwidth}
\includegraphics[width=\linewidth]{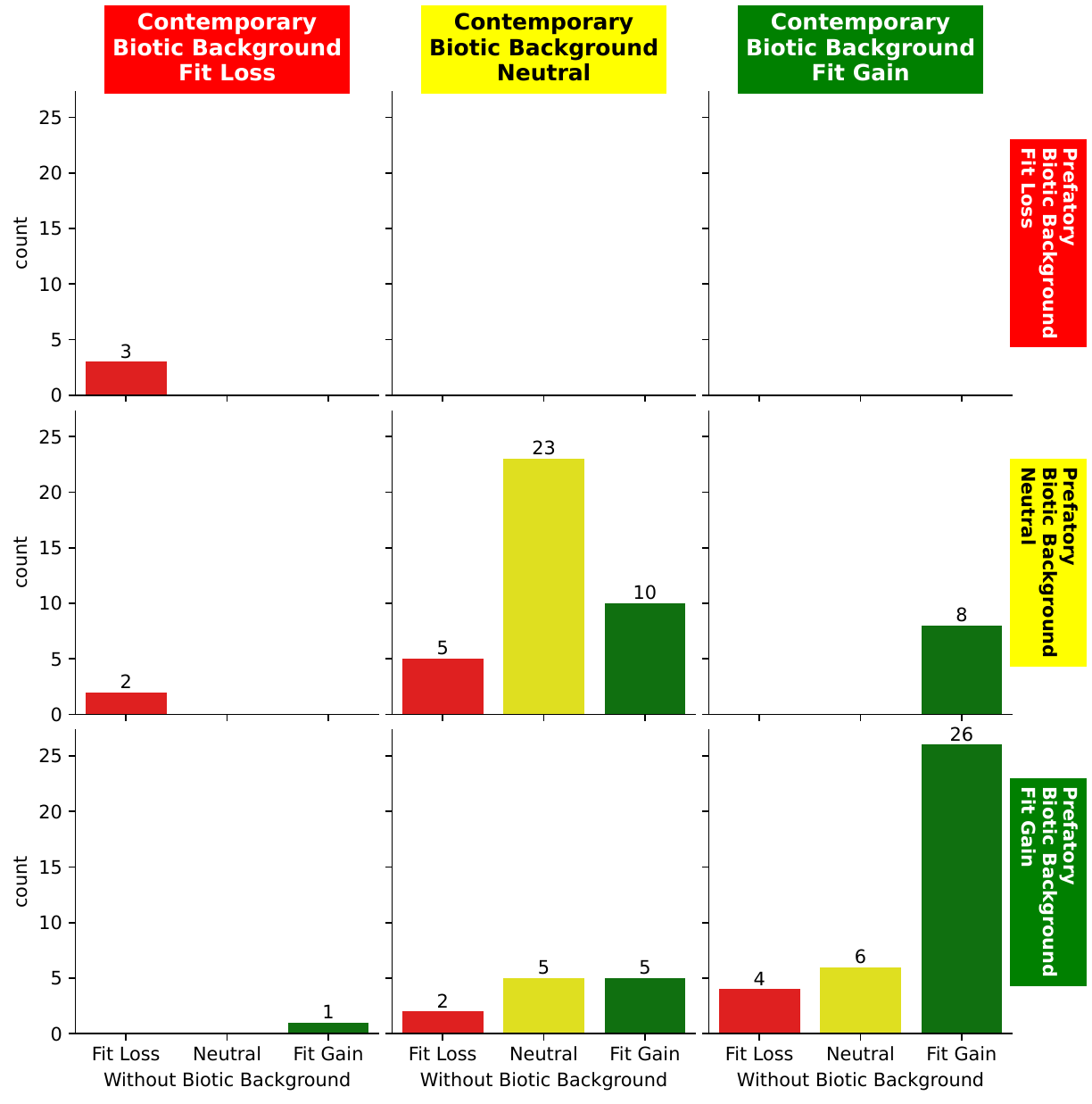}
\caption{Joint distribution of adaptation assay on focal strain population over biotic backgrounds, with diversity maintenance during competition.}
\label{fig:outcome_count_joint_distns:population}
\end{subfigure}

\caption{
\textbf{Effect of biotic background on adaptation assay outcomes.}
\footnotesize
Joint distribution of adaptation assay outcomes across biotic backgrounds.
For each adaptation assay, three outcomes were possible: significant fitness gain, significant fitness loss, or no significant fitness change (``neutral'').
Significance cutoff $p < 0.005$ was used.
A fitness loss --- color-coded red --- corresponds to winning 2 or fewer competitions out of 20 against the preceding stint's focal strain population.
A fitness gain --- color-coded green --- corresponds to winning 18 or more competitions out of 20 against the preceding stint's focal strain population.
Neutral fitness outcomes are color-coded yellow.
Outcome counts are accumulated over experiments from stint 1 through stint 100.
Counts in each subfigure therefore sum to 100.
Column position in facet grid indicates outcome with contemporary biotic background, row position indicates outcome with prefatory biotic background, and bar color and $x$ position indicates outcome without biotic background.
See Figure \ref{fig:adaptation_assay_cartoon} for explanation of competition biotic backgrounds.
See Figure \ref{fig:with_vs_without_diversity_maintenance} for detail on joint distribution of outcomes with and without diversity maintenance, which were mostly identical.
}
\label{fig:outcome_count_joint_distns}
\end{sidewaysfigure*}

After incorporating the background strain into our measure of fitness, we detected fewer whole-population deleterious outcomes --- only six under contemporary biotic background conditions and only three under prefatory biotic background conditions (Figure \ref{fig:outcome_count_distns}).
To determine whether the presence of the background strain caused the overall reduction in whole-population deleterious outcomes, we performed control competitions under biotic background conditions, but with the focal strain population substituted for the background strain population (Supplementary Figure \ref{fig:baseline_adaptation_control}).
Under these conditions, nine of the stints where whole-population deleterious outcomes had been detected came up neutral;
one, surprisingly, tested significantly adaptive (Supplementary Figure \ref{fig:baseline_fitness_gain_or_loss}).
Dose-dependent fitness effects and/or reduced experimental sensitivity of the biotic background assay appear to play at least a partial role in explaining the reduction of detected whole-population deleterious outcomes.
However, 10 stints still tested significantly deleterious with the control focal strain biotic background in addition to without biotic background.

Four stints do provide strong, direct evidence of a selective effect by the background strain: four whole-population outcomes that were deleterious without their biotic background were actually significantly advantageous in the presence of both the prefatory and contemporary background strain populations (Figure \ref{fig:outcome_count_joint_distns:population}).
All four of these stints exhibited whole-population deleterious outcomes under the control focal strain biotic background, indicating that the observed fitness sign change was specifically due to the presence of the background strain (Supplementary Figure \ref{fig:baseline_fitness_gain_or_loss}).

Additionally, we detected two deleterious outcomes without biotic background as significantly adaptive under the prefatory biotic background but as neutral under contemporary biotic background (Figure \ref{fig:outcome_count_joint_distns:population}).
Control focal strain biotic background experiments again suggest that the background strain, specifically, is responsible for this effect (Supplementary Figure \ref{fig:baseline_fitness_gain_or_loss}).

We also found one whole-population outcome that was significantly advantageous without biotic background and in the presence of the prefatory background strain population but significantly deleterious in the presence of the contemporary background strain, possibly suggesting an ``arms race''-like evolutionary innovation on the part of the background strain over that stint (Figure \ref{fig:outcome_count_joint_distns:population}).

Nonetheless, we still saw three whole-population outcomes that were significantly deleterious under all three conditions (Figure \ref{fig:outcome_count_joint_distns:population}).
These whole-population outcomes were also deleterious under the control focal strain biotic background experiments (Supplementary Figure \ref{fig:baseline_fitness_gain_or_loss}).
Muller's ratchet \citep{andersson1996muller} or maladaptation due to environmental change \citep{brady2019causes} may offer possible explanations, but a definitive answer will require further study.

We also performed fitness assays on individual sampled specimens with both biotic backgrounds.
Out of 100 stints tested, we observed 20 significantly deleterious outcomes without biotic background, 23 significantly deleterious outcomes under prefatory biotic background, and 12 significantly deleterious outcomes under contemporary biotic background (Figure \ref{fig:outcome_count_distns}).
Unlike the whole-population deleterious outcomes discussed above, observing some deleterious outcomes from sampled specimens is not surprising.
Evolving populations naturally contain standing variation in fitness \citep{martin2016nonstationary}, so occasional sampling of less-fit individuals should be expected.
Reciprocally, we observed 57 significantly adaptive outcomes without biotic background, 44 with prefatory biotic background, and 48 with contemporary biotic background (Figure \ref{fig:outcome_count_distns}).
Greater sensitivity of the ``without biotic background'' adaptation assay could account for the counterintuitive detection of more adaptive outcomes under abiotic conditions (i.e., the absence of the background strain).

As before with the population-level adaptation assays, we detected four specimen outcomes that were deleterious without biotic background but significantly advantageous under both tested background strain populations (Figure \ref{fig:outcome_count_joint_distns:specimen_with_dm}).
Additionally, and again as before, we detected two deleterious outcomes without biotic background as significantly adaptive under the prefatory biotic background but as neutral under the contemporary biotic background (Figure \ref{fig:outcome_count_joint_distns:specimen_with_dm}).
Control focal strain biotic background experiments confirm that the background strain, specifically, is responsible for these effects (Supplementary Figure \ref{fig:baseline_fitness_gain_or_loss}).

We found no specimen outcomes that were advantageous under the prefatory biotic background but deleterious under the contemporary background.
However, we found three stints with opposite dynamics: specimen outcomes deleterious under prefatory biotic background but advantageous under contemporary biotic background (Figure \ref{fig:outcome_count_joint_distns:specimen_with_dm}), further suggesting coincident, interacting evolutionary innovations along focal and background strain lineages (Figure \ref{fig:outcome_count_joint_distns:specimen_with_dm}).

\begin{figure*}
\centering
\captionsetup[subfigure]{font=footnotesize}
\begin{subfigure}[t]{0.49\textwidth}
\includegraphics[width=\linewidth]{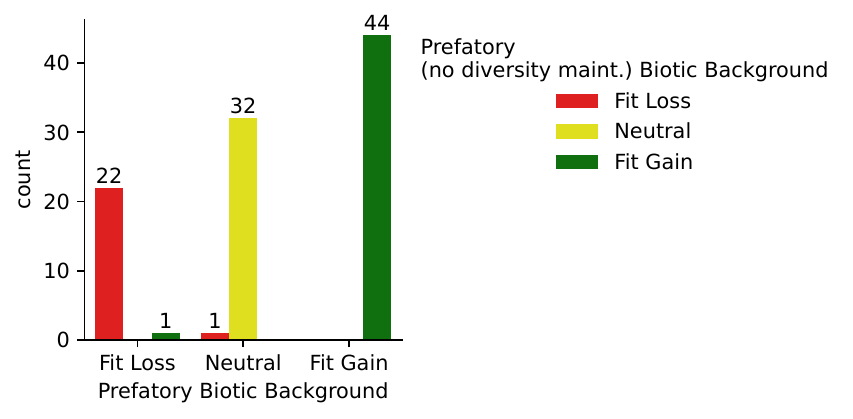}
\caption{prefatory biotic background outcomes with and without diversity maintenance}
\end{subfigure}%
\hfill%
\begin{subfigure}[t]{0.49\textwidth}
\includegraphics[width=\linewidth]{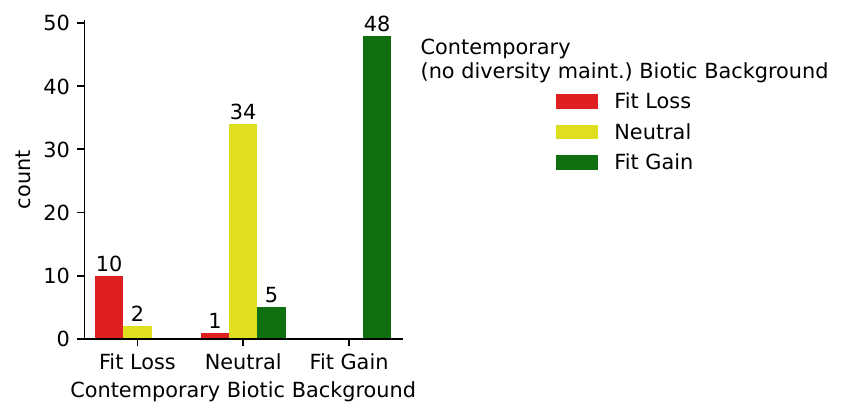}
\caption{contemporary biotic background outcomes with and without diversity maintenance}
\end{subfigure}

\caption{
\textbf{Effect of diversity maintenance on adaptation assays with biotic background.}
\footnotesize
Joint distribution of competition experiments performed under biotic background conditions with diversity maintenance enabled and disabled.
Color coding denotes outcome without diversity maintenance and $x$ position denotes outcome with diversity maintenance.
Note that both plots above show distributions for adaptation assays on representative specimens.
Competition experiments without diversity maintenance were not performed for population-level adaptation.
See Figure \ref{fig:adaptation_assay_cartoon} for explanation of competition biotic backgrounds.
}
\label{fig:with_vs_without_diversity_maintenance}
\end{figure*}

To better characterize the mechanism behind fitness effects caused by the background strain, we performed additional specimen adaptation assays under biotic background conditions with diversity maintenance disabled.
This analysis allowed us to test whether action of the diversity maintenance mechanism, rather than direct interactions between the focal and background strains, caused the observed fitness effects.
Figure \ref{fig:with_vs_without_diversity_maintenance} compares adaptation assay outcomes with and without diversity maintenance under both the prefatory and contemporary biotic background conditions.
Outcomes were generally similar, and we observed only one sign-change difference: one specimen outcome was beneficial under prefatory biotic background conditions without diversity maintenance, but deleterious with diversity maintenance.
Further, without diversity maintenance, we still observed four outcomes that were advantageous under biotic conditions but deleterious under abiotic conditions (Figure \ref{fig:outcome_count_joint_distns:specimen_no_dm}).
So, biotic selective effects cannot be explained as an artifact of the diversity maintenance scheme.%
\footnote{
We also conducted specimen adaptation assays with diversity maintenance disabled under the control focal strain biotic background.
In these experiments, we again found no evidence for impact from the diversity maintenance scheme on results
(Supplementary Figure \ref{fig:baseline_fitness_gain_or_loss}).
}

\begin{sidewaysfigure*}
\thisfloatpagestyle{mylandscape}%
\rotatesidewayslabel%
\centering
\includegraphics[width=\linewidth]{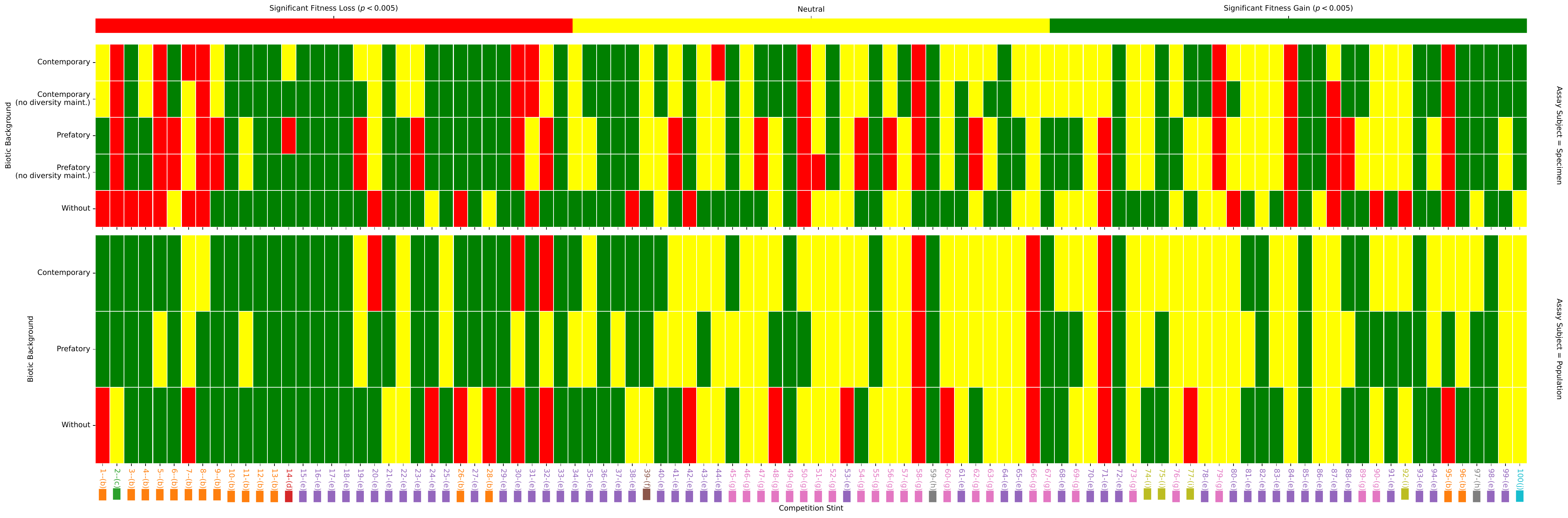}

\caption{
Summary of adaptation assay outcomes for sampled representative specimen (top) and population-level adaptation (bottom).
Color coding and parentheticals of stint labels correspond to qualitative morph codes described in Table \ref{tab:morph_descriptions}.
See Figure \ref{fig:adaptation_assay_cartoon} for explanation of competition biotic backgrounds.
}
\label{fig:fitness_gain_or_loss}
\end{sidewaysfigure*}

Significant increases in fitness occur throughout the evolutionary history of the case study, but not at every stint.
Figure \ref{fig:fitness_gain_or_loss} summarizes the outcome of all adaptation assays stint-by-stint across evolutionary history.
Neutral outcomes appear to occur more frequently at later stints.
This may be indicative of slower evolutionary innovation, but may also result to some extent from simulation of fewer generations during evolutionary stints (Supplementary Figure \ref{fig:simulation}) and during competition experiments (Supplementary Figure \ref{fig:num_updates_elapsed_barplot}) due to slower execution of later genomes.

\begin{sidewaysfigure*}
\thisfloatpagestyle{mylandscape}%
\rotatesidewayslabel%
\centering
\includegraphics[width=\linewidth]{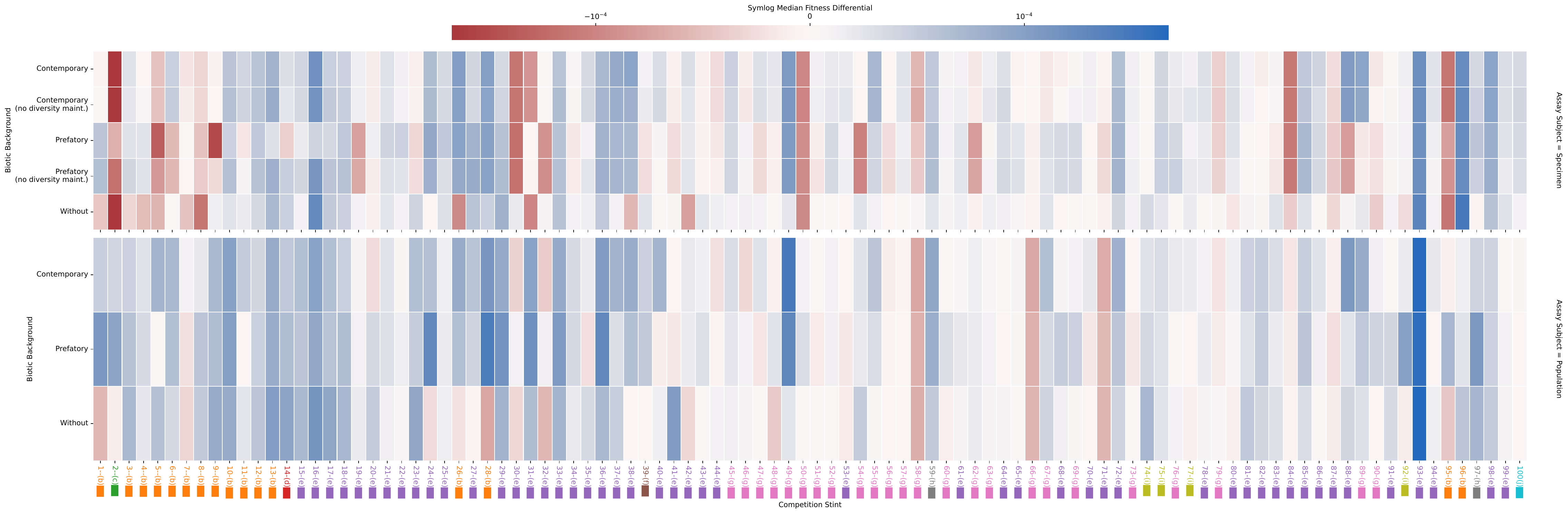}

\caption{
\textbf{Adaptation assay effect strengths.}
\footnotesize
Median calculated fitness differential outcomes of competition experiments.
Zero fitness differential corresponds to a neutral result, color mapped to white.
Blue indicates positive fitness differential (fitness gain) compared to the previous stint and red indicates negative fitness differential (fitness loss).
Color coding and parentheticals of stint labels correspond to qualitative morph codes described in Table \ref{tab:morph_descriptions}.
Note that color intensity is plotted on a symlog scale due to distribution of fitness differentials over multiple orders of magnitude.
Upper panels shows results for sampled focal strain genome, lower panel shows results for entire focal strain population.
See Figure \ref{fig:adaptation_assay_cartoon} for explanation of competition biotic backgrounds.
}
\label{fig:median_fitness_differential_symlog}
\end{sidewaysfigure*}

Figure \ref{fig:median_fitness_differential_symlog} shows the magnitudes of calculated fitness differentials for all adaptation assays.
Fitness differentials during the first 40 stints are generally higher magnitude than later fitness differentials, although a strong fitness differential occurs at stint 93.
Although the emergence of morphology $d$ was associated with significant increases in fitness in some specimen assays and morphologies $e$ and $g$ were associated with significant increases in fitness across all specimen assays (Figure \ref{fig:fitness_gain_or_loss}), the magnitude of these fitness differentials appears ordinary compared to fitness differentials at other stints (Figure \ref{fig:median_fitness_differential_symlog}).
Supplementary Figure \ref{fig:mean_competition_prevalence} shows mean end-competition prevalence across assays, telling a similar story.

\begin{figure}
\centering
\includegraphics[width=0.5\textwidth]{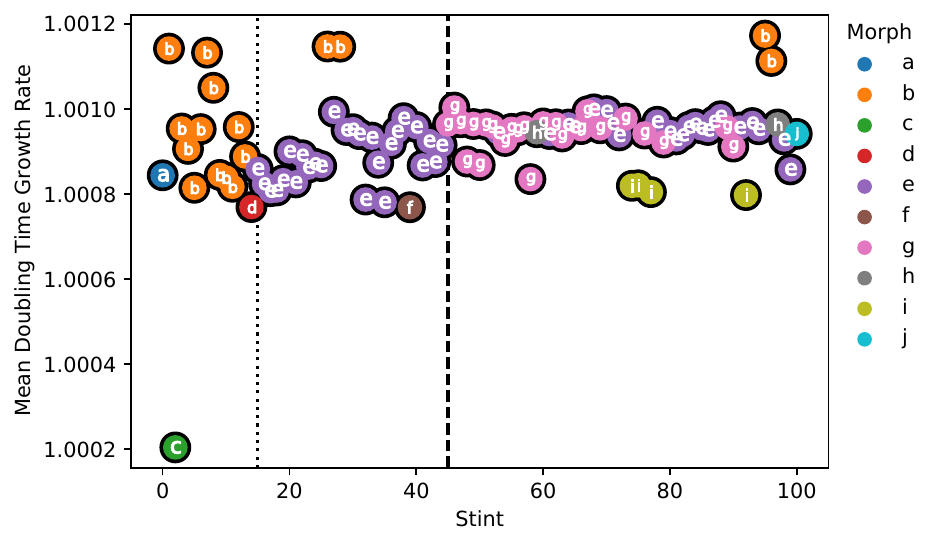}

\caption{
\textbf{Growth rate estimates.}
\footnotesize
Estimates made from doubling time experiments, measuring time for a monoculture to grow from 0.25 maximum population size to 0.5 maximum population size.}
\label{fig:doubling_time}

\end{figure}

In addition to competition assays, we also measured growth rate of specimen strains by tracking doubling time (in updates) when seeded into quarter-full toroidal grids (Figure \ref{fig:doubling_time}).
Morph $b$ exhibited a fast growth rate early on that was never matched by later morphs.
This measure appears to be a poor overall proxy for fitness, highlighting the importance of biotic aspects of the simulation environment, which are not present in the empty space the assayed cells double into.

\subsection{Fitness Complexity}

\begin{figure}
\centering
\includegraphics[width=0.5\textwidth]{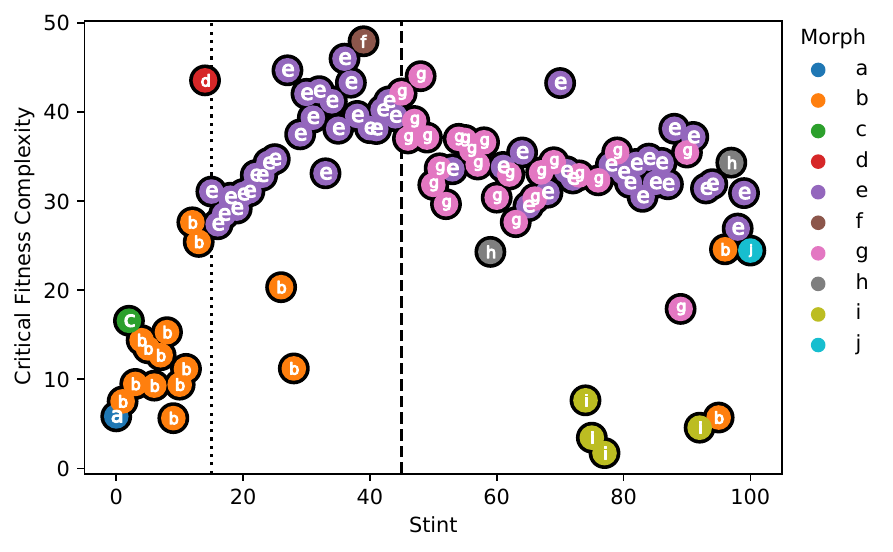}

\caption{
\textbf{Critical fitness complexity.}
\footnotesize
Number of single-site nopouts that significantly decrease fitness, adjusted for expected false positives.
Color coding and letters correspond to qualitative morph codes described in Table \ref{tab:morph_descriptions}.
Dotted vertical line denotes emergence of morph $e$.
Dashed vertical line denotes emergence of morph $g$.
}
\label{fig:critical_fitness_complexity}

\end{figure}%

Figure \ref{fig:critical_fitness_complexity} plots critical fitness complexity of specimens drawn from across the case study's evolutionary history.

Critical fitness complexity reaches more than 20 under morph $b$, jumps to more than 40 under morph $d$, and drops to slightly more than 30 for morph $e$.
Critical fitness complexity reaches a peak of 48 sites around stint 39, then levels out and decreases.
This decrease may in part be due to declining sensitivity of competition experiments due to slower simulation resulting in execution of fewer updates within the fixed-duration jobs (Supplementary Figure \ref{fig:num_updates_elapsed_heatmap}).

Phylogenetic analysis (Figure \ref{fig:phylo_parsimony_tree}) suggests independent origins of the critical fitness complexity in morph $d$ and morph $e$ --- the morph $d$ specimen from stint 14 is more closely related to the morph $b$ specimen from stint 13 than to the morph $e$ specimen from stint 15.
Likewise, specimens of lower complexity morphs $i$ and $b$ that appear past stint 70 appear to have independent evolutionary origins.




\subsection{Interface Complexity}


\providecommand{\dissertationonly}[1]{%
\ifdefined\DISSERTATION%
#1%
\else%
\fi
}

\begin{figure*}
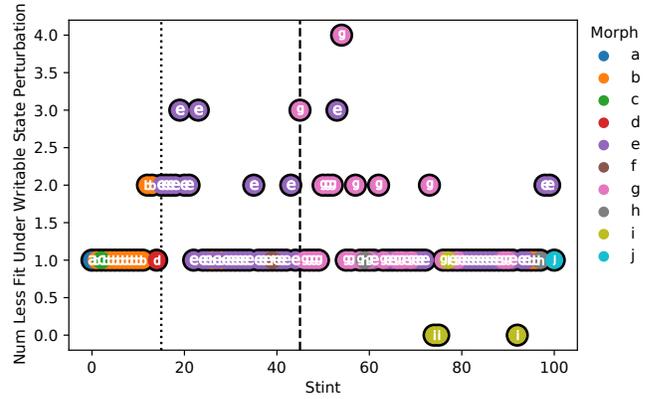
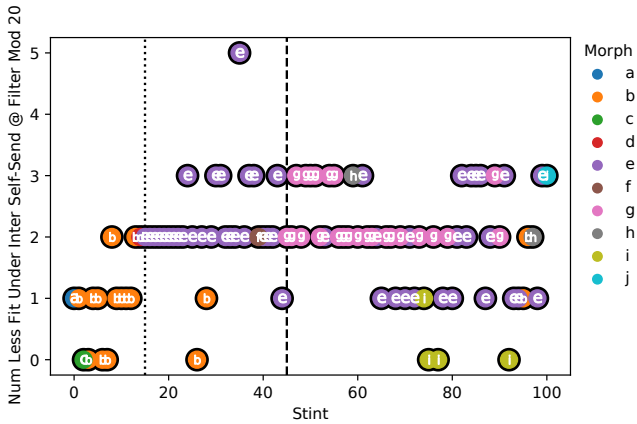
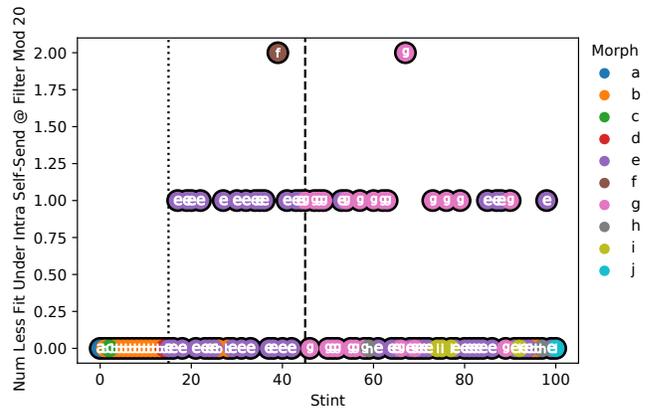
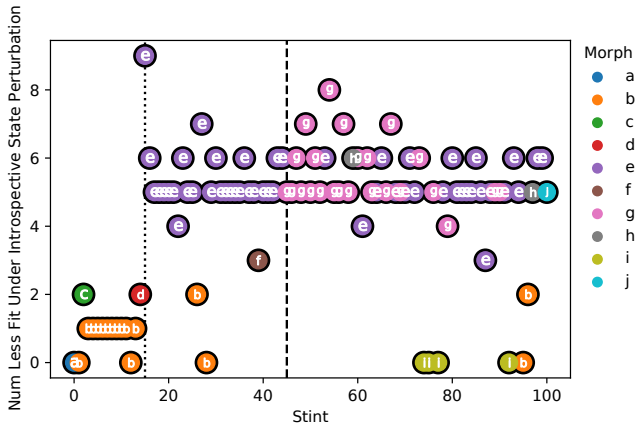
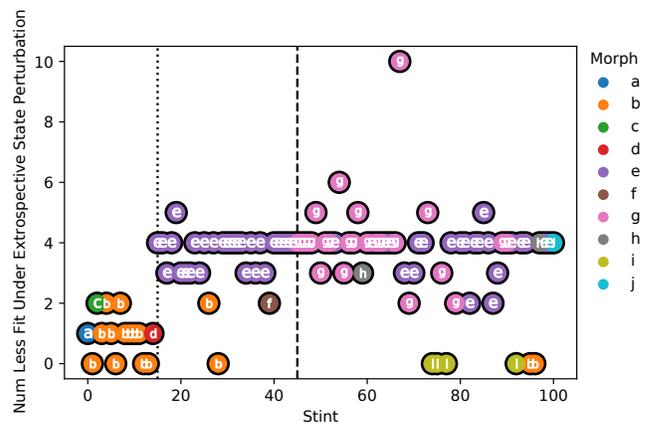

\captionsetup[subfigure]{font=footnotesize}
\input{fig/interface_complexity/cardinal_interface_complexity.tex}
\hfill
\input{fig/interface_complexity/writable_interface_complexity.tex}

\input{fig/interface_complexity/intermessage_interface_complexity.tex}
\hfill
\input{fig/interface_complexity/intramessage_interface_complexity.tex}

\input{fig/interface_complexity/introspective_interface_complexity.tex}
\hfill
\input{fig/interface_complexity/extrospective_interface_complexity.tex}

\caption{
\textbf{Interface complexity estimates.}
\footnotesize
Color coding and letters correspond to qualitative morph codes described in Table \ref{tab:morph_descriptions}.
Dotted vertical line denotes emergence of morph $e$.
Dashed vertical line denotes emergence of morph $g$.}
\label{fig:interface_complexity}
\end{figure*}

Figure \ref{fig:interface_complexity} summarizes cardinal processor interface complexity, as well as its constituent components, for specimens drawn from across the case study's evolutionary history.

Notably, cardinal processor interface complexity more than doubles from 6 interactions to 17 interactions coincident with the emergence of morph $e$ (Figure \ref{fig:interface_complexity:cardinal_interface_complexity}).
This is due to simultaneous increases in extrospective state sensing (2 to 9 states; Figure \ref{fig:interface_complexity:extrospective_interface_complexity}), introspective state sensing (1 to 4 states; Figure \ref{fig:interface_complexity:introspective_interface_complexity}), and writable state usage (1 to 2 states; Figure \ref{fig:interface_complexity:writable_interface_complexity}).

The emergence of morph $g$ coincided with an increase in writable state interface complexity from 1 to 3, as shown in Figure \ref{fig:interface_complexity:writable_interface_complexity}.
However, morph $g$ was not associated with other changes in other aspects of cardinal processor interface complexity.
The greatest observed cardinal processor interface complexity was 22 interactions at stints 54 and 67.

\subsection{Genome Size}

\begin{figure}
\centering
\captionsetup[subfigure]{font=footnotesize}
\begin{subfigure}{0.49\textwidth}
\includegraphics[width=\linewidth]{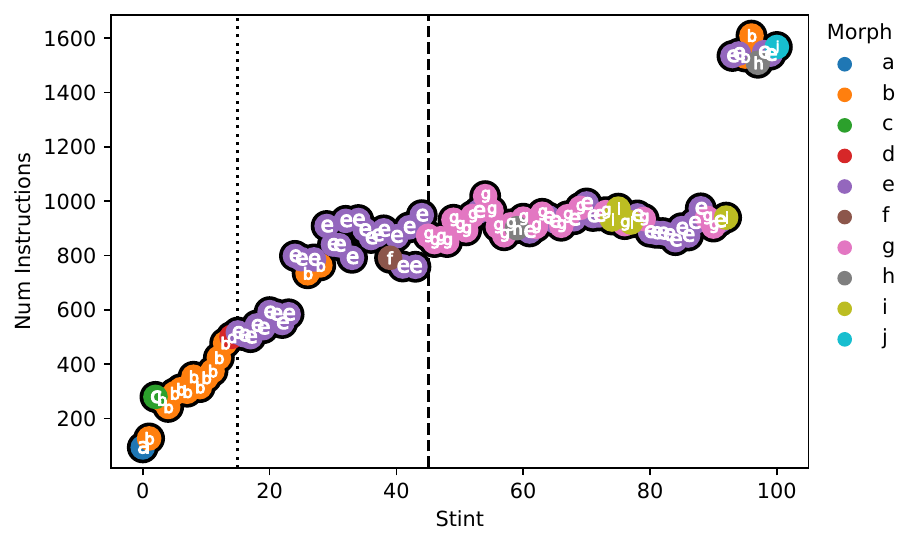}
\caption{instruction count}
\label{fig:instruction_count}
\end{subfigure}
\begin{subfigure}{0.49\textwidth}
\includegraphics[width=\linewidth]{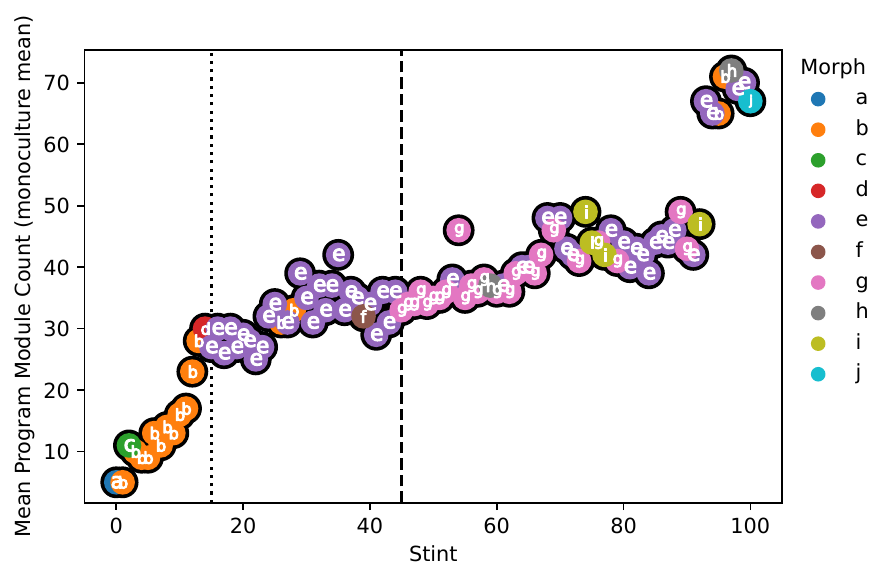}
\caption{module count}
\label{fig:module_count}
\end{subfigure}
\begin{subfigure}{0.49\textwidth}
\includegraphics[width=\linewidth]{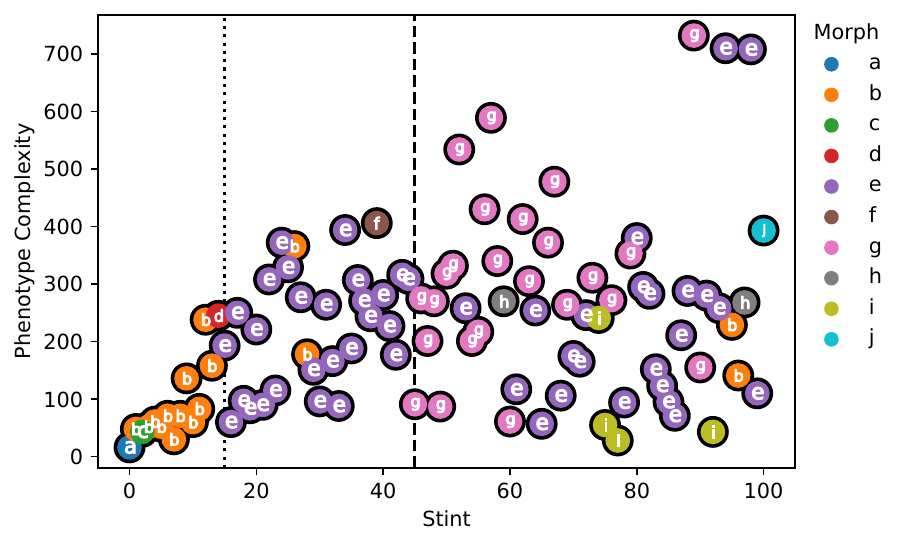}
\caption{phenotype complexity}
\label{fig:phenotype_complexity}
\end{subfigure}

\caption{
\textbf{Genome content measures.}
\footnotesize
Instruction count is the total number of instructions present in the genome.
Module count is the number of tagged linear GP modules available for activation by signals from the environment, from other agents, or from within an agent.
Phenotype complexity is the number of genome sites that contribute to phenotype, measured as number sites remaining after phenotype-neutral nopout (Section \ref{sec:phenotype_neutral_nopout}).
This measure gives a sense of the number of ``active'' instructions that influence agents' behavior.
Color coding and letters correspond to qualitative morph codes described in Table \ref{tab:morph_descriptions}.
Dotted vertical line denotes emergence of morph $e$.
Dashed vertical line denotes emergence of morph $g$.
}
\label{fig:genome_size}

\end{figure}

Figure \ref{fig:genome_size} shows evolutionary trajectories of three genome size metrics in sampled focal strain specimens.
Instruction count and module count increased from 100 and 5 to around 800 and 30, respectively, between stints 0 and 40.
Within this period at stint 24, instruction count jumped from around 600 to more than 800 and module count jumped by about 5.
This was coincident with detection in our adaptation assays of population-level sign-change mediation of adaptation by the background strain (Figure \ref{fig:fitness_gain_or_loss}).
In sampled specimen fitness assays at stint 24, we detected significant increases in fitness in the presence of the background strain but no significant change in fitness in its absence.

Between stints 40 and 90, module count gradually increased to around 40 while instruction count remained stable.
Then, at stint 93, instruction count jumped to around 1,500 and module count jumped to around 60.
This was coincident with the strong fitness differentials observed at stint 93 (Figure \ref{fig:median_fitness_differential_symlog}).

To better understand the functional effects of changes in genome size, we additionally measured the number of instructions that affected agent phenotype, shown as ``phenotype complexity'' in Figure \ref{fig:phenotype_complexity}.
This measure can be considered akin to a count of ``active'' sites.

Phenotype complexity varied greatly stint to stint.
The median value increased from nearly 0 to around 200 sites between stints 0 and 40.
Between stints 40 and 90, we observed phenotype complexity values ranging from less than 100 to more than 500.
Morph $g$ specimens appear to show particularly great variance in phenotype complexity.
The first observed morph $g$ specimen at stint 45 exhibited relatively low phenotype complexity of around 100 active sites.
The highest phenotype complexity values of around 700 were measured from three specimens of morphs $e$ and $g$ in the last ten stints.

\section{Discussion}

Across the case study lineage, we describe an evolutionary sequence of ten qualitatively distinct multicellular morphologies (Table \ref{tab:morph_descriptions}).
The emergence of some, but not all, of these morphologies coincided with an increase in fitness compared to the preceding population.
Fitness assays from the sole observed specimen of morphology $c$ are significantly deleterious in all contexts.
Similarly, the sole observed specimen of morphology $f$ exhibited an advantage only in the absence of the background strain (Figure \ref{fig:fitness_gain_or_loss}).
By contrast, the geneses of morphologies $e$ and $g$ were associated with significant fitness gain in all contexts (Figure \ref{fig:fitness_gain_or_loss}).
Likewise, the sole observed specimen of morphology $d$ exhibited a significant fitness advantage in most, but not all, tested contexts.
This latter set of novelties ($e$, $g$, and $d$) might be described as ``innovations,'' which \citet{hochberg2017innovation} define as qualitative novelty associated with an increase in fitness.
Interestingly, though, the fitness effect sizes observed from these ``innovations'' appear generally comparable in magnitude to other stint-to-stint trials with no substantial novelties (Figure \ref{fig:median_fitness_differential_symlog}).

The relationship between innovation and complexity appears to be loosely coupled.
On one hand, the origination of morphologies $d$ and $e$, respectively, coincided with a critical fitness complexity spike (25 to 43 sites) and an interface complexity spike (5 to 17 interactions).%
\footnote{Note that phylogenetic analysis suggests that morphology $e$ was not a direct descendant of morphology $d$, but that both instead descended from morphology $b$.}
The emergence of morphology $g$, by contrast, coincided with more modest changes in critical fitness complexity and interface complexity (41 to 42 sites and 12 to 15 interactions).
At the opposite extreme, consider the morph $i$ origination observed at stint 75, which exhibited a significant fitness advantage across all tested contexts.
In this case, innovation coincided with sharp decreases in both critical fitness complexity (33 to 3 sites) and interface complexity (12 to 0 interactions).

Although increases in complexity accompanied morphological novelty in several cases, we also observed substantial changes in complexity occurring outside this context.
For example, a spike in critical fitness complexity (11 to 27 sites) was observed between stints 11 and 12 --- with morphology $b$ exhibited at both timepoints.
Similarly, morphology $e$ was retained across a long, gradual increase in critical fitness complexity (31 to 46 sites) between stints 15 and 36.

Finally, we also observed notable disjointedness between alternate measures of functional complexity.
For instance, critical fitness complexity increased by 18 sites with the emergence of morph $d$, but interface complexity increased only marginally (5 to 6 interactions).
Further, compared against morph $d$, the first specimen of morph $e$ exhibited near triple the interface complexity (6 vs. 17 interactions), but 12 fewer sites of critical fitness complexity.
Consider also the increase in critical fitness complexity between stints 15 and 36 under morphology $e$ (31 to 46 sites), which is not accompanied by a clear change in interface complexity (Figures \ref{fig:interface_complexity:cardinal_interface_complexity} and \ref{fig:critical_fitness_complexity}).
These discrepancies between alternate metrics for functional complexity suggest underlying multidimensionality, and underscore well-known difficulties in attempts to describe and quantify complexity \citep{bottcher2018molecules}.

\section{Conclusion}

Complexity and novelty are not inevitable outcomes of Darwinian evolution \citep{stanley2017open}.
Instead, how and why some lineages within some model systems evolve complexity and novelty merits explanation.
In this regard, efforts to develop substrates and conditions sufficient to observe meaningful evolution of complexity and novelty play a crucial role in assessing the sufficiency of theory.
The artificial life research community has a rich track record in such work.

The case study reported here tracks a lineage over two phenotypic innovations and several-fold increases in complexity.
DISHTINY relaxes common simulation constraints \citep{goldsby2012task, goldsby2014evolutionary}, enabling broad genetic determination of multicellular life history and allowing for unconstrained cellular interactions between multicellular bodies.
As such, this case study opens a concrete window into evolutionary origins of complexity and novelty under regimes of strong biotic interactions.

Our case study suggests a loose coupling between novelty, complexity, and adaptation.
We observe instances where novelty coincides with adaptation and instances where it does not.
We observe instances where increases in complexity coincide with adaptation and where decreases in complexity coincide with adaptation.
We observe instances where innovation coincides with spikes in complexity and instances where it does not.
We even observe substantial divergences between metrics measuring different aspects of functional complexity.
For example, the specimen sampled at stint 15 had nearly triple the interface complexity of the specimen sampled at stint 14 but lower critical fitness complexity.

Loose coupling between the conceptual threads of novelty, complexity, and adaptation in this case study highlights the importance of considering these factors independently when developing theory around open-ended evolution --- inherent coupling cannot be assumed.

Our observation of significant selective effects by the background strain suggests that it may serve a crucial role in understanding the focal strain.
Future work should characterize trajectories of adaptation, novelty, and complexity in this background strain.
Additionally, success of the biotic background in fleshing out our adaptation assays suggests that complexity measures could be improved through similar incorporation of the biotic background.
It would be particularly interesting to measure the contribution of the background strain to complexity as the difference between complexity statistics with and without the biotic background.
To more systematically test the role of biotic selection on facilitating evolution of complexity, future experiments might test for differences in the rate of high-complexity evolutionary outcomes between evolution experiments with and without long-term coexistence between lineages (i.e., with diversity maintenance mechanisms enabled versus disabled).

This case study highlights the potential usefulness of toolbox-based approaches to analyzing open-ended evolution systems in which an array of analyses are performed to distinguish disparate dimensions of open-endedness \citep{dolson2019modes}.
Our findings underscore, in particular, the critical role of biotic context in such analyses.
In future work, we are interested in further extending this toolbox.
One priority will be estimating epistatic contributions to fitness without resorting to all-pairs knockouts or other even more extensive assays \citep{moreno2024methods}.
Such methodology will be crucial for systems where fitness is implicit and expensive to measure.

\section*{Author Contributions}

\textbf{Conceptualization}: M.A.M., C.A.O. \\
\textbf{Methodology}: M.A.M. \\
\textbf{Software}: M.A.M., S.R.P. \\
\textbf{Writing}: M.A.M. (lead), S.R.P. (supporting contributions) \\
\textbf{Supervision}: C.A.O.

\section*{Competing Interests}

The authors declare no competing interests.

\section*{Data Availability}

All data is available via the Open Science Framework at \url{https://osf.io/prq49} \citep{Moreno_Rodriguez_Papa_2022}.

\section*{Code Availability}

Core DISHTINY code is available under an MIT License at \url{https://github.com/mmore500/dishtiny} \citep{moreno_2025_16990564}.
Code specific to this manuscript is available under an MIT License at \url{https://github.com/mmore500/oee4} and \url{https://github.com/rodsan0/dishtiny_event_tag_phylogenetics} \citep{matthew_andres_moreno_2024_16990558,rodsan0_2025_16990704}.

\section*{Acknowledgements}

Thanks to members of the DEVOLAB, in particular Katherine Perry for help implementing DISHTINY visualizations.

M.A.M. completed substantial portions of this work while affiliated with the BEACON Center at Michigan State University.

This research was supported in part by NSF grants DEB-1655715 and DBI-0939454 as well as by Michigan State University through the computational resources provided by the Institute for Cyber-Enabled Research.
This material is based upon work supported by the National Science Foundation Graduate Research Fellowship under Grant No. DGE-1424871.
Any opinions, findings, and conclusions or recommendations expressed in this material are those of the author(s) and do not necessarily reflect the views of the National Science Foundation.

This material is based upon work supported by the Eric and Wendy Schmidt AI in Science Postdoctoral Fellowship, a Schmidt Sciences program.

This material is based upon work supported by the U.S. Department of Energy, Office of Science, Office of Advanced Scientific Computing Research (ASCR), under Award Number DE-SC0025634.
This report was prepared as an account of work sponsored by an agency of the United States Government.
Neither the United States Government nor any agency thereof, nor any of their employees, makes any warranty, express or implied, or assumes any legal liability or responsibility for the accuracy, completeness, or usefulness of any information, apparatus, product, or process disclosed, or represents that its use would not infringe privately owned rights.
Reference herein to any specific commercial product, process, or service by trade name, trademark, manufacturer, or otherwise does not necessarily constitute or imply its endorsement, recommendation, or favoring by the United States Government or any agency thereof.
The views and opinions of authors expressed herein do not necessarily state or reflect those of the United States Government or any agency thereof.

\footnotesize
\bibliographystyle{apalike-ejor}
\bibliography{bibl} 

\begin{thebibliography}{}

\bibitem[Ackley \& Small, 2014]{ackley2014indefinitely}
Ackley, D. \& Small, T. (2014).
\newblock Indefinitely scalable computing = artificial life engineering.
\newblock {\em Artificial Life 14: Proceedings of the Fourteenth International
  Conference on the Synthesis and Simulation of Living Systems}, Alife,
  606--613.
\newblock \url{https://doi.org/10.7551/978-0-262-32621-6-ch098}

\bibitem[Ackley, 2023]{ackley2023robust}
Ackley, D.~H. (2023).
\newblock A robust programmable replicator for an indefinitely scalable
  machine.
\newblock {\em The 2023 Conference on Artificial Life}, Alife 2023.
\newblock \url{https://doi.org/10.1162/isal_a_00701}

\bibitem[Adami et~al., 2000]{adami2000evolution}
Adami, C., Ofria, C., \& Collier, T.~C. (2000).
\newblock Evolution of biological complexity.
\newblock {\em Proceedings of the National Academy of Sciences}, 97(9),
  4463–4468.
\newblock \url{https://doi.org/10.1073/pnas.97.9.4463}

\bibitem[Andersson \& Hughes, 1996]{andersson1996muller}
Andersson, D.~I. \& Hughes, D. (1996).
\newblock Muller's ratchet decreases fitness of a dna-based microbe.
\newblock {\em Proceedings of the National Academy of Sciences}, 93(2),
  906--907.
\newblock \url{https://doi.org/10.1073/pnas.93.2.906}

\bibitem[Bedau et~al., 1998]{bedau1998classification}
Bedau, M.~A., Snyder, E., \& Packard, N.~H. (1998).
\newblock A classification of long-term evolutionary dynamics.
\newblock {\em Proceedings of the Sixth International Conference on Artificial
  Life}, Alife, 228--237.

\bibitem[B\"{o}ttcher, 2017]{bottcher2018molecules}
B\"{o}ttcher, T. (2017).
\newblock From molecules to life: Quantifying the complexity of chemical and
  biological systems in the universe.
\newblock {\em Journal of Molecular Evolution}, 86(1), 1--10.
\newblock \url{https://doi.org/10.1007/s00239-017-9824-6}

\bibitem[Brady et~al., 2019]{brady2019causes}
Brady, S.~P., Bolnick, D.~I., Angert, A.~L., Gonzalez, A., Barrett, R.~D.,
  Crispo, E., Derry, A.~M., Eckert, C.~G., Fraser, D.~J., Fussmann, G.~F.,
  Guichard, F., Lamy, T., McAdam, A.~G., Newman, A.~E., Paccard, A.,
  Rolshausen, G., Simons, A.~M., \& Hendry, A.~P. (2019).
\newblock Causes of maladaptation.
\newblock {\em Evolutionary Applications}, 12(7), 1229--1242.
\newblock \url{https://doi.org/10.1111/eva.12844}

\bibitem[Cock et~al., 2009]{cock2009biopython}
Cock, P. J.~A., Antao, T., Chang, J.~T., Chapman, B.~A., Cox, C.~J., Dalke, A.,
  Friedberg, I., Hamelryck, T., Kauff, F., Wilczynski, B., \& de~Hoon, M. J.~L.
  (2009).
\newblock Biopython: freely available python tools for computational molecular
  biology and bioinformatics.
\newblock {\em Bioinformatics}, 25(11), 1422--1423.
\newblock \url{https://doi.org/10.1093/bioinformatics/btp163}

\bibitem[Dolson, 2019]{dolson2019constructive}
Dolson, E. (2019).
\newblock {\em On the Constructive Power of Ecology in Open-ended Evolving
  Systems}.
\newblock Michigan State University.

\bibitem[Dolson et~al., 2019]{dolson2019modes}
Dolson, E.~L., Vostinar, A.~E., Wiser, M.~J., \& Ofria, C. (2019).
\newblock The modes toolbox: Measurements of open-ended dynamics in evolving
  systems.
\newblock {\em Artificial Life}, 25(1), 50--73.
\newblock \url{https://doi.org/10.1162/artl_a_00280}

\bibitem[Downing, 2015]{downing2015intelligence}
Downing, K.~L. (2015).
\newblock {\em Intelligence Emerging: Adaptivity and Search in Evolving Neural
  Systems}.
\newblock The MIT Press.
\newblock \url{https://doi.org/10.7551/mitpress/9898.001.0001}

\bibitem[Goldsby et~al., 2012]{goldsby2012task}
Goldsby, H.~J., Dornhaus, A., Kerr, B., \& Ofria, C. (2012).
\newblock Task-switching costs promote the evolution of division of labor and
  shifts in individuality.
\newblock {\em Proceedings of the National Academy of Sciences}, 109(34),
  13686--13691.
\newblock \url{https://doi.org/10.1073/pnas.1202233109}

\bibitem[Goldsby et~al., 2014]{goldsby2014evolutionary}
Goldsby, H.~J., Knoester, D.~B., Ofria, C., \& Kerr, B. (2014).
\newblock The evolutionary origin of somatic cells under the dirty work
  hypothesis.
\newblock {\em PLoS Biology}, 12(5), e1001858.
\newblock \url{https://doi.org/10.1371/journal.pbio.1001858}

\bibitem[Harris et~al., 2020]{harris2020array}
Harris, C.~R., Millman, K.~J., van~der Walt, S.~J., Gommers, R., Virtanen, P.,
  Cournapeau, D., Wieser, E., Taylor, J., Berg, S., Smith, N.~J., Kern, R.,
  Picus, M., Hoyer, S., van Kerkwijk, M.~H., Brett, M., Haldane, A., del
  R{\'{i}}o, J.~F., Wiebe, M., Peterson, P., G{\'{e}}rard-Marchant, P.,
  Sheppard, K., Reddy, T., Weckesser, W., Abbasi, H., Gohlke, C., \& Oliphant,
  T.~E. (2020).
\newblock Array programming with {NumPy}.
\newblock {\em Nature}, 585(7825), 357--362.
\newblock \url{https://doi.org/10.1038/s41586-020-2649-2}

\bibitem[Hochberg et~al., 2017]{hochberg2017innovation}
Hochberg, M.~E., Marquet, P.~A., Boyd, R., \& Wagner, A. (2017).
\newblock Innovation: an emerging focus from cells to societies.
\newblock {\em Philosophical Transactions of the Royal Society B: Biological
  Sciences}, 372(1735), 20160414.
\newblock \url{https://doi.org/10.1098/rstb.2016.0414}

\bibitem[Huizinga et~al., 2018]{huizinga2018emergence}
Huizinga, J., Stanley, K.~O., \& Clune, J. (2018).
\newblock The emergence of canalization and evolvability in an open-ended,
  interactive evolutionary system.
\newblock {\em Artificial Life}, 24(3), 157--181.
\newblock \url{https://doi.org/10.1162/artl_a_00263}

\bibitem[Hunter, 2007]{hunter2007matplotlib}
Hunter, J.~D. (2007).
\newblock Matplotlib: A 2d graphics environment.
\newblock {\em Computing in Science \& Engineering}, 9(3), 90--95.
\newblock \url{https://doi.org/10.1109/mcse.2007.55}

\bibitem[Kirschner \& Gerhart, 1998]{Kirschner8420}
Kirschner, M. \& Gerhart, J. (1998).
\newblock Evolvability.
\newblock {\em Proceedings of the National Academy of Sciences}, 95(15),
  8420--8427.
\newblock \url{https://doi.org/10.1073/pnas.95.15.8420}

\bibitem[Lalejini et~al., 2019]{lalejini2019data}
Lalejini, A., Dolson, E., Bohm, C., Ferguson, A.~J., Parsons, D.~P., Rainford,
  P.~F., Richmond, P., \& Ofria, C. (2019).
\newblock Data standards for artificial life software.
\newblock {\em The 2019 Conference on Artificial Life}, Alife 2019.
\newblock \url{https://doi.org/10.1162/isal_a_00213}

\bibitem[Lalejini et~al., 2021]{lalejini2021tag}
Lalejini, A., Moreno, M.~A., \& Ofria, C. (2021).
\newblock Tag-based regulation of modules in genetic programming improves
  context-dependent problem solving.
\newblock {\em Genetic Programming and Evolvable Machines}, 22(3), 325--355.
\newblock \url{https://doi.org/10.1007/s10710-021-09406-8}

\bibitem[Lalejini \& Ofria, 2018]{lalejini2018evolving}
Lalejini, A. \& Ofria, C. (2018).
\newblock Evolving event-driven programs with signalgp.
\newblock {\em Proceedings of the Genetic and Evolutionary Computation
  Conference}, Gecco '18, 1135--1142.
\newblock \url{https://doi.org/10.1145/3205455.3205523}

\bibitem[Lauterbach, 2021]{lauterbach2021path}
Lauterbach, G. (2021).
\newblock The path to successful wafer-scale integration: The cerebras story.
\newblock {\em IEEE Micro}, 41(6), 52--57.
\newblock \url{https://doi.org/10.1109/mm.2021.3112025}

\bibitem[Lehman, 2012]{lehman2012evolution}
Lehman, J. (2012).
\newblock {\em Evolution through the Search for Novelty}.
\newblock University of Central Florida.

\bibitem[Lehman \& Stanley, 2011]{lehman2011abandoning}
Lehman, J. \& Stanley, K.~O. (2011).
\newblock Abandoning objectives: Evolution through the search for novelty
  alone.
\newblock {\em Evolutionary Computation}, 19(2), 189--223.
\newblock \url{https://doi.org/10.1162/evco_a_00025}

\bibitem[Lehman \& Stanley, 2012]{lehman2012beyond}
Lehman, J. \& Stanley, K.~O. (2012).
\newblock Beyond open-endedness: Quantifying impressiveness.
\newblock {\em Artificial Life Conference Proceedings 12}, 75--82.

\bibitem[Lehman \& Stanley, 2013]{lehman2013evolvability}
Lehman, J. \& Stanley, K.~O. (2013).
\newblock Evolvability is inevitable: Increasing evolvability without the
  pressure to adapt.
\newblock {\em PloS One}, 8(4), e62186.

\bibitem[Lynch, 2007]{Lynch8597}
Lynch, M. (2007).
\newblock The frailty of adaptive hypotheses for the origins of organismal
  complexity.
\newblock {\em Proceedings of the National Academy of Sciences}, 104(suppl 1),
  8597--8604.
\newblock \url{https://doi.org/10.1073/pnas.0702207104}

\bibitem[Martin \& Roques, 2016]{martin2016nonstationary}
Martin, G. \& Roques, L. (2016).
\newblock The nonstationary dynamics of fitness distributions: Asexual model
  with epistasis and standing variation.
\newblock {\em Genetics}, 204(4), 1541--1558.
\newblock \url{https://doi.org/10.1534/genetics.116.187385}

\bibitem[Moreno, 2023]{moreno2023teeplot}
Moreno, M.~A. (2023).
\newblock {\em mmore500/teeplot}.
\newblock \url{https://doi.org/10.5281/zenodo.10440670}

\bibitem[Moreno, 2024]{moreno2024methods}
Moreno, M.~A. (2024).
\newblock Methods to estimate cryptic sequence complexity.
\newblock {\em The 2024 Conference on Artificial Life}, Alife 2024.
\newblock \url{https://doi.org/10.1162/isal_a_00776}

\bibitem[Moreno et~al., 2022]{moreno2022hereditary}
Moreno, M.~A., Dolson, E., \& Ofria, C. (2022).
\newblock Hereditary stratigraphy: Genome annotations to enable phylogenetic
  inference over distributed populations.
\newblock {\em The 2022 Conference on Artificial Life}, Alife 2022, 64.
\newblock \url{https://doi.org/10.1162/isal_a_00550}.
\newblock 64

\bibitem[Moreno et~al., 2023]{moreno2023matchmaker}
Moreno, M.~A., Lalejini, A., \& Ofria, C. (2023).
\newblock Matchmaker, matchmaker, make me a match: geometric, variational, and
  evolutionary implications of criteria for tag affinity.
\newblock {\em Genetic Programming and Evolvable Machines}, 24(1), 4.
\newblock \url{https://doi.org/10.1007/s10710-023-09448-0}

\bibitem[Moreno \& Ofria, 2019]{moreno2019toward}
Moreno, M.~A. \& Ofria, C. (2019).
\newblock Toward open-ended fraternal transitions in individuality.
\newblock {\em Artificial Life}, 25(2), 117--133.

\bibitem[Moreno \& Ofria, 2022]{moreno2021exploring}
Moreno, M.~A. \& Ofria, C. (2022).
\newblock Exploring evolved multicellular life histories in a open-ended
  digital evolution system.
\newblock {\em Frontiers in Ecology and Evolution}, 10.
\newblock \url{https://doi.org/10.3389/fevo.2022.750837}

\bibitem[Moreno et~al., 2021a]{moreno2021conduit}
Moreno, M.~A., Papa, S.~R., \& Ofria, C. (2021a).
\newblock Conduit: a c++ library for best-effort high performance computing.
\newblock {\em Proceedings of the Genetic and Evolutionary Computation
  Conference Companion}, Gecco '21, 1795--1800.
\newblock \url{https://doi.org/10.1145/3449726.3463205}

\bibitem[Moreno \& Rodriguez~Papa, 2022]{Moreno_Rodriguez_Papa_2022}
Moreno, M.~A. \& Rodriguez~Papa, S. (2022).
\newblock {\em Dishtiny case study}.
\newblock \url{https://doi.org/10.17605/osf.io/prq49}

\bibitem[Moreno et~al., 2024a]{moreno2024analysis}
Moreno, M.~A., {Rodriguez Papa}, S., \& Dolson, E. (2024a).
\newblock {\em Analysis of phylogeny tracking algorithms for serial and
  multiprocess applications}.
\newblock \url{https://doi.org/10.48550/arXiv.2403.00246}

\bibitem[Moreno et~al., 2021b]{moreno2021signalgp}
Moreno, M.~A., {Rodriguez Papa}, S., Lalejini, A., \& Ofria, C. (2021b).
\newblock {\em Signalgp-lite: Event driven genetic programming library for
  large-scale artificial life applications}.
\newblock \url{https://doi.org/10.48550/arxiv.2108.00382}

\bibitem[Moreno et~al., 2025]{moreno_2025_16990564}
Moreno, M.~A., Rodriguez~Papa, S., \& Ofria, C. (2025).
\newblock {\em Toward open-ended fraternal transitions in individuality}.
\newblock \url{https://doi.org/10.5281/zenodo.16990564}

\bibitem[Moreno \& rodsan0, 2024]{matthew_andres_moreno_2024_16990558}
Moreno, M.~A. \& rodsan0 (2024).
\newblock {\em mmore500/oee4: As uploaded to arxiv}.
\newblock \url{https://doi.org/10.5281/zenodo.16990557}

\bibitem[Moreno et~al., 2024b]{moreno2024trackable}
Moreno, M.~A., Yang, C., Dolson, E., \& Zaman, L. (2024b).
\newblock Trackable agent-based evolution models at wafer scale.
\newblock {\em The 2024 Conference on Artificial Life}, Alife 2024.
\newblock \url{https://doi.org/10.1162/isal_a_00830}

\bibitem[Nguyen et~al., 2015]{nguyen2015}
Nguyen, A.~M., Yosinski, J., \& Clune, J. (2015).
\newblock Innovation engines: Automated creativity and improved stochastic
  optimization via deep learning.
\newblock {\em Proceedings of the 2015 Annual Conference on Genetic and
  Evolutionary Computation}, Gecco '15, 959--966.
\newblock \url{https://doi.org/10.1145/2739480.2754703}

\bibitem[Packard et~al., 2019]{packard2019overview}
Packard, N., Bedau, M.~A., Channon, A., Ikegami, T., Rasmussen, S., Stanley,
  K.~O., \& Taylor, T. (2019).
\newblock An overview of open-ended evolution: Editorial introduction to the
  open-ended evolution ii special issue.
\newblock {\em Artificial Life}, 25(2), 93--103.

\bibitem[pandas developers, 2020]{reback2020pandas}
pandas developers (2020).
\newblock pandas-dev/pandas: Pandas.
\newblock {\em Zenodo}.
\newblock \url{https://doi.org/10.5281/zenodo.3509134}

\bibitem[rodsan0 \& Moreno, 2025]{rodsan0_2025_16990704}
rodsan0 \& Moreno, M.~A. (2025).
\newblock {\em mmore500/dishtiny\_event\_tag\_phylogenetics: prq49}.
\newblock \url{https://doi.org/10.5281/zenodo.16990704}

\bibitem[Seabold \& Perktold, 2010]{seabold2010statsmodels}
Seabold, S. \& Perktold, J. (2010).
\newblock Statsmodels: Econometric and statistical modeling with python.
\newblock {\em Proceedings of the 9th Python in Science Conference}, SciPy,
  92--96.
\newblock \url{https://doi.org/10.25080/majora-92bf1922-011}

\bibitem[Soros \& Stanley, 2014]{soros2014identifying}
Soros, L. \& Stanley, K. (2014).
\newblock Identifying necessary conditions for open-ended evolution through the
  artificial life world of chromaria.
\newblock {\em Artificial Life 14: Proceedings of the Fourteenth International
  Conference on the Synthesis and Simulation of Living Systems}, Alife,
  793--800.
\newblock \url{https://doi.org/10.1162/978-0-262-32621-6-ch128}

\bibitem[Stanley, 2019]{stanley2019open}
Stanley, K.~O. (2019).
\newblock Why open-endedness matters.
\newblock {\em Artificial Life}, 25(3), 232--235.
\newblock \url{https://doi.org/10.1162/artl_a_00294}

\bibitem[Stanley et~al., 2017]{stanley2017open}
Stanley, K.~O., Lehman, J., \& Soros, L. (2017).
\newblock Open-endedness: The last grand challenge you've never heard of.
\newblock {\em Radar / AI \& ML}.

\bibitem[Taylor et~al., 2016]{taylor2016open}
Taylor, T., Bedau, M., Channon, A., Ackley, D., Banzhaf, W., Beslon, G.,
  Dolson, E., Froese, T., Hickinbotham, S., Ikegami, T., McMullin, B., Packard,
  N., Rasmussen, S., Virgo, N., Agmon, E., Clark, E., McGregor, S., Ofria, C.,
  Ropella, G., Spector, L., Stanley, K.~O., Stanton, A., Timperley, C.,
  Vostinar, A., \& Wiser, M. (2016).
\newblock Open-ended evolution: Perspectives from the oee workshop in york.
\newblock {\em Artificial Life}, 22(3), 408--423.
\newblock \url{https://doi.org/10.1162/artl_a_00210}

\bibitem[Virtanen et~al., 2020]{2020SciPy-NMeth}
Virtanen, P., Gommers, R., Oliphant, T.~E., Haberland, M., Reddy, T.,
  Cournapeau, D., Burovski, E., Peterson, P., Weckesser, W., Bright, J., {van
  der Walt}, S.~J., Brett, M., Wilson, J., Millman, K.~J., Mayorov, N., Nelson,
  A. R.~J., Jones, E., Kern, R., Larson, E., Carey, C.~J., Polat, {\.I}., Feng,
  Y., Moore, E.~W., {VanderPlas}, J., Laxalde, D., Perktold, J., Cimrman, R.,
  Henriksen, I., Quintero, E.~A., Harris, C.~R., Archibald, A.~M., Ribeiro,
  A.~H., Pedregosa, F., {van Mulbregt}, P., \& {SciPy 1.0 Contributors} (2020).
\newblock {{SciPy} 1.0: Fundamental Algorithms for Scientific Computing in
  Python}.
\newblock {\em Nature Methods}, 17, 261--272.
\newblock \url{https://doi.org/10.1038/s41592-019-0686-2}

\bibitem[Vostinar et~al., 2024]{vostinar2024empirical}
Vostinar, A., Lalejini, A., Ofria, C., Dolson, E., \& Moreno, M.~A. (2024).
\newblock Empirical: A scientific software library for research, education, and
  public engagement.
\newblock {\em Journal of Open Source Software}, 9(98), 6617.
\newblock \url{https://doi.org/10.21105/joss.06617}

\bibitem[Waskom, 2021]{waskom2021seaborn}
Waskom, M.~L. (2021).
\newblock seaborn: statistical data visualization.
\newblock {\em Journal of Open Source Software}, 6(60), 3021.
\newblock \url{https://doi.org/10.21105/joss.03021}

\bibitem[{W}es {M}c{K}inney, 2010]{mckinney-proc-scipy-2010}
{W}es {M}c{K}inney (2010).
\newblock {D}ata {S}tructures for {S}tatistical {C}omputing in {P}ython.
\newblock {\em {P}roceedings of the 9th {P}ython in {S}cience {C}onference},
  56--61.
\newblock \url{https://doi.org/10.25080/Majora-92bf1922-00a}

\end{thebibliography}

\clearpage
\newpage

\appendix

\clearpage
\onecolumn

\vspace*{\fill}
{\Huge
Supplement for ``Case Study of Novelty, Complexity, and Adaptation in a Multicellular System'' by Matthew Andres Moreno, Santiago Rodriguez Papa, and Charles Ofria
in OEE4: The Fourth Workshop on Open-Ended Evolution}
\vspace*{\fill}

\clearpage

\setcounter{secnumdepth}{2}


\providecommand{\dissertationexclude}[1]{%
\ifdefined\DISSERTATION
\else
#1
\fi
}

\dissertationexclude{\clearpage}
\dissertationexclude{\onecolumn}


\providecommand{\dissertationelse}[2]{%
\ifdefined\DISSERTATION
#1
\else
#2
\fi
}

\newcommand{\tinyurl}[1]{\dissertationelse{\fontsize{8}{8}\selectfont}{\tiny}\url{#1}}

\newcommand{\morphcode}[1]{ \raisebox{-0.8em}{\color[HTML]{FFFFFF} \large \textbf{#1} }}

\begin{longtable}[]{ccp{2.5cm} p{2.5cm} p{2.5cm} p{2.5cm} p{2.5cm}}
\caption{Qualitative morph categorization of representative specimens sampled across evolutionary stints.
Video links provide a time-lapse animation of each specimen's morphology when grown in monoculture.
Web viewer URLs link to an in-browser simulation of each genome or population.
The human-readable genome link will download the corresponding genome as an annotated JSON file.} \label{tab:morph_by_stint} \\
\multicolumn{1}{l}{\textbf{Stint}}               & \textbf{Morph} & \textbf{Monoculture Video Link} & \textbf{Monoculture Web Viewer Link} & \textbf{Population Web Viewer Link} & \textbf{Human-readable Genome Link} \\
\endfirsthead
\caption*{\tablename{} \thetable{} (cont'd)} \\
\multicolumn{1}{l}{\textbf{Stint}}               & \textbf{Morph} & \textbf{Monoculture Video Link} & \textbf{Monoculture Web Viewer Link} & \textbf{Population Web Viewer Link} & \textbf{Human-readable Genome Link} \\
\endhead
0 & \cellcolor[HTML]{4C72B0}{\morphcode{a}} & \tinyurl{https://hopth.ru/21/b=prq49+s=16005+t=0+v=video+w=specimen} & \tinyurl{https://hopth.ru/21/b=prq49+s=16005+t=0+v=simulation+w=specimen} & \tinyurl{https://hopth.ru/21/b=prq49+s=16005+t=0+v=simulation+w=population} & \tinyurl{https://hopth.ru/21/b=prq49+s=16005+t=0+v=text+w=specimen} & \\
1 & \cellcolor[HTML]{DD8452}{\morphcode{b} } & \tinyurl{https://hopth.ru/21/b=prq49+s=16005+t=1+v=video+w=specimen} & \tinyurl{https://hopth.ru/21/b=prq49+s=16005+t=1+v=simulation+w=specimen} & \tinyurl{https://hopth.ru/21/b=prq49+s=16005+t=1+v=simulation+w=population} & \tinyurl{https://hopth.ru/21/b=prq49+s=16005+t=1+v=text+w=specimen} \\
2 & \cellcolor[HTML]{55A868}{\morphcode{c} } & \tinyurl{https://hopth.ru/21/b=prq49+s=16005+t=2+v=video+w=specimen} & \tinyurl{https://hopth.ru/21/b=prq49+s=16005+t=2+v=simulation+w=specimen} & \tinyurl{https://hopth.ru/21/b=prq49+s=16005+t=2+v=simulation+w=population} & \tinyurl{https://hopth.ru/21/b=prq49+s=16005+t=2+v=text+w=specimen} \\
3 & \cellcolor[HTML]{DD8452}{\morphcode{b} } & \tinyurl{https://hopth.ru/21/b=prq49+s=16005+t=3+v=video+w=specimen} & \tinyurl{https://hopth.ru/21/b=prq49+s=16005+t=3+v=simulation+w=specimen} & \tinyurl{https://hopth.ru/21/b=prq49+s=16005+t=3+v=simulation+w=population} & \tinyurl{https://hopth.ru/21/b=prq49+s=16005+t=3+v=text+w=specimen} \\
4 & \cellcolor[HTML]{DD8452}{\morphcode{b} } & \tinyurl{https://hopth.ru/21/b=prq49+s=16005+t=4+v=video+w=specimen} & \tinyurl{https://hopth.ru/21/b=prq49+s=16005+t=4+v=simulation+w=specimen} & \tinyurl{https://hopth.ru/21/b=prq49+s=16005+t=4+v=simulation+w=population} & \tinyurl{https://hopth.ru/21/b=prq49+s=16005+t=4+v=text+w=specimen} \\
5 & \cellcolor[HTML]{DD8452}{\morphcode{b} } & \tinyurl{https://hopth.ru/21/b=prq49+s=16005+t=5+v=video+w=specimen} & \tinyurl{https://hopth.ru/21/b=prq49+s=16005+t=5+v=simulation+w=specimen} & \tinyurl{https://hopth.ru/21/b=prq49+s=16005+t=5+v=simulation+w=population} & \tinyurl{https://hopth.ru/21/b=prq49+s=16005+t=5+v=text+w=specimen} \\
6 & \cellcolor[HTML]{DD8452}{\morphcode{b} } & \tinyurl{https://hopth.ru/21/b=prq49+s=16005+t=6+v=video+w=specimen} & \tinyurl{https://hopth.ru/21/b=prq49+s=16005+t=6+v=simulation+w=specimen} & \tinyurl{https://hopth.ru/21/b=prq49+s=16005+t=6+v=simulation+w=population} & \tinyurl{https://hopth.ru/21/b=prq49+s=16005+t=6+v=text+w=specimen} \\
7 & \cellcolor[HTML]{DD8452}{\morphcode{b} } & \tinyurl{https://hopth.ru/21/b=prq49+s=16005+t=7+v=video+w=specimen} & \tinyurl{https://hopth.ru/21/b=prq49+s=16005+t=7+v=simulation+w=specimen} & \tinyurl{https://hopth.ru/21/b=prq49+s=16005+t=7+v=simulation+w=population} & \tinyurl{https://hopth.ru/21/b=prq49+s=16005+t=7+v=text+w=specimen} \\
8 & \cellcolor[HTML]{DD8452}{\morphcode{b} } & \tinyurl{https://hopth.ru/21/b=prq49+s=16005+t=8+v=video+w=specimen} & \tinyurl{https://hopth.ru/21/b=prq49+s=16005+t=8+v=simulation+w=specimen} & \tinyurl{https://hopth.ru/21/b=prq49+s=16005+t=8+v=simulation+w=population} & \tinyurl{https://hopth.ru/21/b=prq49+s=16005+t=8+v=text+w=specimen} \\
9 & \cellcolor[HTML]{DD8452}{\morphcode{b} } & \tinyurl{https://hopth.ru/21/b=prq49+s=16005+t=9+v=video+w=specimen} & \tinyurl{https://hopth.ru/21/b=prq49+s=16005+t=9+v=simulation+w=specimen} & \tinyurl{https://hopth.ru/21/b=prq49+s=16005+t=9+v=simulation+w=population} & \tinyurl{https://hopth.ru/21/b=prq49+s=16005+t=9+v=text+w=specimen} \\
10 & \cellcolor[HTML]{DD8452}{\morphcode{b} } & \tinyurl{https://hopth.ru/21/b=prq49+s=16005+t=10+v=video+w=specimen} & \tinyurl{https://hopth.ru/21/b=prq49+s=16005+t=10+v=simulation+w=specimen} & \tinyurl{https://hopth.ru/21/b=prq49+s=16005+t=10+v=simulation+w=population} & \tinyurl{https://hopth.ru/21/b=prq49+s=16005+t=10+v=text+w=specimen} \\
11 & \cellcolor[HTML]{DD8452}{\morphcode{b} } & \tinyurl{https://hopth.ru/21/b=prq49+s=16005+t=11+v=video+w=specimen} & \tinyurl{https://hopth.ru/21/b=prq49+s=16005+t=11+v=simulation+w=specimen} & \tinyurl{https://hopth.ru/21/b=prq49+s=16005+t=11+v=simulation+w=population} & \tinyurl{https://hopth.ru/21/b=prq49+s=16005+t=11+v=text+w=specimen} \\
12 & \cellcolor[HTML]{DD8452}{\morphcode{b} } & \tinyurl{https://hopth.ru/21/b=prq49+s=16005+t=12+v=video+w=specimen} & \tinyurl{https://hopth.ru/21/b=prq49+s=16005+t=12+v=simulation+w=specimen} & \tinyurl{https://hopth.ru/21/b=prq49+s=16005+t=12+v=simulation+w=population} & \tinyurl{https://hopth.ru/21/b=prq49+s=16005+t=12+v=text+w=specimen} \\
13 & \cellcolor[HTML]{DD8452}{\morphcode{b} } & \tinyurl{https://hopth.ru/21/b=prq49+s=16005+t=13+v=video+w=specimen} & \tinyurl{https://hopth.ru/21/b=prq49+s=16005+t=13+v=simulation+w=specimen} & \tinyurl{https://hopth.ru/21/b=prq49+s=16005+t=13+v=simulation+w=population} & \tinyurl{https://hopth.ru/21/b=prq49+s=16005+t=13+v=text+w=specimen} \\
14 & \cellcolor[HTML]{C44E52}{\morphcode{d} } & \tinyurl{https://hopth.ru/21/b=prq49+s=16005+t=14+v=video+w=specimen} & \tinyurl{https://hopth.ru/21/b=prq49+s=16005+t=14+v=simulation+w=specimen} & \tinyurl{https://hopth.ru/21/b=prq49+s=16005+t=14+v=simulation+w=population} & \tinyurl{https://hopth.ru/21/b=prq49+s=16005+t=14+v=text+w=specimen} \\
15 & \cellcolor[HTML]{8172B3}{\morphcode{e} } & \tinyurl{https://hopth.ru/21/b=prq49+s=16005+t=15+v=video+w=specimen} & \tinyurl{https://hopth.ru/21/b=prq49+s=16005+t=15+v=simulation+w=specimen} & \tinyurl{https://hopth.ru/21/b=prq49+s=16005+t=15+v=simulation+w=population} & \tinyurl{https://hopth.ru/21/b=prq49+s=16005+t=15+v=text+w=specimen} \\
16 & \cellcolor[HTML]{8172B3}{\morphcode{e} } & \tinyurl{https://hopth.ru/21/b=prq49+s=16005+t=16+v=video+w=specimen} & \tinyurl{https://hopth.ru/21/b=prq49+s=16005+t=16+v=simulation+w=specimen} & \tinyurl{https://hopth.ru/21/b=prq49+s=16005+t=16+v=simulation+w=population} & \tinyurl{https://hopth.ru/21/b=prq49+s=16005+t=16+v=text+w=specimen} \\
17 & \cellcolor[HTML]{8172B3}{\morphcode{e} } & \tinyurl{https://hopth.ru/21/b=prq49+s=16005+t=17+v=video+w=specimen} & \tinyurl{https://hopth.ru/21/b=prq49+s=16005+t=17+v=simulation+w=specimen} & \tinyurl{https://hopth.ru/21/b=prq49+s=16005+t=17+v=simulation+w=population} & \tinyurl{https://hopth.ru/21/b=prq49+s=16005+t=17+v=text+w=specimen} \\
18 & \cellcolor[HTML]{8172B3}{\morphcode{e} } & \tinyurl{https://hopth.ru/21/b=prq49+s=16005+t=18+v=video+w=specimen} & \tinyurl{https://hopth.ru/21/b=prq49+s=16005+t=18+v=simulation+w=specimen} & \tinyurl{https://hopth.ru/21/b=prq49+s=16005+t=18+v=simulation+w=population} & \tinyurl{https://hopth.ru/21/b=prq49+s=16005+t=18+v=text+w=specimen} \\
19 & \cellcolor[HTML]{8172B3}{\morphcode{e} } & \tinyurl{https://hopth.ru/21/b=prq49+s=16005+t=19+v=video+w=specimen} & \tinyurl{https://hopth.ru/21/b=prq49+s=16005+t=19+v=simulation+w=specimen} & \tinyurl{https://hopth.ru/21/b=prq49+s=16005+t=19+v=simulation+w=population} & \tinyurl{https://hopth.ru/21/b=prq49+s=16005+t=19+v=text+w=specimen} \\
20 & \cellcolor[HTML]{8172B3}{\morphcode{e} } & \tinyurl{https://hopth.ru/21/b=prq49+s=16005+t=20+v=video+w=specimen} & \tinyurl{https://hopth.ru/21/b=prq49+s=16005+t=20+v=simulation+w=specimen} & \tinyurl{https://hopth.ru/21/b=prq49+s=16005+t=20+v=simulation+w=population} & \tinyurl{https://hopth.ru/21/b=prq49+s=16005+t=20+v=text+w=specimen} \\
21 & \cellcolor[HTML]{8172B3}{\morphcode{e} } & \tinyurl{https://hopth.ru/21/b=prq49+s=16005+t=21+v=video+w=specimen} & \tinyurl{https://hopth.ru/21/b=prq49+s=16005+t=21+v=simulation+w=specimen} & \tinyurl{https://hopth.ru/21/b=prq49+s=16005+t=21+v=simulation+w=population} & \tinyurl{https://hopth.ru/21/b=prq49+s=16005+t=21+v=text+w=specimen} \\
22 & \cellcolor[HTML]{8172B3}{\morphcode{e} } & \tinyurl{https://hopth.ru/21/b=prq49+s=16005+t=22+v=video+w=specimen} & \tinyurl{https://hopth.ru/21/b=prq49+s=16005+t=22+v=simulation+w=specimen} & \tinyurl{https://hopth.ru/21/b=prq49+s=16005+t=22+v=simulation+w=population} & \tinyurl{https://hopth.ru/21/b=prq49+s=16005+t=22+v=text+w=specimen} \\
23 & \cellcolor[HTML]{8172B3}{\morphcode{e} } & \tinyurl{https://hopth.ru/21/b=prq49+s=16005+t=23+v=video+w=specimen} & \tinyurl{https://hopth.ru/21/b=prq49+s=16005+t=23+v=simulation+w=specimen} & \tinyurl{https://hopth.ru/21/b=prq49+s=16005+t=23+v=simulation+w=population} & \tinyurl{https://hopth.ru/21/b=prq49+s=16005+t=23+v=text+w=specimen} \\
24 & \cellcolor[HTML]{8172B3}{\morphcode{e} } & \tinyurl{https://hopth.ru/21/b=prq49+s=16005+t=24+v=video+w=specimen} & \tinyurl{https://hopth.ru/21/b=prq49+s=16005+t=24+v=simulation+w=specimen} & \tinyurl{https://hopth.ru/21/b=prq49+s=16005+t=24+v=simulation+w=population} & \tinyurl{https://hopth.ru/21/b=prq49+s=16005+t=24+v=text+w=specimen} \\
25 & \cellcolor[HTML]{8172B3}{\morphcode{e} } & \tinyurl{https://hopth.ru/21/b=prq49+s=16005+t=25+v=video+w=specimen} & \tinyurl{https://hopth.ru/21/b=prq49+s=16005+t=25+v=simulation+w=specimen} & \tinyurl{https://hopth.ru/21/b=prq49+s=16005+t=25+v=simulation+w=population} & \tinyurl{https://hopth.ru/21/b=prq49+s=16005+t=25+v=text+w=specimen} \\
26 & \cellcolor[HTML]{DD8452}{\morphcode{b} } & \tinyurl{https://hopth.ru/21/b=prq49+s=16005+t=26+v=video+w=specimen} & \tinyurl{https://hopth.ru/21/b=prq49+s=16005+t=26+v=simulation+w=specimen} & \tinyurl{https://hopth.ru/21/b=prq49+s=16005+t=26+v=simulation+w=population} & \tinyurl{https://hopth.ru/21/b=prq49+s=16005+t=26+v=text+w=specimen} \\
27 & \cellcolor[HTML]{8172B3}{\morphcode{e} } & \tinyurl{https://hopth.ru/21/b=prq49+s=16005+t=27+v=video+w=specimen} & \tinyurl{https://hopth.ru/21/b=prq49+s=16005+t=27+v=simulation+w=specimen} & \tinyurl{https://hopth.ru/21/b=prq49+s=16005+t=27+v=simulation+w=population} & \tinyurl{https://hopth.ru/21/b=prq49+s=16005+t=27+v=text+w=specimen} \\
28 & \cellcolor[HTML]{DD8452}{\morphcode{b} } & \tinyurl{https://hopth.ru/21/b=prq49+s=16005+t=28+v=video+w=specimen} & \tinyurl{https://hopth.ru/21/b=prq49+s=16005+t=28+v=simulation+w=specimen} & \tinyurl{https://hopth.ru/21/b=prq49+s=16005+t=28+v=simulation+w=population} & \tinyurl{https://hopth.ru/21/b=prq49+s=16005+t=28+v=text+w=specimen} \\
29 & \cellcolor[HTML]{8172B3}{\morphcode{e} } & \tinyurl{https://hopth.ru/21/b=prq49+s=16005+t=29+v=video+w=specimen} & \tinyurl{https://hopth.ru/21/b=prq49+s=16005+t=29+v=simulation+w=specimen} & \tinyurl{https://hopth.ru/21/b=prq49+s=16005+t=29+v=simulation+w=population} & \tinyurl{https://hopth.ru/21/b=prq49+s=16005+t=29+v=text+w=specimen} \\
30 & \cellcolor[HTML]{8172B3}{\morphcode{e} } & \tinyurl{https://hopth.ru/21/b=prq49+s=16005+t=30+v=video+w=specimen} & \tinyurl{https://hopth.ru/21/b=prq49+s=16005+t=30+v=simulation+w=specimen} & \tinyurl{https://hopth.ru/21/b=prq49+s=16005+t=30+v=simulation+w=population} & \tinyurl{https://hopth.ru/21/b=prq49+s=16005+t=30+v=text+w=specimen} \\
31 & \cellcolor[HTML]{8172B3}{\morphcode{e} } & \tinyurl{https://hopth.ru/21/b=prq49+s=16005+t=31+v=video+w=specimen} & \tinyurl{https://hopth.ru/21/b=prq49+s=16005+t=31+v=simulation+w=specimen} & \tinyurl{https://hopth.ru/21/b=prq49+s=16005+t=31+v=simulation+w=population} & \tinyurl{https://hopth.ru/21/b=prq49+s=16005+t=31+v=text+w=specimen} \\
32 & \cellcolor[HTML]{8172B3}{\morphcode{e} } & \tinyurl{https://hopth.ru/21/b=prq49+s=16005+t=32+v=video+w=specimen} & \tinyurl{https://hopth.ru/21/b=prq49+s=16005+t=32+v=simulation+w=specimen} & \tinyurl{https://hopth.ru/21/b=prq49+s=16005+t=32+v=simulation+w=population} & \tinyurl{https://hopth.ru/21/b=prq49+s=16005+t=32+v=text+w=specimen} \\
33 & \cellcolor[HTML]{8172B3}{\morphcode{e} } & \tinyurl{https://hopth.ru/21/b=prq49+s=16005+t=33+v=video+w=specimen} & \tinyurl{https://hopth.ru/21/b=prq49+s=16005+t=33+v=simulation+w=specimen} & \tinyurl{https://hopth.ru/21/b=prq49+s=16005+t=33+v=simulation+w=population} & \tinyurl{https://hopth.ru/21/b=prq49+s=16005+t=33+v=text+w=specimen} \\
34 & \cellcolor[HTML]{8172B3}{\morphcode{e} } & \tinyurl{https://hopth.ru/21/b=prq49+s=16005+t=34+v=video+w=specimen} & \tinyurl{https://hopth.ru/21/b=prq49+s=16005+t=34+v=simulation+w=specimen} & \tinyurl{https://hopth.ru/21/b=prq49+s=16005+t=34+v=simulation+w=population} & \tinyurl{https://hopth.ru/21/b=prq49+s=16005+t=34+v=text+w=specimen} \\
35 & \cellcolor[HTML]{8172B3}{\morphcode{e} } & \tinyurl{https://hopth.ru/21/b=prq49+s=16005+t=35+v=video+w=specimen} & \tinyurl{https://hopth.ru/21/b=prq49+s=16005+t=35+v=simulation+w=specimen} & \tinyurl{https://hopth.ru/21/b=prq49+s=16005+t=35+v=simulation+w=population} & \tinyurl{https://hopth.ru/21/b=prq49+s=16005+t=35+v=text+w=specimen} \\
36 & \cellcolor[HTML]{8172B3}{\morphcode{e} } & \tinyurl{https://hopth.ru/21/b=prq49+s=16005+t=36+v=video+w=specimen} & \tinyurl{https://hopth.ru/21/b=prq49+s=16005+t=36+v=simulation+w=specimen} & \tinyurl{https://hopth.ru/21/b=prq49+s=16005+t=36+v=simulation+w=population} & \tinyurl{https://hopth.ru/21/b=prq49+s=16005+t=36+v=text+w=specimen} \\
37 & \cellcolor[HTML]{8172B3}{\morphcode{e} } & \tinyurl{https://hopth.ru/21/b=prq49+s=16005+t=37+v=video+w=specimen} & \tinyurl{https://hopth.ru/21/b=prq49+s=16005+t=37+v=simulation+w=specimen} & \tinyurl{https://hopth.ru/21/b=prq49+s=16005+t=37+v=simulation+w=population} & \tinyurl{https://hopth.ru/21/b=prq49+s=16005+t=37+v=text+w=specimen} \\
38 & \cellcolor[HTML]{8172B3}{\morphcode{e} } & \tinyurl{https://hopth.ru/21/b=prq49+s=16005+t=38+v=video+w=specimen} & \tinyurl{https://hopth.ru/21/b=prq49+s=16005+t=38+v=simulation+w=specimen} & \tinyurl{https://hopth.ru/21/b=prq49+s=16005+t=38+v=simulation+w=population} & \tinyurl{https://hopth.ru/21/b=prq49+s=16005+t=38+v=text+w=specimen} \\
39 & \cellcolor[HTML]{937860}{\morphcode{f} } & \tinyurl{https://hopth.ru/21/b=prq49+s=16005+t=39+v=video+w=specimen} & \tinyurl{https://hopth.ru/21/b=prq49+s=16005+t=39+v=simulation+w=specimen} & \tinyurl{https://hopth.ru/21/b=prq49+s=16005+t=39+v=simulation+w=population} & \tinyurl{https://hopth.ru/21/b=prq49+s=16005+t=39+v=text+w=specimen} \\
40 & \cellcolor[HTML]{8172B3}{\morphcode{e} } & \tinyurl{https://hopth.ru/21/b=prq49+s=16005+t=40+v=video+w=specimen} & \tinyurl{https://hopth.ru/21/b=prq49+s=16005+t=40+v=simulation+w=specimen} & \tinyurl{https://hopth.ru/21/b=prq49+s=16005+t=40+v=simulation+w=population} & \tinyurl{https://hopth.ru/21/b=prq49+s=16005+t=40+v=text+w=specimen} \\
41 & \cellcolor[HTML]{8172B3}{\morphcode{e} } & \tinyurl{https://hopth.ru/21/b=prq49+s=16005+t=41+v=video+w=specimen} & \tinyurl{https://hopth.ru/21/b=prq49+s=16005+t=41+v=simulation+w=specimen} & \tinyurl{https://hopth.ru/21/b=prq49+s=16005+t=41+v=simulation+w=population} & \tinyurl{https://hopth.ru/21/b=prq49+s=16005+t=41+v=text+w=specimen} \\
42 & \cellcolor[HTML]{8172B3}{\morphcode{e} } & \tinyurl{https://hopth.ru/21/b=prq49+s=16005+t=42+v=video+w=specimen} & \tinyurl{https://hopth.ru/21/b=prq49+s=16005+t=42+v=simulation+w=specimen} & \tinyurl{https://hopth.ru/21/b=prq49+s=16005+t=42+v=simulation+w=population} & \tinyurl{https://hopth.ru/21/b=prq49+s=16005+t=42+v=text+w=specimen} \\
43 & \cellcolor[HTML]{8172B3}{\morphcode{e} } & \tinyurl{https://hopth.ru/21/b=prq49+s=16005+t=43+v=video+w=specimen} & \tinyurl{https://hopth.ru/21/b=prq49+s=16005+t=43+v=simulation+w=specimen} & \tinyurl{https://hopth.ru/21/b=prq49+s=16005+t=43+v=simulation+w=population} & \tinyurl{https://hopth.ru/21/b=prq49+s=16005+t=43+v=text+w=specimen} \\
44 & \cellcolor[HTML]{8172B3}{\morphcode{e} } & \tinyurl{https://hopth.ru/21/b=prq49+s=16005+t=44+v=video+w=specimen} & \tinyurl{https://hopth.ru/21/b=prq49+s=16005+t=44+v=simulation+w=specimen} & \tinyurl{https://hopth.ru/21/b=prq49+s=16005+t=44+v=simulation+w=population} & \tinyurl{https://hopth.ru/21/b=prq49+s=16005+t=44+v=text+w=specimen} \\
45 & \cellcolor[HTML]{DA8BC3}{\morphcode{g} } & \tinyurl{https://hopth.ru/21/b=prq49+s=16005+t=45+v=video+w=specimen} & \tinyurl{https://hopth.ru/21/b=prq49+s=16005+t=45+v=simulation+w=specimen} & \tinyurl{https://hopth.ru/21/b=prq49+s=16005+t=45+v=simulation+w=population} & \tinyurl{https://hopth.ru/21/b=prq49+s=16005+t=45+v=text+w=specimen} \\
46 & \cellcolor[HTML]{DA8BC3}{\morphcode{g} } & \tinyurl{https://hopth.ru/21/b=prq49+s=16005+t=46+v=video+w=specimen} & \tinyurl{https://hopth.ru/21/b=prq49+s=16005+t=46+v=simulation+w=specimen} & \tinyurl{https://hopth.ru/21/b=prq49+s=16005+t=46+v=simulation+w=population} & \tinyurl{https://hopth.ru/21/b=prq49+s=16005+t=46+v=text+w=specimen} \\
47 & \cellcolor[HTML]{DA8BC3}{\morphcode{g} } & \tinyurl{https://hopth.ru/21/b=prq49+s=16005+t=47+v=video+w=specimen} & \tinyurl{https://hopth.ru/21/b=prq49+s=16005+t=47+v=simulation+w=specimen} & \tinyurl{https://hopth.ru/21/b=prq49+s=16005+t=47+v=simulation+w=population} & \tinyurl{https://hopth.ru/21/b=prq49+s=16005+t=47+v=text+w=specimen} \\
48 & \cellcolor[HTML]{DA8BC3}{\morphcode{g} } & \tinyurl{https://hopth.ru/21/b=prq49+s=16005+t=48+v=video+w=specimen} & \tinyurl{https://hopth.ru/21/b=prq49+s=16005+t=48+v=simulation+w=specimen} & \tinyurl{https://hopth.ru/21/b=prq49+s=16005+t=48+v=simulation+w=population} & \tinyurl{https://hopth.ru/21/b=prq49+s=16005+t=48+v=text+w=specimen} \\
49 & \cellcolor[HTML]{DA8BC3}{\morphcode{g} } & \tinyurl{https://hopth.ru/21/b=prq49+s=16005+t=49+v=video+w=specimen} & \tinyurl{https://hopth.ru/21/b=prq49+s=16005+t=49+v=simulation+w=specimen} & \tinyurl{https://hopth.ru/21/b=prq49+s=16005+t=49+v=simulation+w=population} & \tinyurl{https://hopth.ru/21/b=prq49+s=16005+t=49+v=text+w=specimen} \\
50 & \cellcolor[HTML]{DA8BC3}{\morphcode{g} } & \tinyurl{https://hopth.ru/21/b=prq49+s=16005+t=50+v=video+w=specimen} & \tinyurl{https://hopth.ru/21/b=prq49+s=16005+t=50+v=simulation+w=specimen} & \tinyurl{https://hopth.ru/21/b=prq49+s=16005+t=50+v=simulation+w=population} & \tinyurl{https://hopth.ru/21/b=prq49+s=16005+t=50+v=text+w=specimen} \\
51 & \cellcolor[HTML]{DA8BC3}{\morphcode{g} } & \tinyurl{https://hopth.ru/21/b=prq49+s=16005+t=51+v=video+w=specimen} & \tinyurl{https://hopth.ru/21/b=prq49+s=16005+t=51+v=simulation+w=specimen} & \tinyurl{https://hopth.ru/21/b=prq49+s=16005+t=51+v=simulation+w=population} & \tinyurl{https://hopth.ru/21/b=prq49+s=16005+t=51+v=text+w=specimen} \\
52 & \cellcolor[HTML]{DA8BC3}{\morphcode{g} } & \tinyurl{https://hopth.ru/21/b=prq49+s=16005+t=52+v=video+w=specimen} & \tinyurl{https://hopth.ru/21/b=prq49+s=16005+t=52+v=simulation+w=specimen} & \tinyurl{https://hopth.ru/21/b=prq49+s=16005+t=52+v=simulation+w=population} & \tinyurl{https://hopth.ru/21/b=prq49+s=16005+t=52+v=text+w=specimen} \\
53 & \cellcolor[HTML]{8172B3}{\morphcode{e} } & \tinyurl{https://hopth.ru/21/b=prq49+s=16005+t=53+v=video+w=specimen} & \tinyurl{https://hopth.ru/21/b=prq49+s=16005+t=53+v=simulation+w=specimen} & \tinyurl{https://hopth.ru/21/b=prq49+s=16005+t=53+v=simulation+w=population} & \tinyurl{https://hopth.ru/21/b=prq49+s=16005+t=53+v=text+w=specimen} \\
54 & \cellcolor[HTML]{DA8BC3}{\morphcode{g} } & \tinyurl{https://hopth.ru/21/b=prq49+s=16005+t=54+v=video+w=specimen} & \tinyurl{https://hopth.ru/21/b=prq49+s=16005+t=54+v=simulation+w=specimen} & \tinyurl{https://hopth.ru/21/b=prq49+s=16005+t=54+v=simulation+w=population} & \tinyurl{https://hopth.ru/21/b=prq49+s=16005+t=54+v=text+w=specimen} \\
55 & \cellcolor[HTML]{DA8BC3}{\morphcode{g} } & \tinyurl{https://hopth.ru/21/b=prq49+s=16005+t=55+v=video+w=specimen} & \tinyurl{https://hopth.ru/21/b=prq49+s=16005+t=55+v=simulation+w=specimen} & \tinyurl{https://hopth.ru/21/b=prq49+s=16005+t=55+v=simulation+w=population} & \tinyurl{https://hopth.ru/21/b=prq49+s=16005+t=55+v=text+w=specimen} \\
56 & \cellcolor[HTML]{DA8BC3}{\morphcode{g} } & \tinyurl{https://hopth.ru/21/b=prq49+s=16005+t=56+v=video+w=specimen} & \tinyurl{https://hopth.ru/21/b=prq49+s=16005+t=56+v=simulation+w=specimen} & \tinyurl{https://hopth.ru/21/b=prq49+s=16005+t=56+v=simulation+w=population} & \tinyurl{https://hopth.ru/21/b=prq49+s=16005+t=56+v=text+w=specimen} \\
57 & \cellcolor[HTML]{DA8BC3}{\morphcode{g} } & \tinyurl{https://hopth.ru/21/b=prq49+s=16005+t=57+v=video+w=specimen} & \tinyurl{https://hopth.ru/21/b=prq49+s=16005+t=57+v=simulation+w=specimen} & \tinyurl{https://hopth.ru/21/b=prq49+s=16005+t=57+v=simulation+w=population} & \tinyurl{https://hopth.ru/21/b=prq49+s=16005+t=57+v=text+w=specimen} \\
58 & \cellcolor[HTML]{DA8BC3}{\morphcode{g} } & \tinyurl{https://hopth.ru/21/b=prq49+s=16005+t=58+v=video+w=specimen} & \tinyurl{https://hopth.ru/21/b=prq49+s=16005+t=58+v=simulation+w=specimen} & \tinyurl{https://hopth.ru/21/b=prq49+s=16005+t=58+v=simulation+w=population} & \tinyurl{https://hopth.ru/21/b=prq49+s=16005+t=58+v=text+w=specimen} \\
59 & \cellcolor[HTML]{8C8C8C}{\morphcode{h} } & \tinyurl{https://hopth.ru/21/b=prq49+s=16005+t=59+v=video+w=specimen} & \tinyurl{https://hopth.ru/21/b=prq49+s=16005+t=59+v=simulation+w=specimen} & \tinyurl{https://hopth.ru/21/b=prq49+s=16005+t=59+v=simulation+w=population} & \tinyurl{https://hopth.ru/21/b=prq49+s=16005+t=59+v=text+w=specimen} \\
60 & \cellcolor[HTML]{DA8BC3}{\morphcode{g} } & \tinyurl{https://hopth.ru/21/b=prq49+s=16005+t=60+v=video+w=specimen} & \tinyurl{https://hopth.ru/21/b=prq49+s=16005+t=60+v=simulation+w=specimen} & \tinyurl{https://hopth.ru/21/b=prq49+s=16005+t=60+v=simulation+w=population} & \tinyurl{https://hopth.ru/21/b=prq49+s=16005+t=60+v=text+w=specimen} \\
61 & \cellcolor[HTML]{8172B3}{\morphcode{e} } & \tinyurl{https://hopth.ru/21/b=prq49+s=16005+t=61+v=video+w=specimen} & \tinyurl{https://hopth.ru/21/b=prq49+s=16005+t=61+v=simulation+w=specimen} & \tinyurl{https://hopth.ru/21/b=prq49+s=16005+t=61+v=simulation+w=population} & \tinyurl{https://hopth.ru/21/b=prq49+s=16005+t=61+v=text+w=specimen} \\
62 & \cellcolor[HTML]{DA8BC3}{\morphcode{g} } & \tinyurl{https://hopth.ru/21/b=prq49+s=16005+t=62+v=video+w=specimen} & \tinyurl{https://hopth.ru/21/b=prq49+s=16005+t=62+v=simulation+w=specimen} & \tinyurl{https://hopth.ru/21/b=prq49+s=16005+t=62+v=simulation+w=population} & \tinyurl{https://hopth.ru/21/b=prq49+s=16005+t=62+v=text+w=specimen} \\
63 & \cellcolor[HTML]{DA8BC3}{\morphcode{g} } & \tinyurl{https://hopth.ru/21/b=prq49+s=16005+t=63+v=video+w=specimen} & \tinyurl{https://hopth.ru/21/b=prq49+s=16005+t=63+v=simulation+w=specimen} & \tinyurl{https://hopth.ru/21/b=prq49+s=16005+t=63+v=simulation+w=population} & \tinyurl{https://hopth.ru/21/b=prq49+s=16005+t=63+v=text+w=specimen} \\
64 & \cellcolor[HTML]{8172B3}{\morphcode{e} } & \tinyurl{https://hopth.ru/21/b=prq49+s=16005+t=64+v=video+w=specimen} & \tinyurl{https://hopth.ru/21/b=prq49+s=16005+t=64+v=simulation+w=specimen} & \tinyurl{https://hopth.ru/21/b=prq49+s=16005+t=64+v=simulation+w=population} & \tinyurl{https://hopth.ru/21/b=prq49+s=16005+t=64+v=text+w=specimen} \\
65 & \cellcolor[HTML]{8172B3}{\morphcode{e} } & \tinyurl{https://hopth.ru/21/b=prq49+s=16005+t=65+v=video+w=specimen} & \tinyurl{https://hopth.ru/21/b=prq49+s=16005+t=65+v=simulation+w=specimen} & \tinyurl{https://hopth.ru/21/b=prq49+s=16005+t=65+v=simulation+w=population} & \tinyurl{https://hopth.ru/21/b=prq49+s=16005+t=65+v=text+w=specimen} \\
66 & \cellcolor[HTML]{DA8BC3}{\morphcode{g} } & \tinyurl{https://hopth.ru/21/b=prq49+s=16005+t=66+v=video+w=specimen} & \tinyurl{https://hopth.ru/21/b=prq49+s=16005+t=66+v=simulation+w=specimen} & \tinyurl{https://hopth.ru/21/b=prq49+s=16005+t=66+v=simulation+w=population} & \tinyurl{https://hopth.ru/21/b=prq49+s=16005+t=66+v=text+w=specimen} \\
67 & \cellcolor[HTML]{DA8BC3}{\morphcode{g} } & \tinyurl{https://hopth.ru/21/b=prq49+s=16005+t=67+v=video+w=specimen} & \tinyurl{https://hopth.ru/21/b=prq49+s=16005+t=67+v=simulation+w=specimen} & \tinyurl{https://hopth.ru/21/b=prq49+s=16005+t=67+v=simulation+w=population} & \tinyurl{https://hopth.ru/21/b=prq49+s=16005+t=67+v=text+w=specimen} \\
68 & \cellcolor[HTML]{8172B3}{\morphcode{e} } & \tinyurl{https://hopth.ru/21/b=prq49+s=16005+t=68+v=video+w=specimen} & \tinyurl{https://hopth.ru/21/b=prq49+s=16005+t=68+v=simulation+w=specimen} & \tinyurl{https://hopth.ru/21/b=prq49+s=16005+t=68+v=simulation+w=population} & \tinyurl{https://hopth.ru/21/b=prq49+s=16005+t=68+v=text+w=specimen} \\
69 & \cellcolor[HTML]{DA8BC3}{\morphcode{g} } & \tinyurl{https://hopth.ru/21/b=prq49+s=16005+t=69+v=video+w=specimen} & \tinyurl{https://hopth.ru/21/b=prq49+s=16005+t=69+v=simulation+w=specimen} & \tinyurl{https://hopth.ru/21/b=prq49+s=16005+t=69+v=simulation+w=population} & \tinyurl{https://hopth.ru/21/b=prq49+s=16005+t=69+v=text+w=specimen} \\
70 & \cellcolor[HTML]{8172B3}{\morphcode{e} } & \tinyurl{https://hopth.ru/21/b=prq49+s=16005+t=70+v=video+w=specimen} & \tinyurl{https://hopth.ru/21/b=prq49+s=16005+t=70+v=simulation+w=specimen} & \tinyurl{https://hopth.ru/21/b=prq49+s=16005+t=70+v=simulation+w=population} & \tinyurl{https://hopth.ru/21/b=prq49+s=16005+t=70+v=text+w=specimen} \\
71 & \cellcolor[HTML]{8172B3}{\morphcode{e} } & \tinyurl{https://hopth.ru/21/b=prq49+s=16005+t=71+v=video+w=specimen} & \tinyurl{https://hopth.ru/21/b=prq49+s=16005+t=71+v=simulation+w=specimen} & \tinyurl{https://hopth.ru/21/b=prq49+s=16005+t=71+v=simulation+w=population} & \tinyurl{https://hopth.ru/21/b=prq49+s=16005+t=71+v=text+w=specimen} \\
72 & \cellcolor[HTML]{8172B3}{\morphcode{e} } & \tinyurl{https://hopth.ru/21/b=prq49+s=16005+t=72+v=video+w=specimen} & \tinyurl{https://hopth.ru/21/b=prq49+s=16005+t=72+v=simulation+w=specimen} & \tinyurl{https://hopth.ru/21/b=prq49+s=16005+t=72+v=simulation+w=population} & \tinyurl{https://hopth.ru/21/b=prq49+s=16005+t=72+v=text+w=specimen} \\
73 & \cellcolor[HTML]{DA8BC3}{\morphcode{g} } & \tinyurl{https://hopth.ru/21/b=prq49+s=16005+t=73+v=video+w=specimen} & \tinyurl{https://hopth.ru/21/b=prq49+s=16005+t=73+v=simulation+w=specimen} & \tinyurl{https://hopth.ru/21/b=prq49+s=16005+t=73+v=simulation+w=population} & \tinyurl{https://hopth.ru/21/b=prq49+s=16005+t=73+v=text+w=specimen} \\
74 & \cellcolor[HTML]{CCB974}{\morphcode{i} } & \tinyurl{https://hopth.ru/21/b=prq49+s=16005+t=74+v=video+w=specimen} & \tinyurl{https://hopth.ru/21/b=prq49+s=16005+t=74+v=simulation+w=specimen} & \tinyurl{https://hopth.ru/21/b=prq49+s=16005+t=74+v=simulation+w=population} & \tinyurl{https://hopth.ru/21/b=prq49+s=16005+t=74+v=text+w=specimen} \\
75 & \cellcolor[HTML]{CCB974}{\morphcode{i} } & \tinyurl{https://hopth.ru/21/b=prq49+s=16005+t=75+v=video+w=specimen} & \tinyurl{https://hopth.ru/21/b=prq49+s=16005+t=75+v=simulation+w=specimen} & \tinyurl{https://hopth.ru/21/b=prq49+s=16005+t=75+v=simulation+w=population} & \tinyurl{https://hopth.ru/21/b=prq49+s=16005+t=75+v=text+w=specimen} \\
76 & \cellcolor[HTML]{DA8BC3}{\morphcode{g} } & \tinyurl{https://hopth.ru/21/b=prq49+s=16005+t=76+v=video+w=specimen} & \tinyurl{https://hopth.ru/21/b=prq49+s=16005+t=76+v=simulation+w=specimen} & \tinyurl{https://hopth.ru/21/b=prq49+s=16005+t=76+v=simulation+w=population} & \tinyurl{https://hopth.ru/21/b=prq49+s=16005+t=76+v=text+w=specimen} \\
77 & \cellcolor[HTML]{CCB974}{\morphcode{i} } & \tinyurl{https://hopth.ru/21/b=prq49+s=16005+t=77+v=video+w=specimen} & \tinyurl{https://hopth.ru/21/b=prq49+s=16005+t=77+v=simulation+w=specimen} & \tinyurl{https://hopth.ru/21/b=prq49+s=16005+t=77+v=simulation+w=population} & \tinyurl{https://hopth.ru/21/b=prq49+s=16005+t=77+v=text+w=specimen} \\
78 & \cellcolor[HTML]{8172B3}{\morphcode{e} } & \tinyurl{https://hopth.ru/21/b=prq49+s=16005+t=78+v=video+w=specimen} & \tinyurl{https://hopth.ru/21/b=prq49+s=16005+t=78+v=simulation+w=specimen} & \tinyurl{https://hopth.ru/21/b=prq49+s=16005+t=78+v=simulation+w=population} & \tinyurl{https://hopth.ru/21/b=prq49+s=16005+t=78+v=text+w=specimen} \\
79 & \cellcolor[HTML]{DA8BC3}{\morphcode{g} } & \tinyurl{https://hopth.ru/21/b=prq49+s=16005+t=79+v=video+w=specimen} & \tinyurl{https://hopth.ru/21/b=prq49+s=16005+t=79+v=simulation+w=specimen} & \tinyurl{https://hopth.ru/21/b=prq49+s=16005+t=79+v=simulation+w=population} & \tinyurl{https://hopth.ru/21/b=prq49+s=16005+t=79+v=text+w=specimen} \\
80 & \cellcolor[HTML]{8172B3}{\morphcode{e} } & \tinyurl{https://hopth.ru/21/b=prq49+s=16005+t=80+v=video+w=specimen} & \tinyurl{https://hopth.ru/21/b=prq49+s=16005+t=80+v=simulation+w=specimen} & \tinyurl{https://hopth.ru/21/b=prq49+s=16005+t=80+v=simulation+w=population} & \tinyurl{https://hopth.ru/21/b=prq49+s=16005+t=80+v=text+w=specimen} \\
81 & \cellcolor[HTML]{8172B3}{\morphcode{e} } & \tinyurl{https://hopth.ru/21/b=prq49+s=16005+t=81+v=video+w=specimen} & \tinyurl{https://hopth.ru/21/b=prq49+s=16005+t=81+v=simulation+w=specimen} & \tinyurl{https://hopth.ru/21/b=prq49+s=16005+t=81+v=simulation+w=population} & \tinyurl{https://hopth.ru/21/b=prq49+s=16005+t=81+v=text+w=specimen} \\
82 & \cellcolor[HTML]{8172B3}{\morphcode{e} } & \tinyurl{https://hopth.ru/21/b=prq49+s=16005+t=82+v=video+w=specimen} & \tinyurl{https://hopth.ru/21/b=prq49+s=16005+t=82+v=simulation+w=specimen} & \tinyurl{https://hopth.ru/21/b=prq49+s=16005+t=82+v=simulation+w=population} & \tinyurl{https://hopth.ru/21/b=prq49+s=16005+t=82+v=text+w=specimen} \\
83 & \cellcolor[HTML]{8172B3}{\morphcode{e} } & \tinyurl{https://hopth.ru/21/b=prq49+s=16005+t=83+v=video+w=specimen} & \tinyurl{https://hopth.ru/21/b=prq49+s=16005+t=83+v=simulation+w=specimen} & \tinyurl{https://hopth.ru/21/b=prq49+s=16005+t=83+v=simulation+w=population} & \tinyurl{https://hopth.ru/21/b=prq49+s=16005+t=83+v=text+w=specimen} \\
84 & \cellcolor[HTML]{8172B3}{\morphcode{e} } & \tinyurl{https://hopth.ru/21/b=prq49+s=16005+t=84+v=video+w=specimen} & \tinyurl{https://hopth.ru/21/b=prq49+s=16005+t=84+v=simulation+w=specimen} & \tinyurl{https://hopth.ru/21/b=prq49+s=16005+t=84+v=simulation+w=population} & \tinyurl{https://hopth.ru/21/b=prq49+s=16005+t=84+v=text+w=specimen} \\
85 & \cellcolor[HTML]{8172B3}{\morphcode{e} } & \tinyurl{https://hopth.ru/21/b=prq49+s=16005+t=85+v=video+w=specimen} & \tinyurl{https://hopth.ru/21/b=prq49+s=16005+t=85+v=simulation+w=specimen} & \tinyurl{https://hopth.ru/21/b=prq49+s=16005+t=85+v=simulation+w=population} & \tinyurl{https://hopth.ru/21/b=prq49+s=16005+t=85+v=text+w=specimen} \\
86 & \cellcolor[HTML]{8172B3}{\morphcode{e} } & \tinyurl{https://hopth.ru/21/b=prq49+s=16005+t=86+v=video+w=specimen} & \tinyurl{https://hopth.ru/21/b=prq49+s=16005+t=86+v=simulation+w=specimen} & \tinyurl{https://hopth.ru/21/b=prq49+s=16005+t=86+v=simulation+w=population} & \tinyurl{https://hopth.ru/21/b=prq49+s=16005+t=86+v=text+w=specimen} \\
87 & \cellcolor[HTML]{8172B3}{\morphcode{e} } & \tinyurl{https://hopth.ru/21/b=prq49+s=16005+t=87+v=video+w=specimen} & \tinyurl{https://hopth.ru/21/b=prq49+s=16005+t=87+v=simulation+w=specimen} & \tinyurl{https://hopth.ru/21/b=prq49+s=16005+t=87+v=simulation+w=population} & \tinyurl{https://hopth.ru/21/b=prq49+s=16005+t=87+v=text+w=specimen} \\
88 & \cellcolor[HTML]{8172B3}{\morphcode{e} } & \tinyurl{https://hopth.ru/21/b=prq49+s=16005+t=88+v=video+w=specimen} & \tinyurl{https://hopth.ru/21/b=prq49+s=16005+t=88+v=simulation+w=specimen} & \tinyurl{https://hopth.ru/21/b=prq49+s=16005+t=88+v=simulation+w=population} & \tinyurl{https://hopth.ru/21/b=prq49+s=16005+t=88+v=text+w=specimen} \\
89 & \cellcolor[HTML]{DA8BC3}{\morphcode{g} } & \tinyurl{https://hopth.ru/21/b=prq49+s=16005+t=89+v=video+w=specimen} & \tinyurl{https://hopth.ru/21/b=prq49+s=16005+t=89+v=simulation+w=specimen} & \tinyurl{https://hopth.ru/21/b=prq49+s=16005+t=89+v=simulation+w=population} & \tinyurl{https://hopth.ru/21/b=prq49+s=16005+t=89+v=text+w=specimen} \\
90 & \cellcolor[HTML]{DA8BC3}{\morphcode{g} } & \tinyurl{https://hopth.ru/21/b=prq49+s=16005+t=90+v=video+w=specimen} & \tinyurl{https://hopth.ru/21/b=prq49+s=16005+t=90+v=simulation+w=specimen} & \tinyurl{https://hopth.ru/21/b=prq49+s=16005+t=90+v=simulation+w=population} & \tinyurl{https://hopth.ru/21/b=prq49+s=16005+t=90+v=text+w=specimen} \\
91 & \cellcolor[HTML]{8172B3}{\morphcode{e} } & \tinyurl{https://hopth.ru/21/b=prq49+s=16005+t=91+v=video+w=specimen} & \tinyurl{https://hopth.ru/21/b=prq49+s=16005+t=91+v=simulation+w=specimen} & \tinyurl{https://hopth.ru/21/b=prq49+s=16005+t=91+v=simulation+w=population} & \tinyurl{https://hopth.ru/21/b=prq49+s=16005+t=91+v=text+w=specimen} \\
92 & \cellcolor[HTML]{CCB974}{\morphcode{i} } & \tinyurl{https://hopth.ru/21/b=prq49+s=16005+t=92+v=video+w=specimen} & \tinyurl{https://hopth.ru/21/b=prq49+s=16005+t=92+v=simulation+w=specimen} & \tinyurl{https://hopth.ru/21/b=prq49+s=16005+t=92+v=simulation+w=population} & \tinyurl{https://hopth.ru/21/b=prq49+s=16005+t=92+v=text+w=specimen} \\
93 & \cellcolor[HTML]{8172B3}{\morphcode{e} } & \tinyurl{https://hopth.ru/21/b=prq49+s=16005+t=93+v=video+w=specimen} & \tinyurl{https://hopth.ru/21/b=prq49+s=16005+t=93+v=simulation+w=specimen} & \tinyurl{https://hopth.ru/21/b=prq49+s=16005+t=93+v=simulation+w=population} & \tinyurl{https://hopth.ru/21/b=prq49+s=16005+t=93+v=text+w=specimen} \\
94 & \cellcolor[HTML]{8172B3}{\morphcode{e} } & \tinyurl{https://hopth.ru/21/b=prq49+s=16005+t=94+v=video+w=specimen} & \tinyurl{https://hopth.ru/21/b=prq49+s=16005+t=94+v=simulation+w=specimen} & \tinyurl{https://hopth.ru/21/b=prq49+s=16005+t=94+v=simulation+w=population} & \tinyurl{https://hopth.ru/21/b=prq49+s=16005+t=94+v=text+w=specimen} \\
95 & \cellcolor[HTML]{DD8452}{\morphcode{b} } & \tinyurl{https://hopth.ru/21/b=prq49+s=16005+t=95+v=video+w=specimen} & \tinyurl{https://hopth.ru/21/b=prq49+s=16005+t=95+v=simulation+w=specimen} & \tinyurl{https://hopth.ru/21/b=prq49+s=16005+t=95+v=simulation+w=population} & \tinyurl{https://hopth.ru/21/b=prq49+s=16005+t=95+v=text+w=specimen} \\
96 & \cellcolor[HTML]{DD8452}{\morphcode{b} } & \tinyurl{https://hopth.ru/21/b=prq49+s=16005+t=96+v=video+w=specimen} & \tinyurl{https://hopth.ru/21/b=prq49+s=16005+t=96+v=simulation+w=specimen} & \tinyurl{https://hopth.ru/21/b=prq49+s=16005+t=96+v=simulation+w=population} & \tinyurl{https://hopth.ru/21/b=prq49+s=16005+t=96+v=text+w=specimen} \\
97 & \cellcolor[HTML]{8C8C8C}{\morphcode{h} } & \tinyurl{https://hopth.ru/21/b=prq49+s=16005+t=97+v=video+w=specimen} & \tinyurl{https://hopth.ru/21/b=prq49+s=16005+t=97+v=simulation+w=specimen} & \tinyurl{https://hopth.ru/21/b=prq49+s=16005+t=97+v=simulation+w=population} & \tinyurl{https://hopth.ru/21/b=prq49+s=16005+t=97+v=text+w=specimen} \\
98 & \cellcolor[HTML]{8172B3}{\morphcode{e} } & \tinyurl{https://hopth.ru/21/b=prq49+s=16005+t=98+v=video+w=specimen} & \tinyurl{https://hopth.ru/21/b=prq49+s=16005+t=98+v=simulation+w=specimen} & \tinyurl{https://hopth.ru/21/b=prq49+s=16005+t=98+v=simulation+w=population} & \tinyurl{https://hopth.ru/21/b=prq49+s=16005+t=98+v=text+w=specimen} \\
99 & \cellcolor[HTML]{8172B3}{\morphcode{e} } & \tinyurl{https://hopth.ru/21/b=prq49+s=16005+t=99+v=video+w=specimen} & \tinyurl{https://hopth.ru/21/b=prq49+s=16005+t=99+v=simulation+w=specimen} & \tinyurl{https://hopth.ru/21/b=prq49+s=16005+t=99+v=simulation+w=population} & \tinyurl{https://hopth.ru/21/b=prq49+s=16005+t=99+v=text+w=specimen} \\
100 & \cellcolor[HTML]{64B5CD}{\morphcode{j} } & \tinyurl{https://hopth.ru/21/b=prq49+s=16005+t=100+v=video+w=specimen} & \tinyurl{https://hopth.ru/21/b=prq49+s=16005+t=100+v=simulation+w=specimen} & \tinyurl{https://hopth.ru/21/b=prq49+s=16005+t=100+v=simulation+w=population} & \tinyurl{https://hopth.ru/21/b=prq49+s=16005+t=100+v=text+w=specimen} \\
\end{longtable}

\dissertationexclude{\clearpage}
\dissertationexclude{\twocolumn}

\begin{figure*}

\includegraphics[width=\linewidth]{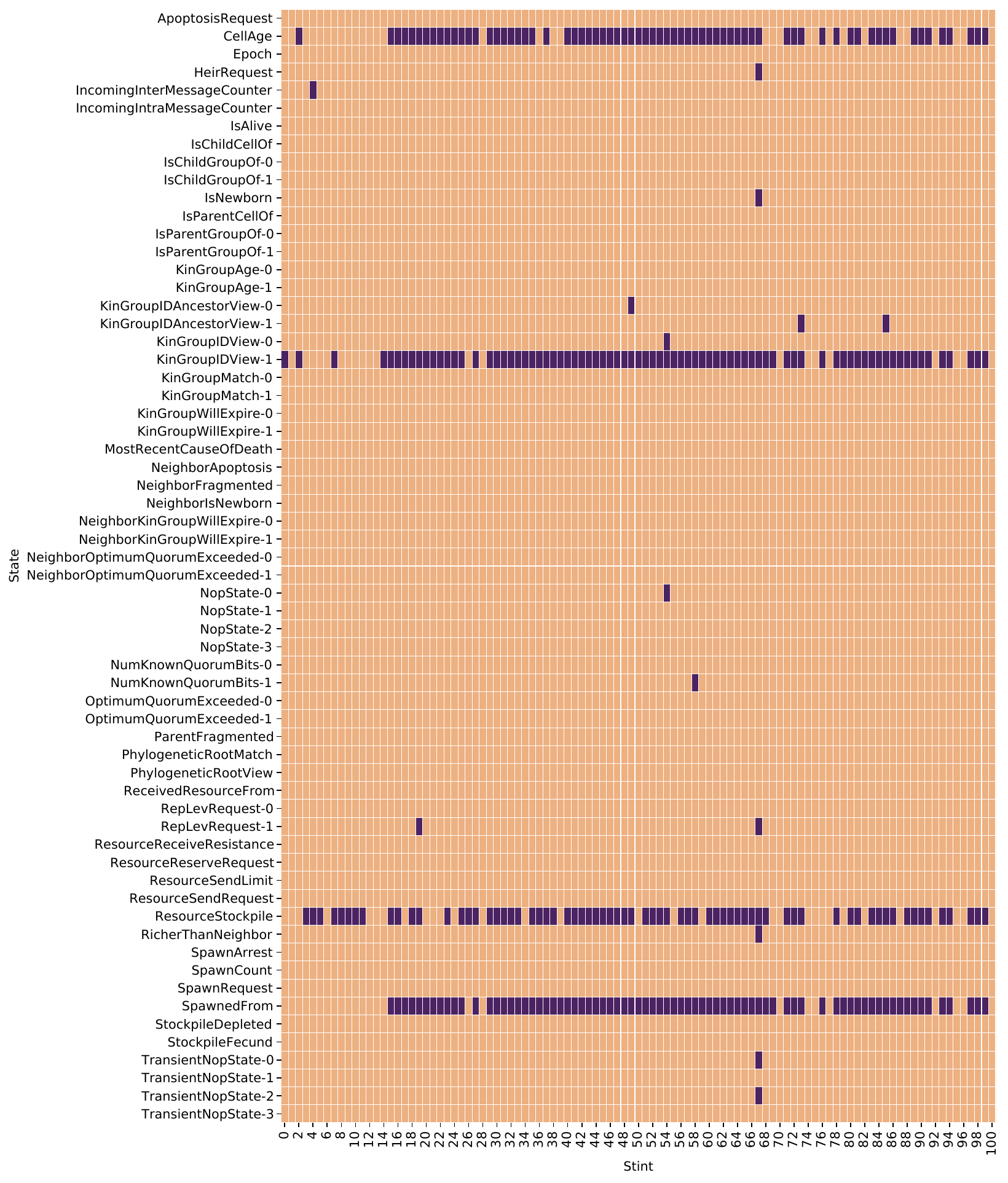}

\caption{
Fitness effect of extrospective states (read-only state information of neighboring cells) for focal strains between stint 0 and 100.
Peach color indicates no fitness effect.
Burgundy indicates a significant fitness effect.
Supplementary material provides a description for each state.
}
\label{fig:extrospective_perturbation}
\end{figure*}

\begin{figure*}

\includegraphics[width=\linewidth]{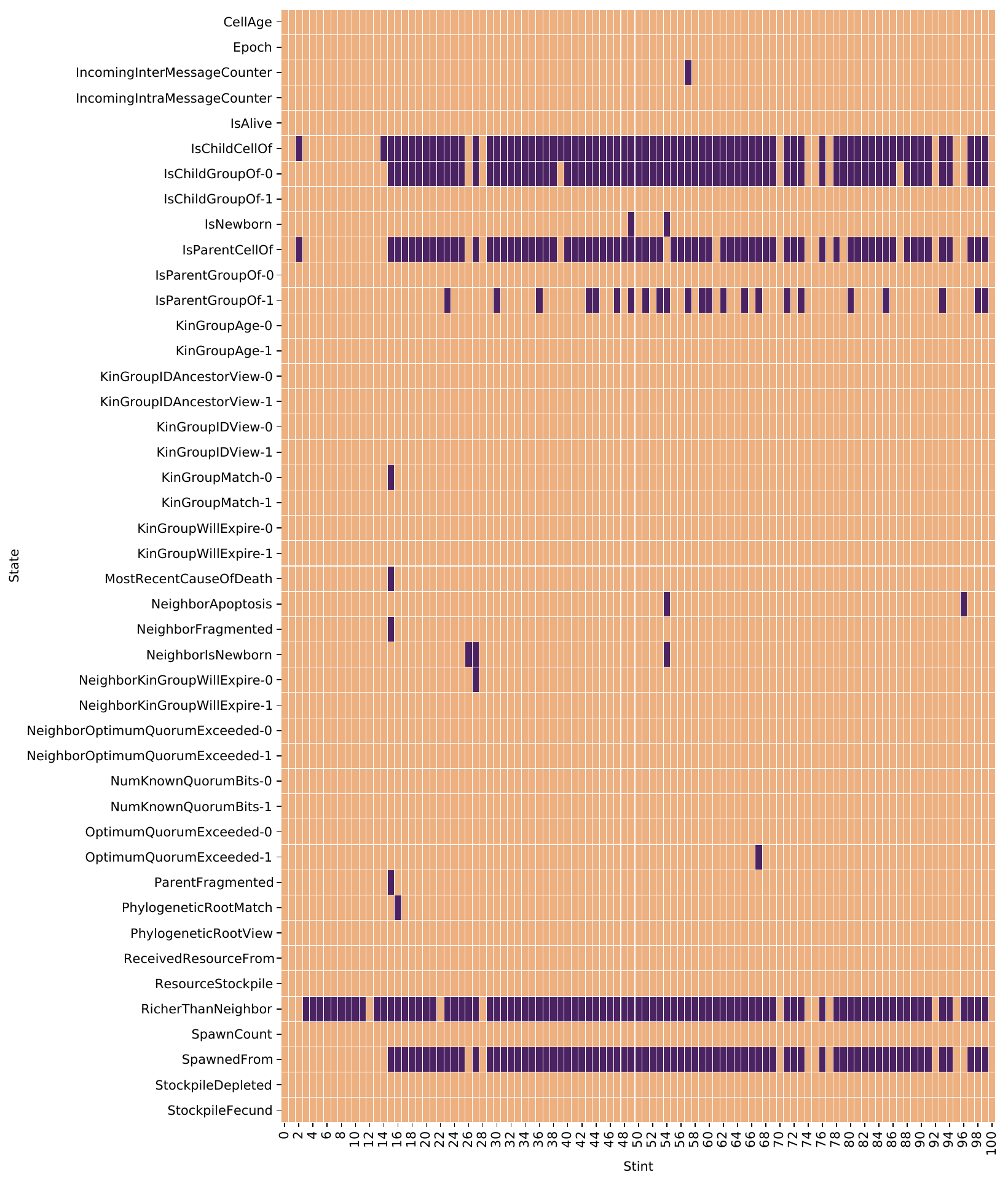}

\caption{ Fitness effect of introspective states (read-only state information of own cell) for focal strains between stint 0 and 100.
Peach color indicates no fitness effect.
Burgundy indicates a significant fitness effect.
Supplementary material provides a description for each state. }
\label{fig:introspective_perturbation}
\end{figure*}

\begin{figure*}

\includegraphics[width=\linewidth]{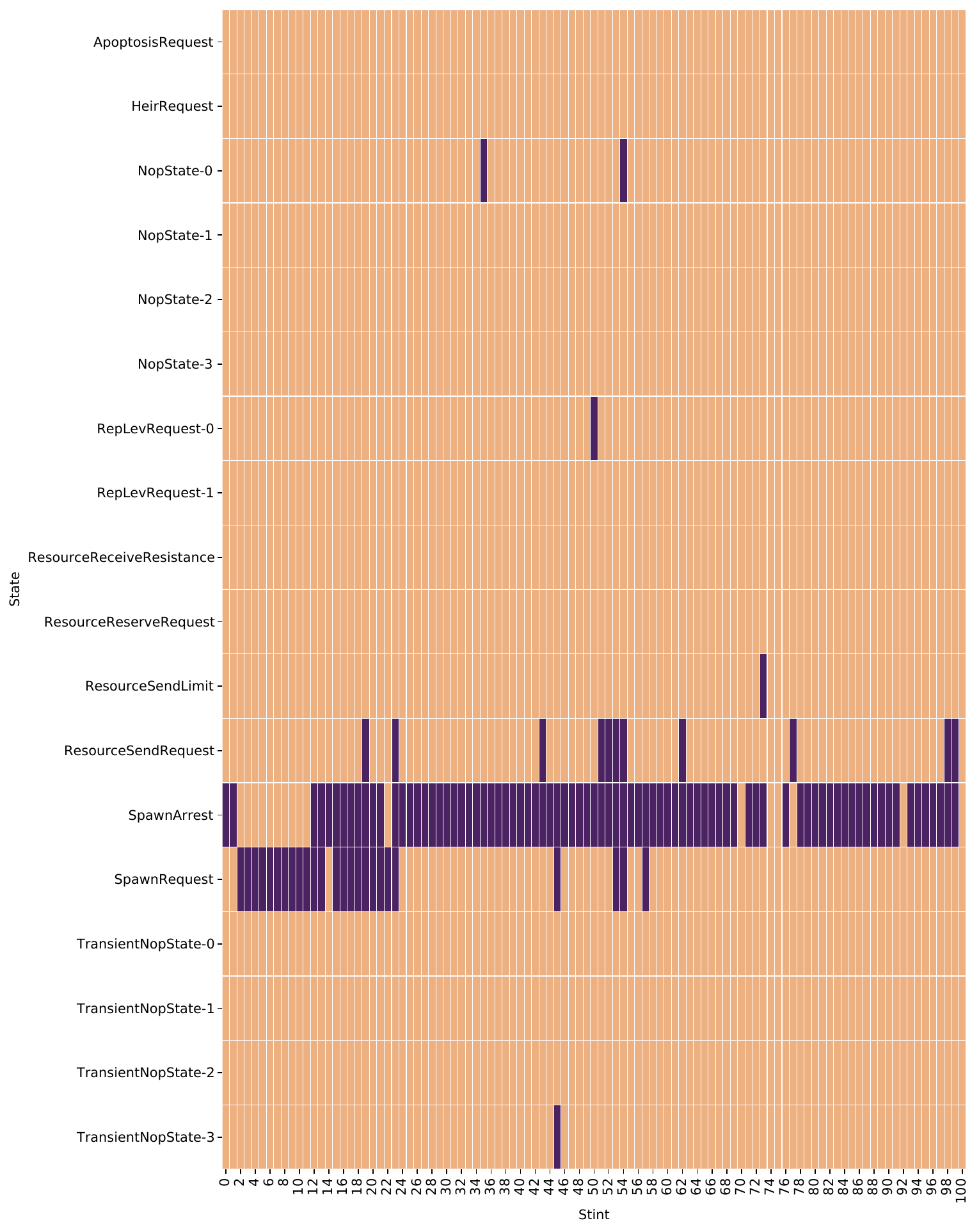}

\caption{ Fitness effect of writable states for focal strains between stint 0 and 100.
Peach color indicates no fitness effect.
Burgundy indicates a significant fitness effect.
Supplementary material provides a description for each state. }
\label{fig:writable_perturbation}
\end{figure*}




In the following sections, $L$ refers to the number of hierarchical kin group levels defined for the simulation.
In this work, we use $L=2$.

\section{Virtual CPU} \label{sec:virtual_cpu}

Each cardinal processor hosts a signalgp-lite virtual CPU \citep{lalejini2018evolving,moreno2021signalgp}.
Each CPU can host up to 16 active virtual cores.
If additional cores are required after all 16 available are in use, the oldest active core is killed and replaced.
Each virtual core contains 8 virtual \texttt{float} registers.

Cores execute round-robin in quasi-parallel, with up to 8 instructions being executed on a single core before execution shifts to the next active core.

Like SignalGP, the signalgp-lite system uses tag-matching to determine which modules to activate in response to incoming signals from the environment, from other agents (i.e., messages), and from internal events (i.e., execution of \texttt{call} and \texttt{fork} instructions).

In the DISHTINY simulation, each CPU hosts two independent module-lookup data structures.
The first module-lookup data structure is used to activate modules in response to internally-generated signals, messages from other cardinal processors within the same cell, and environmental events;
this module-lookup data structure contains \textit{all} modules within the genetic program.
The second module-lookup data structure is used to activate modules in response to messages from other cells;
this module-lookup data structure contains only modules with bitstring tags that end in 0.
(Hence, the subset of modules with bitstring tags that end in 1 \textit{cannot} be activated by messages from other cells, so that sensitive functionality like resource sharing and apoptosis can be protected from potentially malicious exploitation.)

We also use a tag-matching system to route \texttt{jump} instructions executed within a module.
When a module is loaded, all \texttt{local anchor} instructions are registered within a tag-matching data structure.
The local \texttt{jump} instruction routes to the best-matching local anchor.
If no matching local anchor is available, then no jump is performed and execution continues as if a \texttt{nop} instruction had elapsed.

\section{Tag Matching}

We use 64-bit bitstring tags to label modules and jump destinations.
We use a variant of Downing's streak metric to compute tag matches \citep{downing2015intelligence,moreno2023matchmaker}.
We deterministically select the single best-matching result for a tag lookup.
If there is a tie, an arbitrary result is selected.
For module-lookup, the best-matching tag must have a match quality at the 80th or better percentile among match qualities of pairs of randomly generated tags.
Otherwise, no module is activated.
For jump-lookup, the best match must be at the 50th or better percentile.
Otherwise, no jump is performed.

\section{Program Generation and Mutation}

Initial populations were seeded with programs consisting of 128 randomly generated instructions.
Program length was capped at 4096 instructions.

Mutation was applied to one in 10 reproductions where any kin group commonality was maintained and to all reproductions where it was not.
If mutation occurred, bits in the binary representation of the genome were flipped with 0.02\% probability.
If mutation occurred, sequence mutations were also introduced into the program at a per-site rate of 0.1\%.
Half of sequence mutations were deletion events, with a number of sites deleted drawn uniformly between 0 and 8.
Half of sequence mutations were insertion events, with a number of sites inserted drawn uniformly between 0 and 8.
When sites were inserted, half of the time randomly-generated instructions were added and half of the time the preceding sequence of instructions was duplicated.
With 0.1\% probability a sequence mutation took on severe intensity, meaning that the number of sites inserted or deleted was drawn uniformly between 0 and program size rather than between 0 and 8.

\section{Cooperative Resource Collection}

In order to ensure kin group structure had functional ramifications, we based part of cell resource collection on the number of contiguous kin group members.
To do this, we needed an efficient distributed method to approximate kin group size.

Each cell held a 64-bit bitstring with one chosen bit fixed.
We estimated group size by counting the number of distinct set bits (out of 64 available slots) that were contained within a kin group.
We refer to this count of distinct set bits as a group's ``quorum count.''

At every update within each tile, the simulation system broadcasts all bits that were known to be set within that tile's kin group.
This broadcast was only sent to neighboring tiles that were part of the same kin group as the broadcasting tile.
Each tile tracked which neighbor it learned of each set bit from so that when tiles left the kin group, their set bits could be forgotten from the bits known to be set within the kin group.

This scheme was replicated independently for each kin group level simulated.
For the lowest-level kin group, a different fixed bit was chosen independently for each tile.
Thus, the quorum count for these lowest-level kin groups was a function of the number of cells contained.
For the highest-level kin group, each tile's fixed bit was chosen as a deterministic function of its lowest-level kin group ID.
Thus, the quorum count for these highest-level kin groups was a function of the number of lowest-level kin groups contained.

To incentivize kin group formation and maintenance, we gave each cell a 0.02 resource bonus every four updates for each non-self quorum count.
This bonus saturated at the simulation-defined target quorum count.
For both the lowest- and highest-level kin groups we used a target quorum count of 12.
The source code controlling cooperative resource collection can be found at \url{https://github.com/mmore500/dishtiny/blob/prq49/include/dish2/services/CollectiveHarvestingService.hpp}.

In addition to this cooperative resource collection mechanism, cells enjoyed a continuous resource inflow of 0.02 units per update.
The source code controlling base resource inflow can be found at \url{https://github.com/mmore500/dishtiny/blob/master/include/dish2/services/ResourceHarvestingService.hpp}.

To penalize groups that expanded beyond the simulation-defined target quorum count, we decayed held resource by a multiplicative factor of $0.9995^{2^n}$, where $n$ is the excess quorum count beyond the simulation-defined target.
The source code controlling cooperative resource collection can be found at \url{https://github.com/mmore500/dishtiny/blob/prq49/include/dish2/services/CollectiveResourceDecayService.hpp}.

In addition to any decay due to group size, held resource decayed at a rate of 0.05\% per update.
Received resource was decayed by 0.099975\% upon receipt.

\section{Simulation Details}
\label{sec:simulation-details}

\newenvironment{leveldown}
  {\let\section\subsection%
   \let\subsection\subsubsection%
   \let\subsubsection\paragraph%
   \let\paragraph\subparagraph%
  }{}

\begin{leveldown}
\section{Events}

This section enumerates simulation-managed events that were dispatched on virtual CPUs.
In addition to a program, each genome contained an array of 64-bit tags --- one for each event.
When an event's criteria was met in the simulation, the genome's corresponding tag was used to dispatch a module in the program and launch a core executing that module.

All events are also exposed to the cell as a corresponding input sensor.
The state of the event (0 for false, 1 for true) is stored in the sensor prior to virtual CPU execution.
In fact, events are triggered based on the reading of the sensor register (not by re-reading the underlying simulation state).
This means that experimental perturbations that perturb sensor input also disrupted event-handling, allowing the state interface complexity metric to measure both event-driven and sensor-based behaviors.

The source code controlling events can be found at \url{https://github.com/mmore500/dishtiny/tree/prq49/include/dish2/events} and \url{https://github.com/mmore500/dishtiny/blob/prq49/include/dish2/services/InterpretedIntrospectiveStateRefreshService.hpp}.

\subsection{Always}

This event is always dispatched.

\subsection{Is Child Cell Of}

Is this cell a daughter cell of the cardinal processor's neighbor?
Triggered if this cell was spawned from the cardinal processor's neighbor and its cell is younger than the neighbor.

\subsection{Is Child Group Of (0 thru $L-1$)}

Is this cell's kin group a daughter group of the cardinal processor's neighbor cell's kin group?
Triggered if a cell's kin group ancestor ID(s) are equal to the cardinal processor's neighbor's current kin group ID(s).

\subsection{Is Newborn}

This event is dispatched once when a cell is first born.
Triggered if cell age is less than frequency at which events are launched.

\subsection{Is Parent Cell Of}

Is this cardinal processor's cell the parent cell of the cardinal processor's neighbor?
Triggered if neighbor was spawned from cell and cell age is greater than neighbor age.

\subsection{Kin Group Match (0 thru $L-1$)}

Is this cell part of the same kin group as the cardinal processor's neighbor?
Triggered if a cell's kin group ID(s) are equal to the cardinal processor's neighbors' current kin group ID(s).

\subsection{Kin Group Mismatch (0 thru $L-1$)}

Is this cell part of a different kin group from the cardinal processor's neighbor cell?
Triggered if a cell's kin group ID(s) are not equal to the cardinal processor's neighbors' current kin group ID(s).

\subsection{Kin Group will Expire (0 thru $L-1$)}

Triggered if kin group age is greater than 80\% of the kin group expiration duration.
(Depending on experiment configuration, the group may be force-fragmented after expiration.)

\subsection{Kin Group will not Expire (0 thru $L-1$)}

Triggered if kin group age is less than or equal to than 80\% of the kin group expiration duration.

\subsection{Neighbor Apoptosis}

Triggered if the most recent cell death in the cardinal processor's neighbor tile was apoptosis.

\subsection{Neighbor Fragmented}

Triggered if the most recent cell death in the cardinal processor's neighbor tile was fragmentation.

\subsection{Neighbor Is Alive}

Triggered if a cardinal processor's neighbor tile is occupied by a live cell.

\subsection{Neighbor Is Newborn}

Triggered once for each time a newborn spawns into the cardinal processor's neighboring tile.
Triggered if the cardinal processor's neighbor's age is less than the frequency at which events are launched.

\subsection{Neighbor Is Not Alive}

Triggered if the cardinal processor's neighboring tile is not occupied.

\subsection{Neighbor Kin Group Will Expire (0 thru $L-1$)}

Triggered if the cardinal processor's cell neighbor's kin group age is less than or equal to 80\% of the kin group expiration duration.

\subsection{Neighbor Optimum Quorum Exceeded}

Triggered if the cardinal processor's cell neighbor's number of known quorum bits exceed the target quorum count.

\subsection{Optimum Quorum Exceeded (0 thru $L-1$)}

Triggered if the cell's number of known quorum bits exceed the target quorum count.

\subsection{Optimum Quorum Not Exceeded (0 thru $L-1$)}

Triggered if the cell's number of known quorum bits is less than or equal to than the target quorum count.

\subsection{Parent Fragmented}

Triggered if a cell's parent died from fragmentation.
That is, if the last cause of death on the current tile was fragmentation.

\subsection{Phylogenetic Root Match}

Does this cell descend from the same originally-generated genome as its neighbor?
Triggered if a cardinal processor's cell's root ID is equal to that cardinal processor's neighbor cell's root ID.

\subsection{Phylogenetic Root Mismatch}

Does this cell and its neighbor descend from a different originally-generated genomes?
Triggered if a cardinal processor's cell's root ID is not equal to that cardinal processor's neighbor cell's Root ID.

\subsection{Poorer Than Neighbor}

Does this cell have less resource stockpiled than its neighbor?
Triggered if a cardinal processor's cell has less resource than that cardinal processor's neighbor cell.

\subsection{Received Resource From}

Triggered if a cardinal processor's cell has received resource from that cardinal processor's neighbor cell.

\subsection{Richer Than Neighbor}

Does this cell have more resource stockpiled than its neighbor?
Triggered if a cardinal processor's cell has more resource in its stockpile than that cardinal processor's neighbor cell.

\subsection{Stockpile Depleted}

Is this cell's stockpile empty?
Triggered if a cell's stockpile is less than twice the base harvest rate.

\subsection{Stockpile Fecund}

Does this cell have enough stockpiled resource to fund cellular reproduction?
Triggered if a cell's stockpile is greater than 1.0.

\section{Operations}

\newcommand{\opdef}[2]{
    \begin{tabular}{|
        >{\columncolor[HTML]{C0C0C0}}l |l|}
        \hline
        Prevalence & #1 \\ \hline
        Num Args   & #2 \\ \hline
    \end{tabular}
}

This section overviews the operation library made available to evolving signalgp-lite genetic programs within the simulation.

Within the program section of each genome, each instruction contained
\begin{itemize}
\item an op code, specifying which operation should be performed;
\item a 64-bit bitstring, used as a tag for operations that required tag-matching or as data for some configurable operations; and
\item three integer arguments, specifying which registers the operation should apply to (many operations do not use all arguments).
\end{itemize}

In the operation descriptions below, we refer to register access via to the $n$th argument as \texttt{reg[arg\_n]}.
Each core has its own eight \texttt{float} registers.
All core registers are zeroed out at core launch.

In order to prevent bread-and-butter operations like global anchors, local anchors, and terminals from being swamped out by large instruction set size, we manually defined increased ``prevalences'' for some instructions.
This prevalence increased the probability of the operation being selected under mutations and initial program generation.
Prevalence works like increasing the number of identical copies of the operation included in the operation library.
We provide the prevalence of each operation below.

See \url{https://github.com/mmore500/dishtiny/tree/prq49/include/dish2/operations} for the source code of DISHTINY-specific operations and \url{https://github.com/mmore500/signalgp-lite/tree/b6c437f44136651aa6f4051d84bc62a86c2afbbe/include/sgpl/operations} for the source code of generic operations.

Refer to Section \ref{sec:virtual_cpu} for details on the virtual CPU running these instructions.

\subsection{Fork If}

\opdef{1}{1}

If \texttt{reg[arg\_0]} is nonzero, registers a request to activate a new core with the module best-matching the current instruction's tag.
These fork requests are only handled when the current core terminates.
Each core may only register 3 fork requests.

\subsection{Nop, 0 RNG Touches}

\opdef{1}{0}

Performs no operation for one virtual CPU cycle.

\subsection{Nop, 1 RNG Touches}

\opdef{1}{0}

Performs no operation for one virtual CPU cycle, and advances the RNG engine once.
(Important to nop-out operations that perform one RNG touch without causing side effects.)

\subsection{Nop, 2 RNG Touches}

\opdef{1}{1}

Performs no operation for one virtual CPU cycle, and advances the RNG engine twice.
(Important to nop-out operations that perform two RNG touches without causing side effects.)

\subsection{Terminate If}

\opdef{1}{1}

Terminates current core if \texttt{reg[arg\_0]} is nonzero.

\subsection{Add}

\opdef{1}{3}

Adds \texttt{reg[arg\_1]} to \texttt{reg[arg\_2]} and stores the result in \texttt{reg[arg\_0]}.

\subsection{Divide}

\opdef{1}{3}

Divides \texttt{reg[arg\_1]} by \texttt{reg[arg\_2]} and stores the result in \texttt{reg[arg\_0]}.
Division by zero can result in an \texttt{Inf} or \texttt{NaN} value.

\subsection{Modulo}

\opdef{1}{3}

Calculates the modulus of \texttt{reg[arg\_1]} by \texttt{reg[arg\_2]} and stores the result in \texttt{reg[arg\_0]}.
Mod by zero can result in a \texttt{NaN} value.

\subsection{Multiply}

\opdef{1}{3}

Multiplies \texttt{reg[arg\_1]} by \texttt{reg[arg\_2]} and stores the result in \texttt{reg[arg\_0]}.

\subsection{Subtract}

\opdef{1}{3}

Subtracts \texttt{reg[arg\_2]} from \texttt{reg[arg\_1]} and stores the result in \texttt{reg[arg\_0]}.

\subsection{Bitwise And}

\opdef{1}{3}

Performs a bitwise AND of \texttt{reg[arg\_1]} and \texttt{reg[arg\_2]} then stores the result in \texttt{reg[arg\_0]}.

\subsection{Bitwise Not}

\opdef{1}{2}

Computes the bitwise NOT of \texttt{reg[arg\_1]} and stores the result in \texttt{reg[arg\_0]}.

\subsection{Bitwise Or}

\opdef{1}{3}

Performs a bitwise OR of \texttt{reg[arg\_1]} and \texttt{reg[arg\_2]} then stores the result in \texttt{reg[arg\_0]}.

\subsection{Bitwise Shift}

\opdef{1}{3}

Shifts the bits of \texttt{reg[arg\_1]} by \texttt{reg[arg\_2]} positions.
(If \texttt{reg[arg\_2]} is negative, this is a right shift.
Otherwise it is a left shift.)
Stores the result in \texttt{reg[arg\_0]}.

\subsection{Bitwise Xor}

\opdef{1}{3}

Performs a bitwise XOR of \texttt{reg[arg\_1]} and \texttt{reg[arg\_2]} then stores the result in \texttt{reg[arg\_0]}.

\subsection{Count Ones}

\opdef{1}{2}

Counts the number of bits set in \texttt{reg[arg\_1]} and stores the result in \texttt{reg[arg\_0]}.

\subsection{Random Fill}

\opdef{1}{1}

Fills register pointed to by \texttt{reg[arg\_0]} with random bits chosen from a uniform distribution.

\subsection{Equal}

\opdef{1}{3}

Checks whether \texttt{reg[arg\_1]} is equal to \texttt{reg[arg\_2]} and stores the result in \texttt{reg[arg\_0]}.

\subsection{Greater Than}

\opdef{1}{3}

Checks whether \texttt{reg[arg\_1]} is greater than \texttt{reg[arg\_2]} and stores the result in \texttt{reg[arg\_0]}.

\subsection{Less Than}

\opdef{1}{3}

Checks whether \texttt{reg[arg\_1]} is less than \texttt{reg[arg\_2]} and stores the result in \texttt{reg[arg\_0]}.

\subsection{Logical And}

\opdef{1}{3}

Performs a logical AND of \texttt{reg[arg\_1]} and \texttt{reg[arg\_2]}, storing the result in \texttt{reg[arg\_0]}.

\subsection{Logical Or}

\opdef{1}{3}

Performs a logical OR of \texttt{reg[arg\_1]} and \texttt{reg[arg\_2]}, storing the result in \texttt{reg[arg\_0]}.

\subsection{Not Equal}

\opdef{1}{3}

Checks whether \texttt{reg[arg\_1]} is not equal to \texttt{reg[arg\_2]} and stores the result in \texttt{reg[arg\_0]}.

\subsection{Global Anchor} \label{sec:global_anchor}

\opdef{15}{0}

Marks a module-begin position.
Based on tag-lookup, new cores or global jump instructions may set the program counter to this instruction's program position.

This instruction can also mark a module-end position --- executing this instruction can terminate the executing core.
If no local anchor instruction is present between the current global anchor instruction and the preceding global anchor instruction, this operation will not terminate the executing core.
(This way, several global anchors may lead into the same module.)

However, if a local anchor instruction is present between the current global anchor instruction and the preceding global anchor instruction, this operation will terminate the executing core.
Local jump instructions will only consider local anchors between the preceding global anchor and the subsequent global anchor instruction.

\subsection{Global Jump If}

\opdef{1}{2}

Jumps the current core to a global anchor that matches the instruction tag if \texttt{reg[arg\_0]} is nonzero.
If \texttt{reg[arg\_1]} is nonzero, resets registers.

\subsection{Global Jump If Not}

\opdef{1}{2}

Jumps the current core to a global anchor that matches the instruction tag if \texttt{reg[arg\_0]} is nonzero.
If \texttt{reg[arg\_1]} is zero, resets registers.

\subsection{Protected Regulator Adjust}

\opdef{1}{1}

Adjusts the regulator value of global jump table tags matching this instruction's tag by the amount \texttt{reg[arg\_0]}.

This regulator value affects the outcome of tag lookup for internal events and signals from the environment.
(Note, as described in \ref{sec:virtual_cpu}, that independent tag lookup tables handle activating genome modules across different contexts.)

\subsection{Protected Regulator Decay}

\opdef{1}{1}

Ages the regulator decay countdown of global jump table tags matching this instruction's tag by the amount \texttt{reg[arg\_0]}.
If \texttt{reg[arg\_0]} is negative, this can forestall decay.

This decay countdown affects the outcome of tag lookup for internal events, and signals from the environment.
(Note, as described in \ref{sec:virtual_cpu}, that independent tag lookup tables handle activating genome modules across different contexts.)

\subsection{Protected Regulator Get}

\opdef{1}{1}

Gets the regulator value of the global jump table tag that best matches this instruction's tag.
Stores the value in \texttt{reg[arg\_0]}.

If no tag matches, a no-op is performed.

The regulator value gotten controls internal events and signals from the environment.
(Note, as described in \ref{sec:virtual_cpu}, that independent tag lookup tables handle activating genome modules across different contexts.)

\subsection{Protected Regulator Set}

\opdef{1}{1}

Sets the regulator value of global jump table tags matching this instruction's tag to \texttt{reg[arg\_0]}.

This regulator value affects the outcome of tag lookup for internal events and signals from the environment.
(Note, as described in \ref{sec:virtual_cpu}, that independent tag lookup tables handle activating genome modules across different contexts.)

\subsection{Local Anchor}

\opdef{20}{0}

Marks a program location local jump instructions may route to.
This program location is tagged with the instruction's tag.

As described in Section \ref{sec:global_anchor}, this operation also plays a role in determining whether global anchor instructions close a module.

\subsection{Local Jump If}

\opdef{1}{1}

Jumps to a local anchor that matches the instruction tag if \texttt{reg[arg\_0]} is nonzero.

\subsection{Local Jump If Not}

\opdef{1}{1}

Jumps to a local anchor that matches the instruction tag if \texttt{reg[arg\_0]} is zero.

\subsection{Local Regulator Adjust}

\opdef{1}{1}

Adjusts the regulator value of local jump table tags matching this instruction's tag by the amount \texttt{reg[arg\_0]}.

\subsection{Local Regulator Decay}

\opdef{1}{1}

Ages the regulator decay countdown of local jump table tags matching this instruction's tag by the amount \texttt{reg[arg\_0]}.
If \texttt{reg[arg\_0]} is negative, this can forestall decay.

\subsection{Local Regulator Get}

\opdef{1}{1}

Gets the regulator value of the local jump table tag that best matches this instruction's tag.
Stores the value in \texttt{reg[arg\_0]}.

If no tag matches, a no-op is performed.

\subsection{Local Regulator Set}

\opdef{1}{1}

Sets the regulator value of global jump table tags matching this instruction's tag to \texttt{reg[arg\_0]}.

\subsection{Decrement}

\opdef{1}{1}

Takes \texttt{reg[arg\_0]}, decrements it by one, and stores the result in \texttt{reg[arg\_0]}.

\subsection{Increment}

\opdef{1}{1}

Takes \texttt{reg[arg\_0]}, increments it by one, and stores the result in \texttt{reg[arg\_0]}.

\subsection{Negate}

\opdef{1}{1}

Negates \texttt{reg[arg\_0]} and stores the result in \texttt{reg[arg\_0]}.

\subsection{Not}

\opdef{1}{1}

Performs a logical not on \texttt{reg[arg\_0]} and stores the result in \texttt{reg[arg\_0]}.

\subsection{Random Bool}

\opdef{1}{1}

Stores \texttt{1.0f} to \texttt{reg[arg\_0]} with probability determined by this instruction's tag.
Otherwise, stores \texttt{0.0f} to \texttt{reg[arg\_0]}.

\subsection{Random Draw}

\opdef{1}{1}

Stores a randomly drawn float value to \texttt{reg[arg\_0]}.

\subsection{Terminal}

\opdef{50}{1}

Stores a genetically-encoded value to \texttt{reg[arg\_0]}.
This value is determined deterministically using the instruction's tag.

\subsection{Exposed Regulator Adjust}

\opdef{1}{1}

Adjusts the regulator value of global jump table tags matching this instruction's tag by the amount \texttt{reg[arg\_0]}.

This regulator value affects the outcome of tag lookup for messages from neighbor cells.
(Note, as described in \ref{sec:virtual_cpu}, that independent tag lookup tables handle activating genome modules across different contexts.)

\subsection{Exposed Regulator Decay}

\opdef{1}{1}

Ages the regulator decay countdown of global jump table tags matching this instruction's tag by the amount \texttt{reg[arg\_0]}.
If \texttt{reg[arg\_0]} is negative, this can forestall decay.

This decay countdown affects the outcome of tag lookup for messages from neighbor cells.
(Note, as described in \ref{sec:virtual_cpu}, that independent tag lookup tables handle activating genome modules across different contexts.)

\subsection{Exposed Regulator Get}

\opdef{1}{1}

Gets the regulator value of the global jump table tag that best matches this instruction's tag.
Stores the value in \texttt{arg[0]}.

If no tag matches, a no-op is performed.

The regulator value gotten controls messages from other cells.
(Note, as described in \ref{sec:virtual_cpu}, that independent tag lookup tables handle activating genome modules across different contexts.)

\subsection{Exposed Regulator Set}

\opdef{1}{1}

Sets the regulator value of global jump table tags matching this instruction's tag to \texttt{reg[arg\_0]}.

This regulator value affects the outcome of tag lookup for messages from other cells.
(Note, as described in \ref{sec:virtual_cpu}, that independent tag lookup tables handle activating genome modules across different contexts.)

\subsection{Add to Own State}

\opdef{5}{1}

Adds \texttt{reg[arg\_0]} to the current value in a target writable state then stores the sum back in to that target writable state.

To determine the target writable state, interprets the first 32 bits of the instruction tag as an unsigned integer then calculates the remainder of integer division by the number of writable states.

\subsection{Broadcast Intra Message If}

\opdef{1}{1}

If \texttt{reg[arg\_0]} is nonzero, generates a message tagged with the instruction's tag that contains the core's current register state.
Broadcasts this message to every other cardinal processors within the cell.

\subsection{Multiply Own State}

\opdef{5}{1}

Multiplies \texttt{reg[arg\_0]} by the current value in a target writable state then stores the result back in to that target writable state.

To determine the target writable state, interprets the first 32 bits of the instruction tag as an unsigned integer then calculates the remainder of integer division by the number of writable states.

\subsection{Read Neighbor State}

\opdef{10}{1}

Reads a target readable state from the neighboring cell and stores it into \texttt{reg[arg\_0]}.

To determine the target readable state, interprets the first 32 bits of the instruction tag as an unsigned integer then calculates the remainder of integer division by the number of readable states.

\subsection{Read Own State}

\opdef{20}{1}

Reads a target readable state and stores it into \texttt{reg[arg\_0]}.

To determine the target readable state, interprets the first 32 bits of the instruction tag as an unsigned integer then calculates the remainder of integer division by the number of readable states.

\subsection{Send Inter Message If}

\opdef{5}{1}

If \texttt{reg[arg\_0]} is nonzero, generates a message tagged with the instruction's tag that contains the core's current register state.
Sends this message to the neighboring cell.

\subsection{Send Intra Message If}

\opdef{5}{1}

If \texttt{reg[arg\_0]} is nonzero, generates a message tagged with the instruction's tag that contains the core's current register state.
Sends this message to a target cardinal processor within the cell.

To determine the target cardinal processor, sums instruction arguments then calculates the remainder of integer division by the number of other same-cell cardinal processors.

\subsection{Write Own State If}

\opdef{5}{2}

If \texttt{reg[arg\_1]} is nonzero, stores \texttt{reg[arg\_0]} into a target writable state.

To determine the target writable state, interprets the first 32 bits of the instruction tag as an unsigned integer then calculates the remainder of integer division by the number of writable states.

\newcommand{\instrospectivestatedef}[2]{
    \begin{tabular}{|
        >{\columncolor[HTML]{C0C0C0}}l |l|}
        \hline
        Type & #1 \\ \hline
        Category & #2 \\ \hline
    \end{tabular} \\
}

\section{Introspective State}

Introspective state refers to the collection of simulation-generated sensor values that evolving programs running within each cardinal processor can access via read-only operations.
Each cardinal processor has an independent copy of each piece of introspective state state.
(However, some introspective states representing cell state are set to identical values across cardinal processors within the same cell.)

Each cardinal processor's introspective state regularly copied and dispatched to that cardinal processor's neighbor cell, where it serves as read-only extrospective state.

Introspective state is organized into two categories:
\begin{enumerate}
    \item raw introspective state, and
    \item interpreted introspective state.
\end{enumerate}

Raw introspective state directly exposes aspects of simulation state.
Interpreted introspective state is filled with truthy values that are interpreted as booleans to dispatch environmentally-managed events.

See \url{https://github.com/mmore500/dishtiny/tree/prq49/include/dish2/peripheral/readable_state/introspective_state} for source code implementing introspective state.

\subsection{Is Child Cell Of}

\instrospectivestatedef{\texttt{char} (w/ boolean semantics)}{interpreted}

Did this cell spawn from this cardinal processor's neighbor cell?

\subsection{Is Child Group Of (0 thru $L-1$)}

\instrospectivestatedef{\texttt{char} (w/ boolean semantics)}{interpreted}

Does this cell's kin group ID descend directly from the neighbor's kin group ID?

\subsection{Is Newborn}

\instrospectivestatedef{\texttt{char} (w/ boolean semantics)}{interpreted}

Is this cell's age less than \texttt{EVENT\_LAUNCHING\_SERVICE\_FREQUENCY}?

\subsection{Is Parent Cell Of}

\instrospectivestatedef{\texttt{char} (w/ boolean semantics)}{interpreted}

Did this cardinal processor's neighbor cell spawn from this cell?
That is, was neighbor was spawned from this cell and is this cell older than neighbor?

\subsection{Is Parent Group Of}

\instrospectivestatedef{\texttt{char} (w/ boolean semantics)}{interpreted}

Did this cell's kin group descend directly from the cardinal processor's neighbor cell's kin group?
That is, is cell's kin group ancestor ID(s) equal to the cardinal processor's neighbor's current kin group ID(s).

\subsection{Kin Group Match (0 thru $L-1$)}

\instrospectivestatedef{\texttt{char} (w/ boolean semantics)}{interpreted}

Does this cell's kin group ID match the neighbor's kin group ID?

\subsection{Kin Group will Expire (0 thru $L-1$)}

\instrospectivestatedef{\texttt{char} (w/ boolean semantics)}{interpreted}

Is this cell's kin group age greater than 80\% of this level's \texttt{GROUP\_EXPIRATION\_DURATIONS}?

\subsection{Neighbor Apoptosis}

Was the neighbor tile's most recent death apoptosis?

\instrospectivestatedef{\texttt{char} (w/ boolean semantics)}{interpreted}

\subsection{Neighbor Fragmented}

Was group fragmentation the most recent cause of death in the cardinal processor's neighbor cell?

\instrospectivestatedef{\texttt{char} (w/ boolean semantics)}{interpreted}

\subsection{Neighbor Is Newborn}

\instrospectivestatedef{\texttt{char} (w/ boolean semantics)}{interpreted}

Is the neighbor's cell age less than \texttt{EVENT\_LAUNCHING\_SERVICE\_FREQUENCY}?

\subsection{Neighbor Kin Group Will Expire (0 thru $L-1$)}{interpreted}

\instrospectivestatedef{\texttt{char} (w/ boolean semantics)}{interpreted}

Is this cell's kin group age greater than 80\% of the this level's \texttt{GROUP\_EXPIRATION\_DURATIONS}?

\subsection{Neighbor Optimum Quorum Exceeded (0 thru $L-1$)}

\instrospectivestatedef{\texttt{char} (w/ boolean semantics)}{interpreted}

Is this cardinal processor's neighbor cell's kin group quorum count more than the simulation-defined target count \texttt{OPTIMAL\_QUORUM\_COUNT}?

\subsection{Optimum Quorum Exceeded (0 thru $L-1$)}

\instrospectivestatedef{\texttt{char} (w/ boolean semantics)}{interpreted}

Is this cell's kin group quorum count more than the simulation-defined target count \texttt{OPTIMAL\_QUORUM\_COUNT}?

\subsection{Parent Fragmented}

\instrospectivestatedef{\texttt{char} (w/ boolean semantics)}{interpreted}

Did the cell's parent die from fragmentation?
That is, was the last cause of death on the current tile was fragmentation?

\subsection{Phylogenetic Root Match}

\instrospectivestatedef{\texttt{char} (w/ boolean semantics)}{interpreted}

Does this cell's root ID equal the cardinal processor's cell neighbor's root ID?
(This means they originate from the same seed ancestor.)

\subsection{Richer Than Neighbor}

\instrospectivestatedef{\texttt{char} (w/ boolean semantics)}{interpreted}

Does this cell's stockpile more resource than the cardinal processor's cell neighbor?

\subsection{Stockpile Depleted}

\instrospectivestatedef{\texttt{char} (w/ boolean semantics)}{interpreted}

Is this cell's stockpile less than twice the base harvest rate?

\subsection{Stockpile Fecund}

\instrospectivestatedef{\texttt{char} (w/ boolean semantics)}{interpreted}

Does this cell have enough resource stockpiled to fund spawning an offspring cell?

\subsection{Cell Age}

\instrospectivestatedef{\texttt{size\_t}}{raw}

Number CellAgeService calls elapsed since cell was born.

\subsection{Epoch}

\instrospectivestatedef{\texttt{size\_t}}{raw}

Updates elapsed since start of simulation.

\subsection{Incoming Inter Message Counter}

\instrospectivestatedef{\texttt{size\_t}}{raw}

Counter tracking incoming messages from cardinal processor's neighbor cell.
Intermittently reset to zero.

\subsection{Incoming Intra Message Counter}

\instrospectivestatedef{\texttt{size\_t}}{raw}

Counter of incoming messages from other cardinal processors within the cell.
Intermittently reset to zero.

\subsection{Is Alive}

\instrospectivestatedef{\texttt{char} (w/ boolean semantics)}{raw}

Whether the cell is alive.
Although trivial as introspective state, this state is useful for neighbor cell's extrospective state.

\subsection{Kin Group Age ($0$ thru $L - 1$)}

\instrospectivestatedef{\texttt{size\_t}}{raw}

Number of epochs elapsed since kin group formation.

\subsection{Kin Group ID Ancestor View ($0$ thru $L - 1$)}

\instrospectivestatedef{\texttt{size\_t}}{raw}

Kin group ID from which cell's kin group ID is descended.

\subsection{Kin Group ID View ($0$ thru $L - 1$)}

\instrospectivestatedef{\texttt{size\_t}}{raw}

Kin group ID of this cell.

\subsection{Most Recent Cause of Death}{raw}

\instrospectivestatedef{\texttt{char}}{raw}

What was this the most recent cause of death on this tile?
Encoded using the \texttt{CauseOfDeath} enum.

\subsection{Num Known Quorum Bits (0 thru $L-1$)}

\instrospectivestatedef{\texttt{size\_t}}{raw}

What is this cell's known quorum count?
(How many unique quorum bits collected from kin group members are known?)

\subsection{Phylogenetic Root View}

\instrospectivestatedef{\texttt{size\_t}}{raw}

What is this cell's phylogenetic root ID?

(Which initially-generated ancestor is this cell descended from?)

\subsection{Received Resource From}

\instrospectivestatedef{\texttt{float}}{raw}

How much resource is being received from the cardinal processor's cell neighbor?

\subsection{Resource Stockpile}

\instrospectivestatedef{\texttt{float}}{raw}

Amount of resource this cell has.

\subsection{Spawn Count}

\instrospectivestatedef{\texttt{float}}{raw}

Number of offspring generated from this cell and sent to the cardinal processor's neighbor tile.
Includes offspring that do not successfully take into the neighbor tile or have not survived.

\subsection{Spawned From}

\instrospectivestatedef{\texttt{char} (w/ boolean semantics)}{raw}

Did this cell spawn from this cardinal processor's neighbor cell?

\newcommand{\writablestatedef}[2]{
    \begin{tabular}{|
        >{\columncolor[HTML]{C0C0C0}}l |l|}
        \hline
        Type & \texttt{#1} \\ \hline
    \end{tabular} \\
}

\section{Writable State}

Writable state refers to the collection of output values that evolving programs running within each cardinal processor can write to and read from.
Some of these outputs enable interaction with the simulation (i.e., control phenotypic characteristics).
Each cardinal processor has an independent copy of each piece of writable state state.

See \url{https://github.com/mmore500/dishtiny/tree/prq49/include/dish2/peripheral/readable_state/writable_state} for source code implementing writable state.

\subsection{Nop State ($4\times$)}

\writablestatedef{float}

Writing to this state has no external effect.
It can be used as global memory shared between cores.

\subsection{Transient Nop State ($4\times$)}

\writablestatedef{float}

Writing to this state has no external effect.
It is cleared regularly by the decay to baseline service.
It can be used as temporary global memory shared between cores.

\subsection{Apoptosis Request}

\writablestatedef{char}

Writing a nonzero value to this state causes cell death.

\subsection{Heir Request}

\writablestatedef{char}

If this state is set when cell death occurs, the cardinal processor's neighbor cell will inherit leftover resource from the cell's stockpile.

\subsection{RepLev Request (0 thru $L$)}

\writablestatedef{char}

Controls kin group inheritance for daughter cells spawned to this cardinal processor's neighbor tile.

If no copies of this state are set at cell spawn, the daughter cell will have no common kin group IDs.
If one copy of this state is set at cell spawn, the daughter cell will have one common kin group ID.
If $L$ copies of this state are set at cell spawn, the daughter cell will have $L$ common kin group IDs.

\subsection{Resource Receive Resistance}

\writablestatedef{float}

Setting this state reduces the amount of resource received from the cardinal processor's neighbor cell.

\subsection{Resource Reserve Request}

\writablestatedef{float}

Setting this state prevents that amount of stockpiled resource from being drawn from to be sent to the cardinal processor's neighbor cell.

\subsection{Resource Send Limit}

\writablestatedef{float}

Setting this state caps the amount of resource that this cell can send to the cardinal processor's neighbor cell per update.

\subsection{Resource Send Request}

\writablestatedef{float}

Setting this state initiates resource sharing to the cardinal processor's neighbor cell.
The value stored controls the amount of resource shared.

\subsection{Spawn Arrest}

\writablestatedef{char}

Setting this state prevents this cell from spawning offspring into this cardinal processor's neighbor tile, even if sufficient resource is available.

\subsection{Spawn Request}

\writablestatedef{char}

Setting this state attempts to initiate spawning offspring into this cardinal processor's neighbor tile.

\newcommand{\cellsimservicedef}[2]{
    \begin{tabular}{|
        >{\columncolor[HTML]{C0C0C0}}l |l|}
        \hline
        Frequency & every #1 update(s) \\ \hline
    \end{tabular} \\
}

\section{Cellular Simulation Services}

Simulation logic is applied to each cell through a collection of distinct functors, referred to as services.

All services specified to run on a particular update are applied in sequence to a single cell.
(Some services run only every $n$th update.)
Then, to another randomly-chosen cell in a \texttt{thread\_local} population, and another until the entire population has been updated.

See \url{https://github.com/mmore500/dishtiny/tree/prq49/include/dish2/services} for source code implementing these services.

\subsection{Decay to Baseline Service}

\cellsimservicedef{32}

Decays a cell's global regulators, resets its controller-mapped peripheral states, and resets its transient NOP states.

\subsection{Running Log Purge Service}

\cellsimservicedef{64}

Purges a cell's running logs.
(Only affects data collection, not simulation logic.)

\subsection{Controller Mapped State Noise Service}

\cellsimservicedef{8}

Given a non-zero controller-mapped state defect rate, picks a random number $n$ from a Poisson distribution parameterized by \texttt{CONTROLLER\_MAPPED\_STATE\_DEFECT\_RATE}.
Then, it introduces $n$ defects to a cell's writable state.
Half of these defects zero out the state and half randomize it.

\subsection{Interpreted Introspective State Refresh Service}

\cellsimservicedef{i}

Refreshes the interpreted introspective state of a cell.

\subsection{Extrospective State Exchange Service}

\cellsimservicedef{1}

Used for experimental manipulations testing the fitness effect of extrospective state.
(Not part of core simulation logic.)

\subsection{Extrospective State Rotate Service}

\cellsimservicedef{1}

Used for experimental manipulations testing the fitness effect of extrospective state.
(Not part of core simulation logic.)

\subsection{Introspective State Exchange Service}

\cellsimservicedef{1}

Used for experimental manipulations testing the fitness effect of introspective state.
(Not part of core simulation logic.)

\subsection{Introspective State Rotate Service}

\cellsimservicedef{1}

Used for experimental manipulations testing the fitness effect of introspective state.
(Not part of core simulation logic.)

\subsection{CPU Execution Service}

\cellsimservicedef{1}

Executes a cell's genome on its cardinals processors for \texttt{HARDWARE\_EXECUTION\_CYCLES} virtual cycles.
The order of cardinal evaluation is randomized.
This is repeated \texttt{HARDWARE\_EXECUTION\_ROUNDS} times.

\subsection{Event Launching Service}

\cellsimservicedef{8}

Dispatches environmentally-managed events for each cardinal processor.

\subsection{Introspective State Rotate Restore Service}

\cellsimservicedef{1}

Used for experimental manipulations testing the fitness effect of introspective state.
(Not part of core simulation logic.)

\subsection{Introspective State Exchange Restore Service}

\cellsimservicedef{1}

Used for experimental manipulations testing the fitness effect of introspective state.
(Not part of core simulation logic.)

\subsection{Extrospective State Rotate Restore Service}

\cellsimservicedef{1}

Used for experimental manipulations testing the fitness effect of extrospective state.
(Not part of core simulation logic.)

\subsection{Extrospective State Exchange Restore Service}

\cellsimservicedef{1}

Used for experimental manipulations testing the fitness effect of extrospective state.
(Not part of core simulation logic.)

\subsection{Writable State Exchange Service}

\cellsimservicedef{1}

Used for experimental manipulations testing the fitness effect of writable state.
(Not part of core simulation logic.)

\subsection{Writable State Rotate Service}

\cellsimservicedef{1}

Used for experimental manipulations testing the fitness effect of writable state.
(Not part of core simulation logic.)

\subsection{Birth Setup Service}

\cellsimservicedef{16}

Births a new cell into the current cell.

This occurs by first iterating through the cell's cardinal processors' birth request inputs in random order.
While the cell's resource stockpile is greater than the \texttt{SPAWN\_DEFENSE\_COST}, the requests are ignored and the stockpile depleted by that cost.
The first request that cannot be defended against is then acted upon.
The current cell's death routine is called, the old genome is replaced by the incoming genome, and the cell's make-alive routine is called.

\subsection{Cell Age Service}

\cellsimservicedef{1}

Advances the cell age introspective state and refreshes kin group age introspective state.

\subsection{Collective Harvesting Service}

\cellsimservicedef{4}

Calculates the total amount of resource collectively harvested to this cell by the cell's kin group.
This amount increases with quorum count and saturates at \texttt{OPTIMAL\_QUORUM\_COUNT}.
Adds the harvested amount to the cell's resource stockpile.

\subsection{Collective Resource Decay Service}

\cellsimservicedef{1}

If the cell's quorum count exceeds \texttt{OPTIMAL\_QUORUM\_COUNT}, applies multiplicative decay to the cell's resource stockpile.
This effect strengthens exponentially with excess cell quorum count.

\subsection{Conduit Flush Service}

\cellsimservicedef{16}

Flushes each cardinal processors' inter-process and inter-thread output conduits.

\subsection{Inter Message Launching Service}

\cellsimservicedef{8}

Launches new virtual cores to process incoming inter-cell messages.

\subsection{Inter Message Purging Service}

\cellsimservicedef{8}

Purges excess incoming inter-cell messages that couldn't be handled due to virtual core availability.

\subsection{Intra Message Launching Service}

\cellsimservicedef{8}

Launches new virtual cores to process incoming messages from same-cell cardinal processors.

\subsection{Message Counter Clear Service}

\cellsimservicedef{16}

Intermittently resets introspective message count state.

\subsection{Quorum Service}

\cellsimservicedef{1}

Performs distributed estimation of kin group size by simulation.

Each cell has a single randomly-chosen index set within a fixed-length bitstring.
(Depending in parameter settings, some cells may have index set --- all positions within the bitstring are zeroed out.)

Broadcasts bits known to be set are to all neighbor cells within the same kin group.
Incoming bitstrings from neighbors are ORed with known bits.

The original neighbor each non-self bit was first learned from is recorded alongside that bit.
If that neighbor no longer broadcasts that bit, it is erased from the cell's known bits.

Updates latest quorum count into introspective state.

This scheme is replicated independently for each kin group level simulated.

\subsection{Resource Decay Service}

\cellsimservicedef{1}

Decays cell resource stockpile multiplicatively by \texttt{RESOURCE\_DECAY} constant.

\subsection{Resource Harvesting Service}

\cellsimservicedef{1}

Adds a constant amount to cell's resource stockpile.

\subsection{Resource Receiving Service}

\cellsimservicedef{4}

Calculates total amount of resource received across every cardinal processor, and then adds that total to resource stockpile.

If the cell is not alive, it instead refunds all received resources back to each sending cell.

\subsection{Resource Sending Service}

\cellsimservicedef{1}

Based on writable state within each cardinal processor, calculates and dispatches resource that should be shared to each neighbor cell.

\subsection{Spawn Sending Service}

\cellsimservicedef{16}

If available resource is greater than or equal to 1.0, iterates randomly through every cardinal processor to determine whether it requested to spawn and has not arrested spawning.
Then, one of these requests is dispatched at random and stockpile is decreased by one.

\subsection{State Input Jump Service}

\cellsimservicedef{8}

Pulls a fresh copy of each neighboring cardinal processor's current readable state.

\subsection{State Output Put Service}

\cellsimservicedef{8}

Dispatches a copy of each cardinal processor's current readable state to corresponding neighbor cells.

\subsection{Epoch Advance Service}

\cellsimservicedef{8}

The cell's current-known epoch count is advanced by one then set to the maximum of the cell's current-known epoch count and neighbor cells' current-known epoch count.

\subsection{Writable State Rotate Restore Service}

\cellsimservicedef{1}

Used for experimental manipulations testing the fitness effect of writable state.
(Not part of core simulation logic.)

\subsection{Writable State Exchange Restore Service}

\cellsimservicedef{1}

Used for experimental manipulations testing the fitness effect of writable state.
(Not part of core simulation logic.)

\subsection{Group Expiration Service}

\cellsimservicedef{64}

As group age exceeds \texttt{GROUP\_EXPIRATION\_DURATIONS}, with increasing probability fragments cell from its kin group.
This process kills the cell and replaces it in place with a daughter without kin ID commonality.

\subsection{Apoptosis Service}

\cellsimservicedef{16}

If any cardinal processors have requested apoptosis, do death routine on the cell.

\newcommand{\threadlocalsimservicedef}[2]{
    \begin{tabular}{|
        >{\columncolor[HTML]{C0C0C0}}l |l|}
        \hline
        Frequency & every #1 update(s) \\ \hline
    \end{tabular} \\
}

\section{Threadlocal Simulation Services}

Actions that are performed on each \texttt{thread\_local} population.

See \url{https://github.com/mmore500/dishtiny/tree/prq49/include/dish2/services_threadlocal} for source code implementing these services.

\subsection{Cell Update Service}

\threadlocalsimservicedef{1}

Performs each cell's simulation services, iterating over cells in randomized order.

\subsection{Diversity Maintenance Service}

\threadlocalsimservicedef{8}

Prevents any one originally-generated ancestor from sweeping the population, preserving deep phylogenetic diversity.

Counts cells that descend from each originally-seeded ancestor.
If more than \texttt{DIVERSITY\_MAINTENANCE\_PREVALENCE} of cells descend from a single seeded ancestor, decay their resource stockpiles.
The magnitude of this effect increases with excess prevalence.

\subsection{Stint Diversity Maintenance Service}

\threadlocalsimservicedef{n/a}

Prevents any one seeded or reconstituted stint-originating ancestor from sweeping the population, preserving phylogenetic diversity within a single stint.

Counts cells that descend from each seeded or reconstituted stint-originating ancestor.
If more than \texttt{STINT\_DIVERSITY\_MAINTENANCE\_PREVALENCE} of cells descend from a single seeded or reconstituted ancestor, decay their resource stockpiles.
The magnitude of this effect increases with excess prevalence.

\section{Runtime Parameters}

\newcommand{\confdef}[2]{
    ~\\ \begin{tabular}{|
        >{\columncolor[HTML]{C0C0C0}}l |l|}
        \hline
        Type & \texttt{#1} \\ \hline
        Default   & \texttt{#2} \\ \hline
    \end{tabular}
}

This section enumerates simulation parameters and provides default settings that were used.

See \url{https://github.com/mmore500/dishtiny/blob/prq49/include/dish2/config/ConfigBase.hpp} for source code defining run time parameters.

Some parameter settings were overridden in some assays.
See \url{https://github.com/mmore500/dishtiny/tree/prq49/configpacks/bucket=prq49+diversity=0.50_series+mut_freq=1.00+mut_sever=1.00} for configuration files used in each assay and \url{https://github.com/mmore500/dishtiny/tree/prq49/slurm} for runscripts used in each assay.

\subsection{EXECUTION}


\subsubsection{N\_THREADS}

\confdef{size\_t}{4}

How many threads should we run with?

\subsubsection{RUN\_UPDATES}

\confdef{bool}{\texttt{false}}

Should we run evolution or skip directly to post-processing and data collection?

\subsubsection{RUN\_UPDATES}

\confdef{size\_t}{0}

How many updates should we run the experiment for?

\subsubsection{RUN\_SECONDS}

\confdef{double}{0}

How many updates should we run the experiment for?

\subsubsection{MAIN\_TIMEOUT\_SECONDS}

\confdef{double}{10800}

After how many seconds should we time out and fail with an error?

\subsubsection{END\_SNAPSHOT\_TIMEOUT\_SECONDS}

\confdef{double}{1200}

After how many seconds should the end snapshot timeout?

\subsubsection{LOG\_FREQ}

\confdef{double}{20}

How many seconds should pass between logging progress?

\subsubsection{ASYNCHRONOUS}

\confdef{size\_t}{3}

Should updates occur synchronously across threads and processes?

\subsubsection{SYNC\_FREQ\_MILLISECONDS}

\confdef{size\_t}{100}

How often updates occur synchronously across threads and processes for async mode 1?

\subsubsection{RNG\_PRESEED}

\confdef{utin64\_t}{std::numeric\_limits<uint64\_t>::max()}

Optionally override the calculated rng preseed.

\subsubsection{THROW\_ON\_EXTINCTION}

\confdef{bool}{true}

Should we throw an exception if populations go extinct?


\subsection{EXPERIMENT}


\subsubsection{RUN\_SLUG}

\confdef{std::string}{``default''}

Run-identifying slug.

\subsubsection{PHENOTYPIC\_DIVERGENCE\_N\_UPDATES}

\confdef{size\_t}{2048}

How many updates should we run phenotypic divergence experiments for?
If phenotypic divergence is not detected within this many updates, we consider two strains to be phenotypically identical.

\subsubsection{PHENOTYPIC\_DIVERGENCE\_N\_CELLS}

\confdef{size\_t}{100}

How many cells should we simulate while testing for phenotypic divergence?

\subsubsection{STINT}

\confdef{utin64\_t}{std::numeric\_limits<uint64\_t>::max()}

How many evolutionary stints have elapsed?

\subsubsection{SERIES}

\confdef{utin64\_t}{std::numeric\_limits<uint64\_t>::max()}

Which evolutionary series are we running?

\subsubsection{REPLICATE}

\confdef{std::string}{``''}

What replicate are we running?

\subsubsection{TREATMENT}

\confdef{std::string}{``none''}

What experimental treatment has been applied?

\subsubsection{SEED\_FILL\_FRACTION}

\confdef{double}{1.0}

If we are seeding the population, what fraction of available slots should we fill?

\subsubsection{GENESIS}

\confdef{std::string}{"generate"}

How should we initialize the population?
Can be ``generate'' to randomly generate a new population,  ``reconstitute'' to load a population from file, ``monoculture'' to load a single genome from file, or ``inoculate'' to load genomes annotated with root ID keyname attributes from file.


\subsection{DEMOGRAPHICS}


\subsubsection{N\_CELLS}

\confdef{size\_t}{10000}

How many cells should be simulated?

\subsubsection{WEAK\_SCALING}

\confdef{bool}{false}

Should number of total cells be multiplied by the total number of threads (num procs times threads per proc)?

\subsubsection{N\_DIMS}

\confdef{size\_t}{DISH2\_NLEV}

What dimensionality should the toroidal mesh have?

\subsubsection{GROUP\_EXPIRATION\_DURATIONS}

\confdef{internal::nlev\_size\_t\_t}{internal::nlev\_size\_t\_t\{ 256, 1024 \}}

After how many \texttt{epochs} should groups stop collecting resource?

\subsubsection{CELL\_AGE\_DURATION}

\confdef{size\_t}{1024}

After how many epochs should cells die?


\subsection{RESOURCE}


\subsubsection{MIN\_START\_RESOURCE}

\confdef{float}{0.8}

How much resource should a cell start with?

\subsubsection{MAX\_START\_RESOURCE}

\confdef{float}{0.9}

How much resource should a cell start with?

\subsubsection{RESOURCE\_DECAY}

\confdef{float}{0.995}

How much resource should remain each update?

\subsubsection{APOP\_RECOVERY\_FRAC}

\confdef{float}{0.8}

What fraction of \texttt{REP\_THRESH} is recovered to heirs after apoptosis?

\subsubsection{SPAWN\_DEFENSE\_COST}

\confdef{float}{1.1}

What is the cost of repelling an incoming spawn?


\subsection{HARVEST}


\subsubsection{BASE\_HARVEST\_RATE}

\confdef{float}{0.02}

How much resource should cells accrue per update?

\subsubsection{COLLECTIVE\_HARVEST\_RATE}

\confdef{internal::nlev\_float\_t}{internal::nlev\_float\_t\{0.25\}}

How much resource should cells accrue per update?

\subsubsection{OPTIMAL\_QUORUM\_COUNT}

\confdef{internal::nlev\_float\_t}{internal::nlev\_size\_\_t\{12\}}

What group size does collective harvest work most effectively at?


\subsection{QUORUM}


\subsubsection{P\_SET\_QUORUM\_BIT}

\confdef{internal::nlev\_float\_t}{internal::nlev\_float\_t\{1.0\}}

What fraction of cells should have a quorum bit set?


\subsection{GENOME}

\subsubsection{PROGRAM\_START\_SIZE}

\confdef{size\_t}{128}

How many instructions should initial programs be?

\subsubsection{PROGRAM\_MAX\_SIZE}

\confdef{size\_t}{4096}

How many instructions should programs be capped at?

\subsubsection{MUTATION\_RATE}

\confdef{internal::nlev\_float\_t}{internal::nreplev\_float\_t\{0.1\}}

For each replev, what fraction of cells should be mutated at all?

\subsubsection{POINT\_MUTATION\_RATE}

\confdef{float}{0.0002}

What fraction of bits should be scrambled?

\subsubsection{SEQUENCE\_DEFECT\_RATE}

\confdef{float}{0.001}

How often should sloppy copy defect occur?

\subsubsection{MINOR\_SEQUENCE\_MUTATION\_BOUND}

\confdef{size\_t}{8}

For minor sequence mutations, at most how many instructions should be inserted or deleted?

\subsubsection{SEVERE\_SEQUENCE\_MUTATION\_RATE}

\confdef{float}{0.001}

With what probability should sequence mutation be severe?

\subsection{HARDWARE}

\subsubsection{HARDWARE\_EXECUTION\_ROUNDS}

\confdef{size\_t}{1}

How many hardware evaluation rounds to run per update?

\subsubsection{HARDWARE\_EXECUTION\_CYCLES}

\confdef{size\_t}{16}

How many hardware cycles to run per round?

\subsubsection{CONTROLLER\_MAPPED\_STATE\_DEFECT\_RATE}

\confdef{float}{0.0005}

At what rate should bits should be flipped in writable memory?"

\subsection{SERVICES}

\subsubsection{APOPTOSIS\_SERVICE\_FREQUENCY}

\confdef{size\_t}{16}

Run service every ?? updates.
Must be a power of 2.

\subsubsection{BIRTH\_SETUP\_SERVICE\_FREQUENCY}

\confdef{size\_t}{16}

Run service every ?? updates.
Must be a power of 2.

\subsubsection{CONDUIT\_FLUSH\_SERVICE\_FREQUENCY}

\confdef{size\_t}{1}

Run service every ?? updates.
Must be a power of 2.

\subsubsection{COLLECTIVE\_HARVESTING\_SERVICE\_FREQUENCY}

\confdef{size\_t}{16}

Run service every ?? updates.
Must be a power of 2.

\subsubsection{CPU\_EXECUTION\_SERVICE\_FREQUENCY}

\confdef{size\_t}{4}

Run service every ?? updates.
Must be a power of 2.

\subsubsection{GROUP\_EXPIRATION\_SERVICE\_FREQUENCY}

\confdef{size\_t}{1}

Run service every ?? updates.
Must be a power of 2.

\subsubsection{RUNNING\_LOG\_PURGE\_SERVICE\_FREQUENCY}

\confdef{size\_t}{64}

Run service every ?? updates.
Must be a power of 2.

\subsubsection{DIVERSITY\_MAINTENANCE\_SERVICE\_FREQUENCY}

\confdef{size\_t}{64}

Run service every ?? updates.
Must be a power of 2.

\subsubsection{DIVERSITY\_MAINTENANCE\_PREVALENCE}

\confdef{double}{0.25}

If an originally-seeded ancestor's descendants constitute more than this fraction of the population, decay their resource stockpiles.

\subsubsection{STINT\_DIVERSITY\_MAINTENANCE\_SERVICE\_FREQUENCY}

\confdef{size\_t}{0}

Run service every ?? updates.
Must be a power of 2.

\subsubsection{STINT\_DIVERSITY\_MAINTENANCE\_PREVALENCE}

\confdef{double}{0.25}

If a seeded or reconstituted stint-originating ancestor's descendants constitute more than this fraction of the population, decay their resource stockpiles.

\subsubsection{DECAY\_TO\_BASELINE\_SERVICE\_FREQUENCY}

\confdef{size\_t}{32}

Run service every ?? updates.
                                                                                                        Must be a power of 2.

\subsubsection{EPOCH\_ADVANCE\_SERVICE\_FREQUENCY}

\confdef{size\_t}{8}

Run service every ?? updates.
Must be a power of 2.

\subsubsection{EVENT\_LAUNCHING\_SERVICE\_FREQUENCY}

\confdef{size\_t}{8}

Run service every ?? updates.
Must be a power of 2.
Must be > 1.

\subsubsection{INTERMITTENT\_CPU\_RESET\_SERVICE\_FREQUENCY}

\confdef{size\_t}{64}

Run service every ?? updates.
Must be a power of 2.

\subsubsection{INTERMITTENT\_STATE\_PERTURB\_SERVICES\_FREQUENCY}

\confdef{size\_t}{1}

Run service every ?? updates.
Must be a power of 2.

\subsubsection{INTER\_MESSAGE\_COUNTER\_CLEAR\_SERVICE\_FREQUENCY}

\confdef{size\_t}{16}

Run service every ?? updates.
Must be a power of 2.

\subsubsection{INTER\_MESSAGE\_LAUNCHING\_SERVICE\_FREQUENCY}

\confdef{size\_t}{8}

Run service every ?? updates.
Must be a power of 2.

\subsubsection{INTRA\_MESSAGE\_LAUNCHING\_SERVICE\_FREQUENCY}

\confdef{size\_t}{1}

Run service every ?? updates.
Must be a power of 2.

\subsubsection{STATE\_OUTPUT\_PUT\_SERVICE\_FREQUENCY}

\confdef{size\_t}{8}

Run service every ?? updates.
Must be a power of 2.

\subsubsection{PUSH\_SERVICE\_FREQUENCY}

\confdef{size\_t}{16}

Run service every ?? updates.
Must be a power of 2.

\subsubsection{QUORUM\_CAP\_SERVICE\_FREQUENCY}

\confdef{size\_t}{16}

Run service every ?? updates.
Must be a power of 2.

\subsubsection{QUORUM\_SERVICE\_FREQUENCY}

\confdef{size\_t}{1}

Run service every ?? updates.
Must be a power of 2.

\subsubsection{RESOURCE\_DECAY\_SERVICE\_FREQUENCY}

\confdef{size\_t}{1}

Run service every ?? updates.
Must be a power of 2.

\subsubsection{RESOURCE\_HARVESTING\_SERVICE\_FREQUENCY}

\confdef{size\_t}{1}

Run service every ?? updates.
Must be a power of 2.

\subsubsection{RESOURCE\_INPUT\_JUMP\_SERVICE\_FREQUENCY}

\confdef{size\_t}{1}

Run service every ?? updates.
Must be a power of 2.

\subsubsection{RESOURCE\_RECEIVING\_SERVICE\_FREQUENCY}

\confdef{size\_t}{4}

Run service every ?? updates.
Must be a power of 2.

\subsubsection{RESOURCE\_SENDING\_SERVICE\_FREQUENCY}

\confdef{size\_t}{1}

Run service every ?? updates.
Must be a power of 2.

\subsubsection{SPAWN\_SENDING\_SERVICE\_FREQUENCY}

\confdef{size\_t}{16}

Run service every ?? updates.
Must be a power of 2.

\subsubsection{STATE\_INPUT\_JUMP\_SERVICE\_FREQUENCY}

\confdef{size\_t}{8}

Run service every ?? updates.
Must be a power of 2.

\subsubsection{CONTROLLER\_MAPPED\_STATE\_NOISE\_SERVICE\_FREQUENCY}

\confdef{size\_t}{8}

Run service every ?? updates.
Must be a power of 2.

\subsection{DATA}

\subsubsection{PHENOTYPE\_EQUIVALENT\_NOPOUT}

\confdef{bool}{false}

Should we make and record a phenotype equivalent nopout strain at the end of the run? Must also enable ARTIFACTS\_DUMP.

\subsubsection{BATTLESHIP\_PHENOTYPE\_EQUIVALENT\_NOPOUT}

\confdef{bool}{false}

Should we make and record a phenotype equivalent nopout strain at the end of the run?
Must also enable ARTIFACTS\_DUMP.

\subsubsection{JENGA\_PHENOTYPE\_EQUIVALENT\_NOPOUT}

\confdef{bool}{false}

Should we make and record a phenotype equivalent nopout strain at the end of the run?
Must also enable ARTIFACTS\_DUMP.

\subsubsection{JENGA\_NOP\_OUT\_SAVE\_PROGRESS\_AND\_QUIT\_SECONDS}

\confdef{size\_t}{10800}

After how many seconds should we save nop-out progress and quit?

\subsubsection{TEST\_INTERROOT\_PHENOTYPE\_DIFFERENTIATION}

\confdef{bool}{false}

Should we test for phenotype differentiation between roots?

\subsubsection{ALL\_DRAWINGS\_WRITE}

\confdef{bool}{false}

Should we generate and record drawings of the final state of the simulation?
Must also enable DATA\_DUMP.

\subsubsection{DATA\_DUMP}

\confdef{bool}{false}

Should we record data on the final state of the simulation?

\subsubsection{RUNNINGLOGS\_DUMP}

\confdef{bool}{false}

Should we dump running logs at the end of the simulation?
Must also enable DATA\_DUMP.

\subsubsection{CENSUS\_WRITE}

\confdef{bool}{false}

Should we write the cell census at the end of the simulation?
Must also enable DATA\_DUMP.

\subsubsection{ARTIFACTS\_DUMP}

\confdef{bool}{false}

Should we record data on the final state of the simulation?

\subsubsection{BENCHMARKING\_DUMP}

\confdef{bool}{false}

Should we record data for benchmarking the simulation?

\subsubsection{ROOT\_ABUNDANCES\_FREQ}

\confdef{size\_t}{0}

How many updates should elapse between recording phylogenetic root abundances?
If 0, never record phylogenetic root abundances.
Must be power of two.

\subsubsection{ABORT\_IF\_COALESCENT\_FREQ}

\confdef{size\_t}{0}

How many updates should elapse between checking for coalescence? If 0, never check for coalescence.
Must be power of two.

\subsubsection{ABORT\_IF\_EXTINCT\_FREQ}

\confdef{size\_t}{0}

How many updates should elapse between checking for coalescence? If 0, never check for coalescence. Must be power of two.

\subsubsection{ABORT\_AT\_LIVE\_CELL\_FRACTION}

\confdef{double}{0.0}

Should we terminate once a live cell fraction is reached? If 0, will not terminate.

\subsubsection{REGULATION\_VIZ\_CLAMP}

\confdef{double}{10.0}

What bounds should we clamp regulation values into before running PCA visualization?

\subsubsection{RUNNING\_LOG\_DURATION}

\confdef{size\_t}{4}

How many purge epochs should we keep events in the running log?

\subsubsection{SELECTED\_DRAWINGS\_FREQ}

\confdef{size\_t}{0}

How many updates should elapse between outputting snapshot images?

\subsubsection{DRAWING\_WIDTH\_PX}

\confdef{double}{500.0}

What should the width of the drawings be, in pixels?

\subsubsection{DRAWING\_HEIGHT\_PX}

\confdef{double}{500.0}

What should the height of the drawings be, in pixels?

\subsubsection{SELECTED\_DRAWINGS}

\confdef{std::string}{``''}

What drawings should be drawn?
Provide slugified drawer names separated by colons.

\subsubsection{ANIMATE\_FRAMES}

\confdef{bool}{false}

Should we stitch the output images into a video?
Only valid if DRAWING\_FREQ is not 0.

\subsubsection{VIDEO\_FPS}

\confdef{size\_t}{16}

How many frames per second should the video be?

\subsubsection{VIDEO\_MAX\_FRAMES}

\confdef{size\_t}{960}

At most how many frames should output video include?

\section{Compile Time Parameters}

\newcommand{\paramcompiletimedef}[2]{
    ~\\ \begin{tabularx}{\linewidth}{|
        >{\columncolor[HTML]{C0C0C0}}l |X|}
        \hline
        Type & \texttt{#1} \\ \hline
        Value   & \texttt{#2} \\ \hline
    \end{tabularx}
}

This section enumerates simulation parameters and provides default settings that were used.

See \url{https://github.com/mmore500/dishtiny/blob/prq49/include/dish2/spec/Spec_prq49.hpp} for source code defining compile time parameters.

\subsection{NLEV}

\paramcompiletimedef{size\_t}{2}

How many hierarchical kin group levels should be simulated?

\subsection{AMT\_NOP\_MEMORY}

\paramcompiletimedef{size\_t}{4}

How many nop and transient nop states should exist in the peripheral?

\subsection{STATE\_EXCHANGE\_CHAIN\_LENGTH}

\paramcompiletimedef{size\_t}{128}

How many callees should we displace state by in state exchange experiments?

\subsection{sgpl\_spec\_t::num\_cores}

\paramcompiletimedef{size\_t}{32}

How many virtual cores should each cardinal processor's virtual CPU be able to support?

\subsection{sgpl\_spec\_t::num\_fork\_requests}

\paramcompiletimedef{size\_t}{3}

How many fork requests can a virtual core make at most?

\subsection{sgpl\_spec\_t::num\_registers}

\paramcompiletimedef{size\_t}{8}

How many registers should each virtual core contain?

\subsection{sgpl\_spec\_t::switch\_steps}

\paramcompiletimedef{size\_t}{8}

Maximum num steps executed on one core before next core is executed.

\subsection{sgpl\_spec\_t::global\_matching\_t}

\paramcompiletimedef{typedef}{%
emp::MatchDepository<\newline
\hspace*{2em}// \mbox{program index type} \newline
\hspace*{2em}unsigned short, \newline
\hspace*{2em}// \mbox{match metric} \newline
\hspace*{2em}emp::OptimizedApproxDualStreakMetric<64>,\newline
\hspace*{2em}// \mbox{match selector} \newline\hspace*{2em}emp::statics::RankedSelector<\newline
\hspace*{4em}// \mbox{match threshold} \newline
\hspace*{4em}std::ratio<1, 5>\newline
\hspace*{2em}>,\newline
\hspace*{2em}// \mbox{regulator} \newline
\hspace*{2em}emp::PlusCountdownRegulator<\newline
\hspace*{4em}std::deci, // Slope\newline
\hspace*{4em}std::ratio<1,4>, // MaxUpreg\newline
\hspace*{4em}std::deci, // ClampLeeway\newline
\hspace*{4em}2 // CountdownStart\newline
\hspace*{2em}>,\newline
\hspace*{2em}true, // raw caching \newline
\hspace*{2em}8 // regulated caching\newline
>
}

What matching datastructure implementation should we use for global jump tables?

\subsection{sgpl\_spec\_t::local\_matching\_t}

\paramcompiletimedef{typedef}{%
emp::MatchDepository<\newline
\hspace*{2em}// \mbox{program index type} \newline
\hspace*{2em}unsigned short, \newline
\hspace*{2em}// \mbox{match metric} \newline
\hspace*{2em}emp::OptimizedApproxDualStreakMetric<64>,\newline
\hspace*{2em}// \mbox{match selector} \newline\hspace*{2em}emp::statics::RankedSelector<\newline
\hspace*{4em}// \mbox{match threshold} \newline
\hspace*{4em}std::ratio<1, 2>\newline
\hspace*{2em}>,\newline
\hspace*{2em}// \mbox{regulator} \newline
\hspace*{2em}emp::PlusCountdownRegulator<\newline
\hspace*{4em}std::deci, // Slope\newline
\hspace*{4em}std::ratio<1,4>, // MaxUpreg\newline
\hspace*{4em}std::deci, // ClampLeeway\newline
\hspace*{4em}2 // CountdownStart\newline
\hspace*{2em}>,\newline
\hspace*{2em}false, // raw caching \newline
\hspace*{2em}0 // regulated caching\newline
>
}

What matching datastructure implementation should we use for local jump tables?

\subsection{sgpl\_spec\_t::tag\_width}

\paramcompiletimedef{size\_t}{64}

Tag width in bits.

\end{leveldown}





\providecommand{\dissertationonly}[1]{%
\ifdefined\DISSERTATION%
#1%
\else%
\fi
}

\section{Supplementary Figures}

\dissertationonly{This section provides supplementary figures for Chapter \ref{ch:measuring-cna}.}

\begin{figure*}
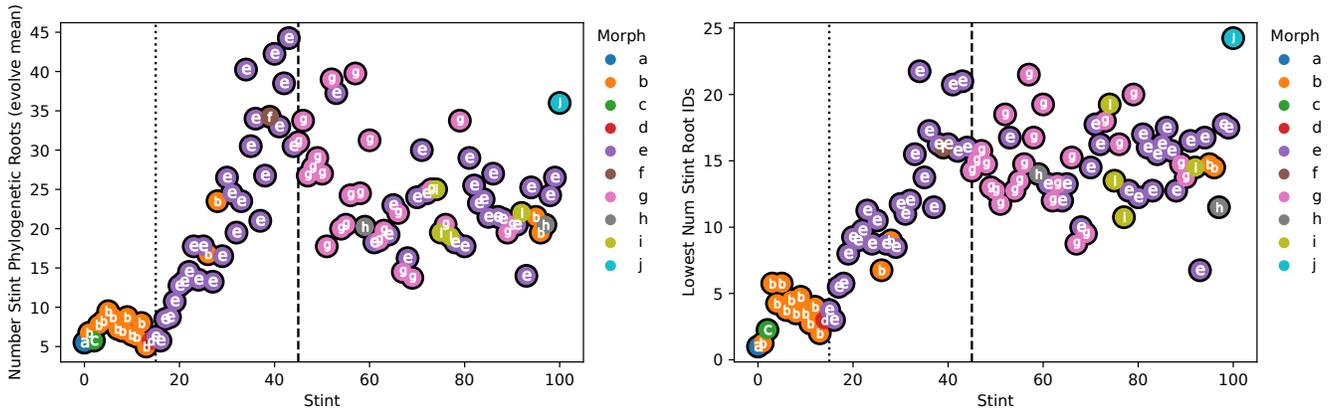
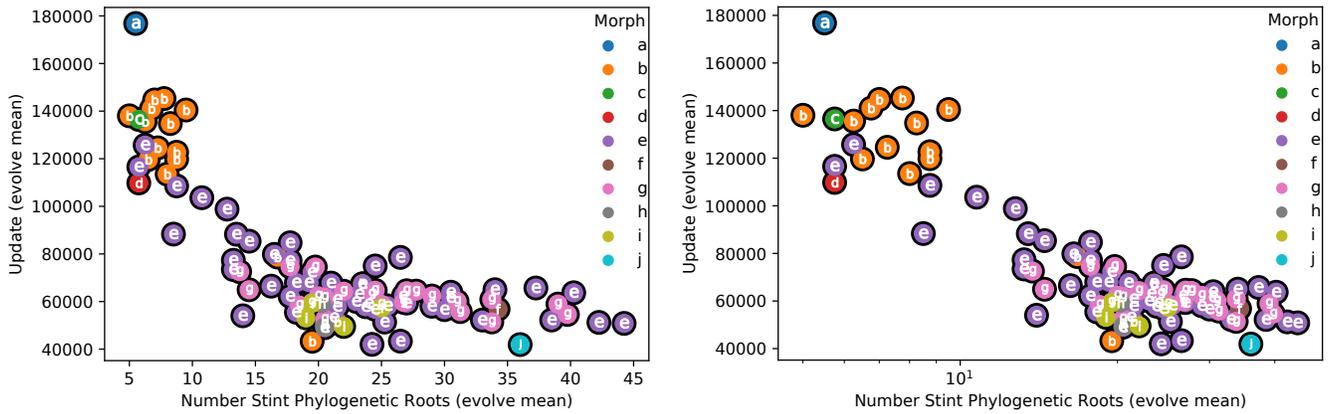


\input{fig/phylogeny/stint_roots.tex}
\input{fig/phylogeny/lowestroot_stint_roots.tex}
\input{fig/phylogeny/updates_vs_stint_roots.tex}
\input{fig/phylogeny/log_updates_vs_stint_roots.tex}

\caption{
Phylogenetic statistics.
Color coding and letters correspond to qualitative morph codes described in Table \ref{tab:morph_descriptions}.
Dotted vertical line denotes emergence of morph $e$.
Dashed vertical line denotes emergence of morph $g$.
}
\label{fig:phylogeny}
\end{figure*}

\begin{figure*}
\centering
\includegraphics[width=0.8\linewidth]{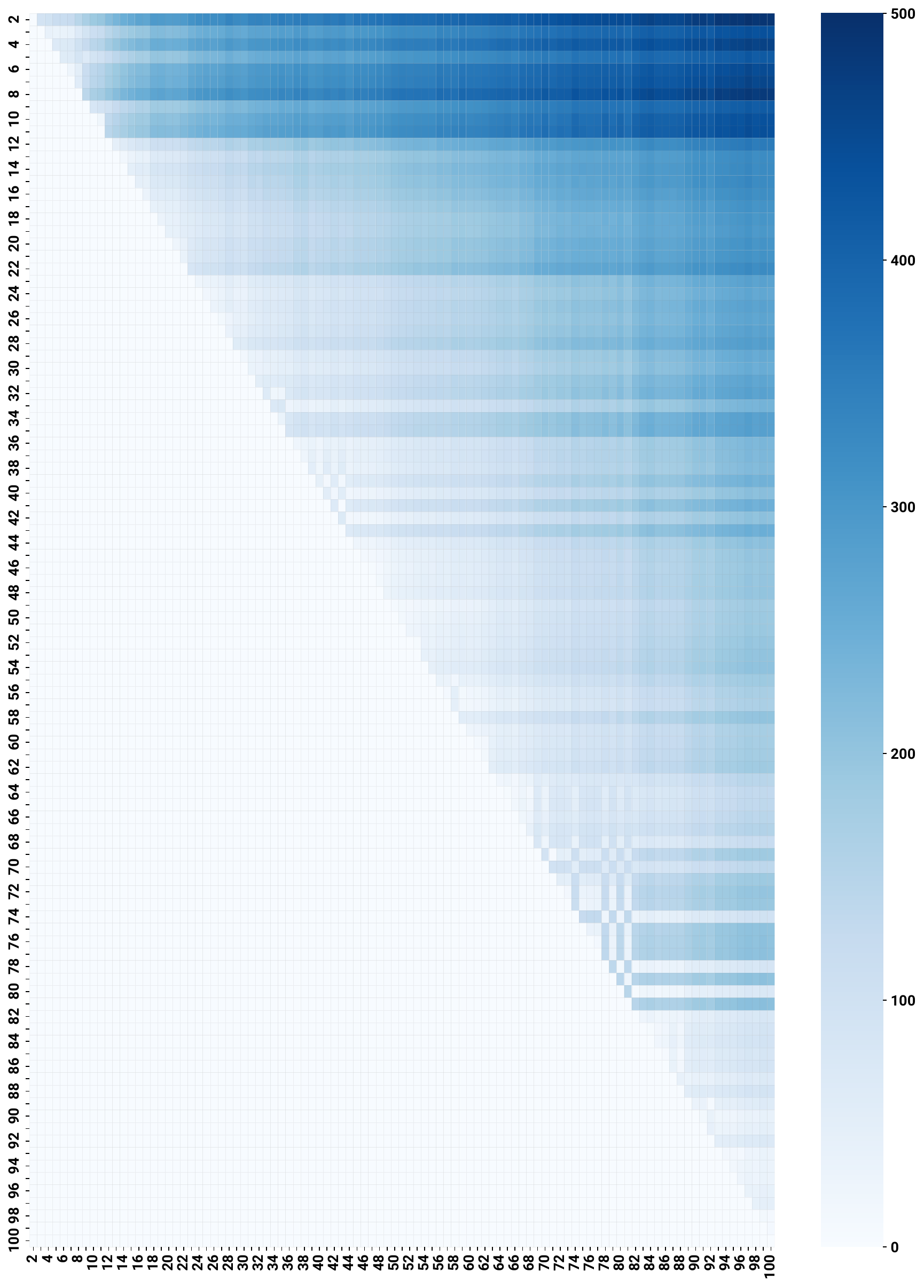}

\caption{
Point mutation distance between tag blocks from sampled focal strain representatives across stints.
These distances were used to reconstruct phylogenetic relationship of sampled focal strain representatives.
Light white indicates high similarity between tag blocks and dark blue indicates low similarity.
Representatives from stints 0 and 1, which share no common ancestry with representatives from other stints, are excluded.
Only upper triangular data plotted.
}
\label{fig:phylo_distance_matrix_heatmap}
\end{figure*}

\begin{figure*}

\input{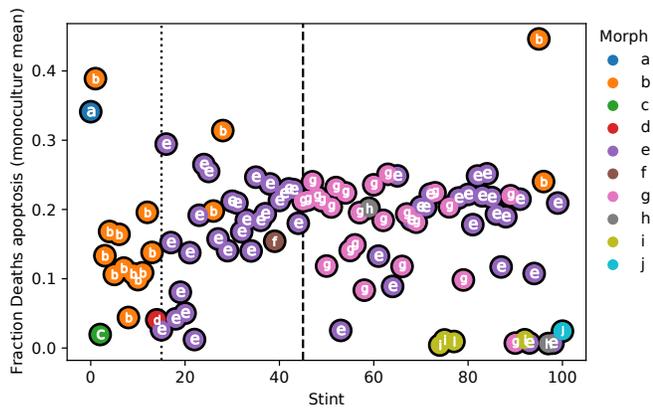}
\input{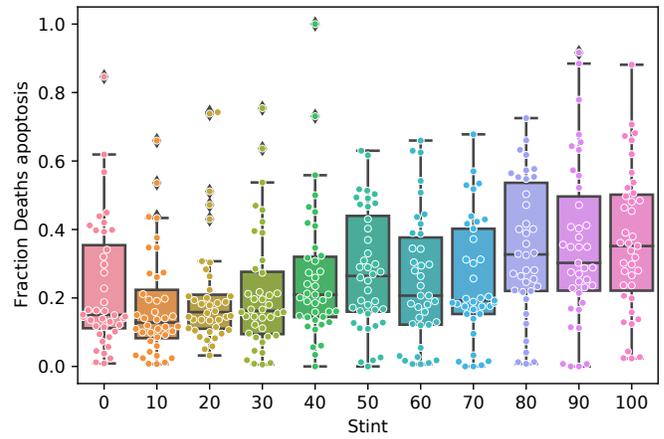}

\caption{ Apoptosis rates in case study strain and across evolutionary replicates. }
\label{fig:apoptosis}
\end{figure*}

\begin{figure*}
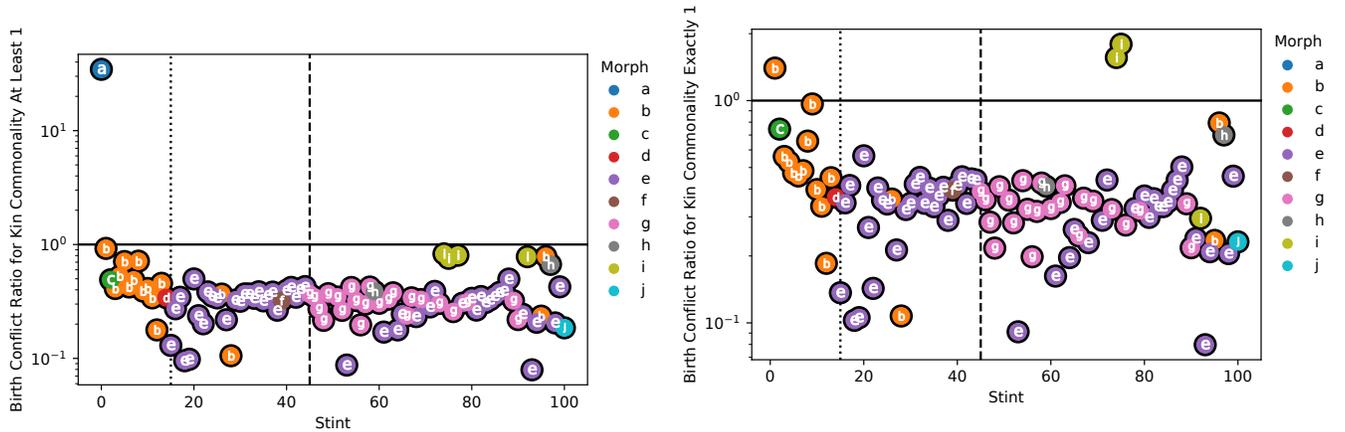
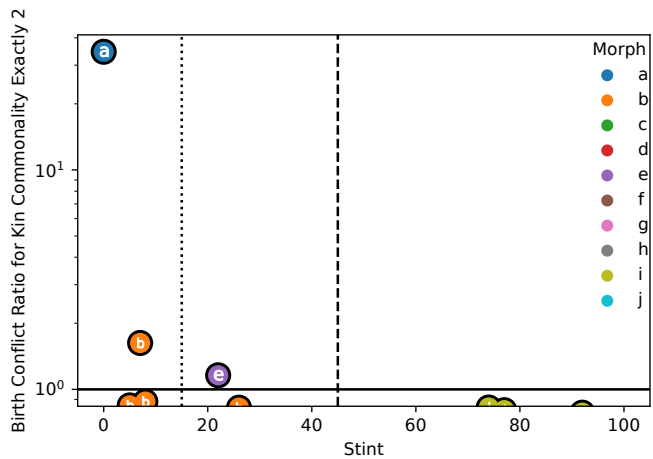


\input{fig/conflict/at_least_one.tex}
\input{fig/conflict/exactly_one.tex}
\input{fig/conflict/exactly_two.tex}

\caption{ Kin conflict rates.
Color coding and letters correspond to qualitative morph codes described in Table \ref{tab:morph_descriptions}.}
\label{fig:conflict}

\end{figure*}

\begin{figure*}
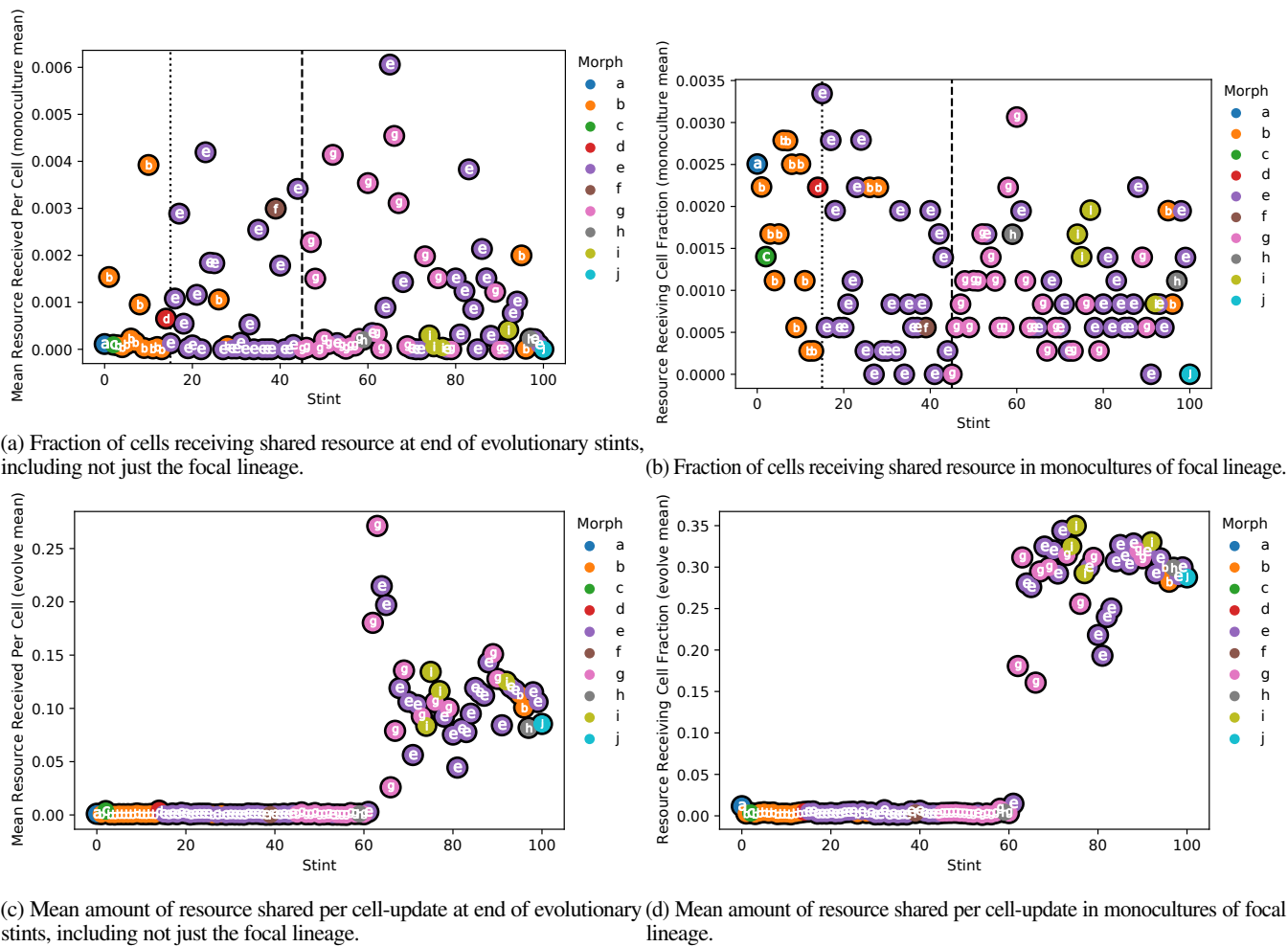


\input{fig/resource_sharing/fraction_sharing_evolve.tex}
\input{fig/resource_sharing/fraction_sharing_monoculture.tex}
\input{fig/resource_sharing/sharing_amount_evolve.tex}
\input{fig/resource_sharing/sharing_amount_monoculture.tex}

\caption{
Resource-sharing phenotypic traits.
Color coding and letters correspond to qualitative morph codes described in Table \ref{tab:morph_descriptions}.
Dotted vertical line denotes emergence of morph $e$.
Dashed vertical line denotes emergence of morph $g$.
}
\label{fig:resource_sharing}
\end{figure*}

\begin{figure*}
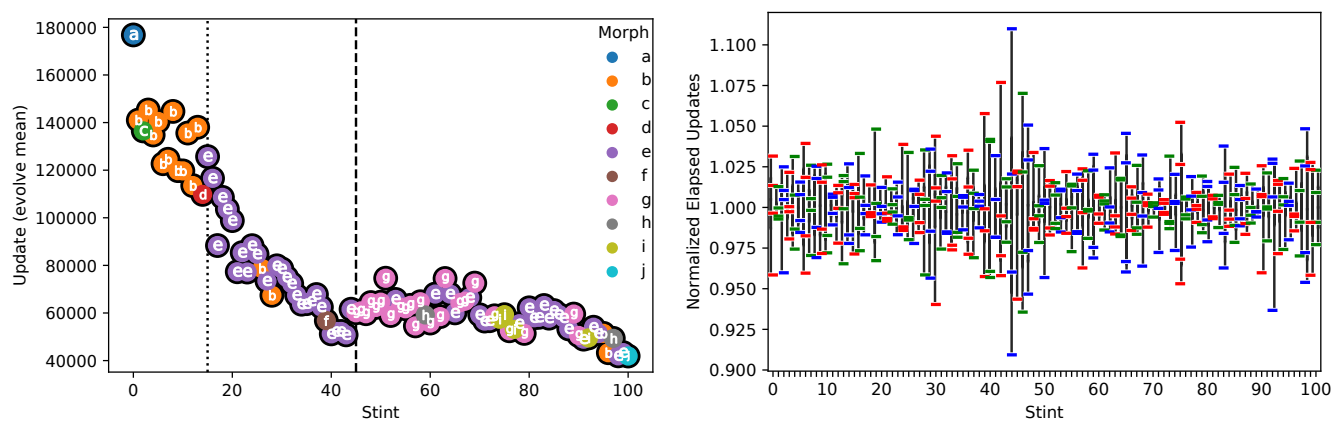


\input{fig/simulation/stint_updates.tex}
\input{fig/simulation/thread_updates.tex}

\caption{ Real-time simulation performance. }
\label{fig:simulation}
\end{figure*}

\begin{sidewaysfigure*}
\thisfloatpagestyle{mylandscape}%
\rotatesidewayslabel%
\centering
\includegraphics[width=\linewidth]{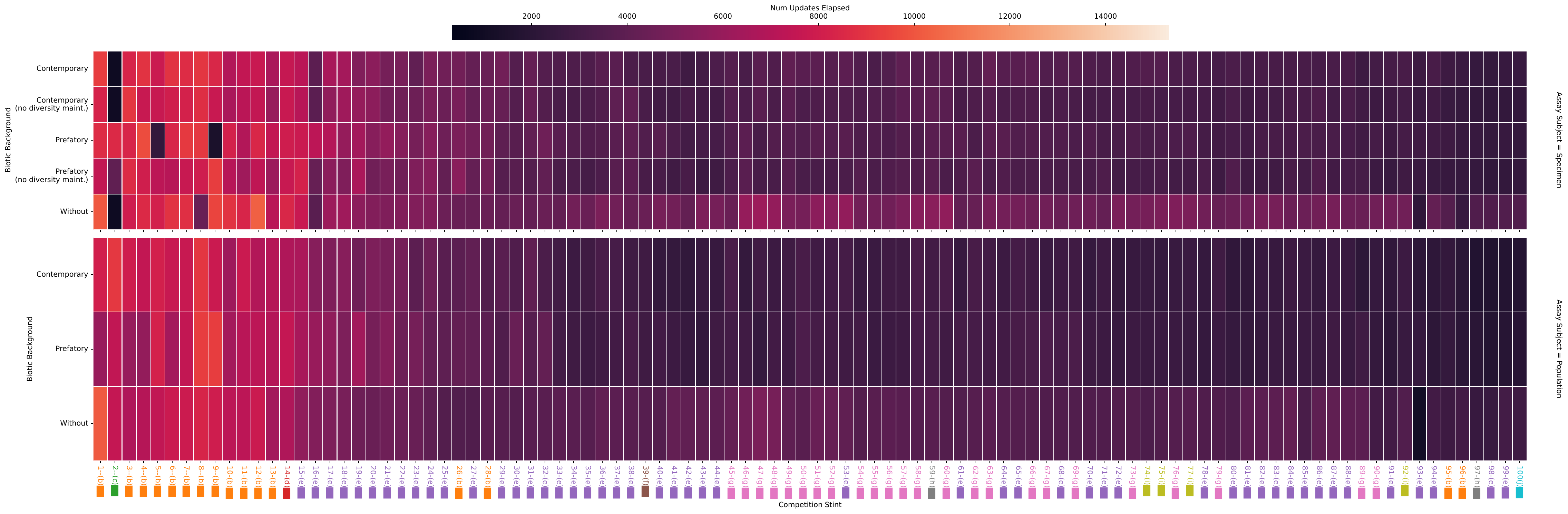}

\caption{
Number updates elapsed during fixed-duration adaptation assay competitions for sampled representative specimen (top) and population-level adaptation (bottom).
See Figure \ref{fig:adaptation_assay_cartoon} for explanation of competition biotic backgrounds.
See Supplementary Figure \ref{fig:num_updates_elapsed_barplot} for confidence interval estimates of mean updates elapsed during competition expedriments and Supplementary Figure \ref{fig:num_updates_elapsed_boxplot} for distributions of updates elapsed during competition experiments.
}
\label{fig:num_updates_elapsed_heatmap}
\end{sidewaysfigure*}

\begin{sidewaysfigure*}
\thisfloatpagestyle{mylandscape}%
\rotatesidewayslabel%
\centering
\includegraphics[width=\linewidth]{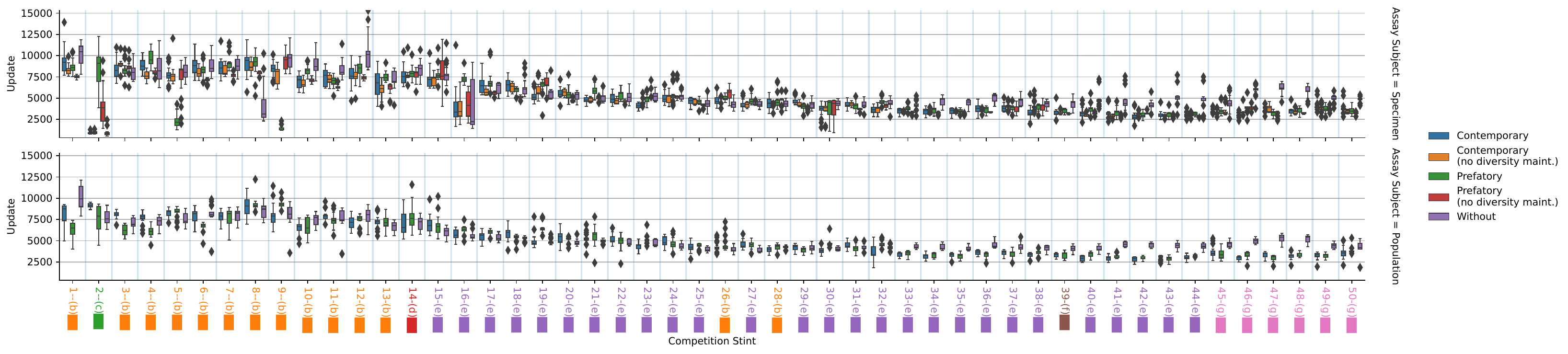}
\includegraphics[width=\linewidth]{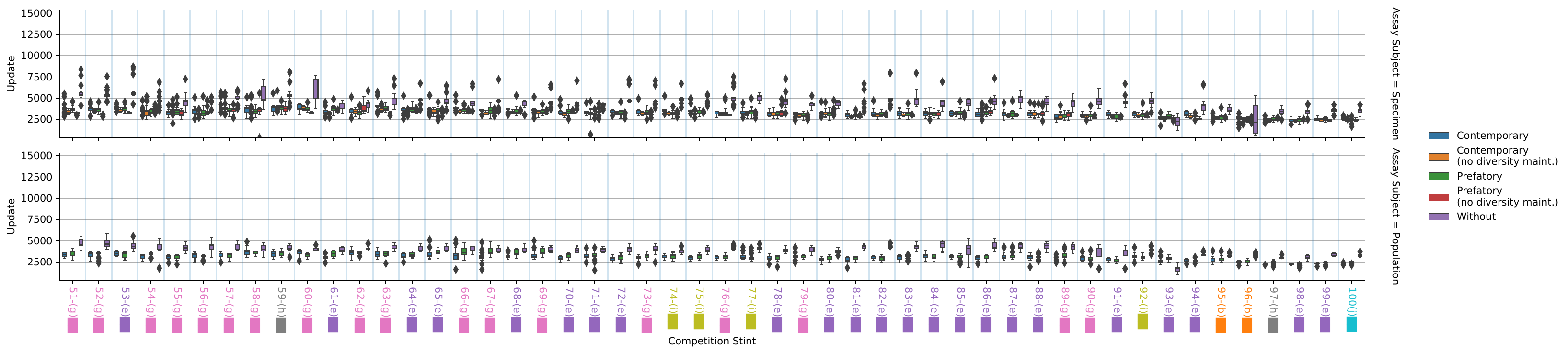}

\caption{
Number updates elapsed during fixed-duration adaptation assay competitions for sampled representative specimen (upper panels) population-level adaptation (lower panels).
Figure is split into two rows due to layout considerations.
See Figure \ref{fig:adaptation_assay_cartoon} for explanation of competition biotic backgrounds.
}
\label{fig:num_updates_elapsed_boxplot}
\end{sidewaysfigure*}

\begin{sidewaysfigure*}
\thisfloatpagestyle{mylandscape}%
\rotatesidewayslabel%
\centering
\includegraphics[width=\linewidth]{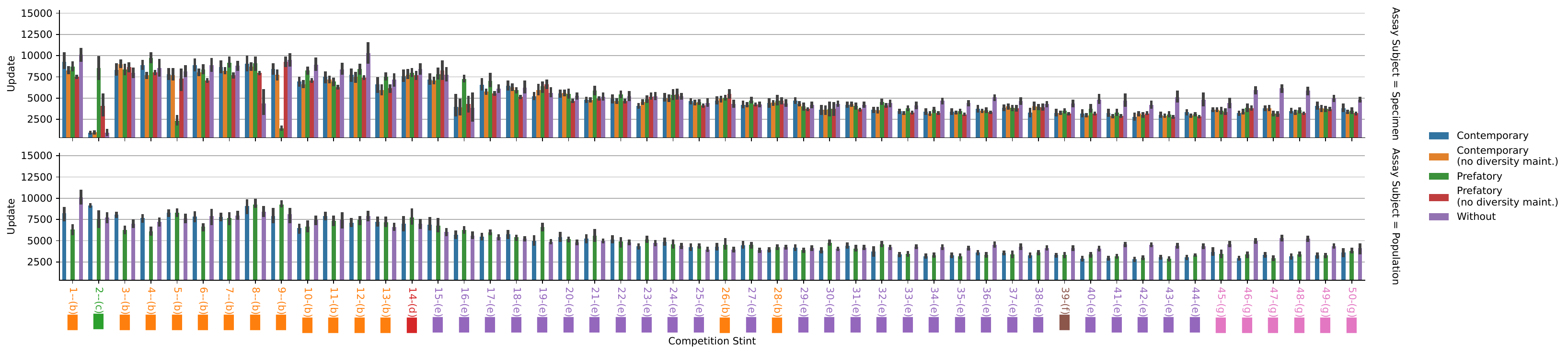}
\includegraphics[width=\linewidth]{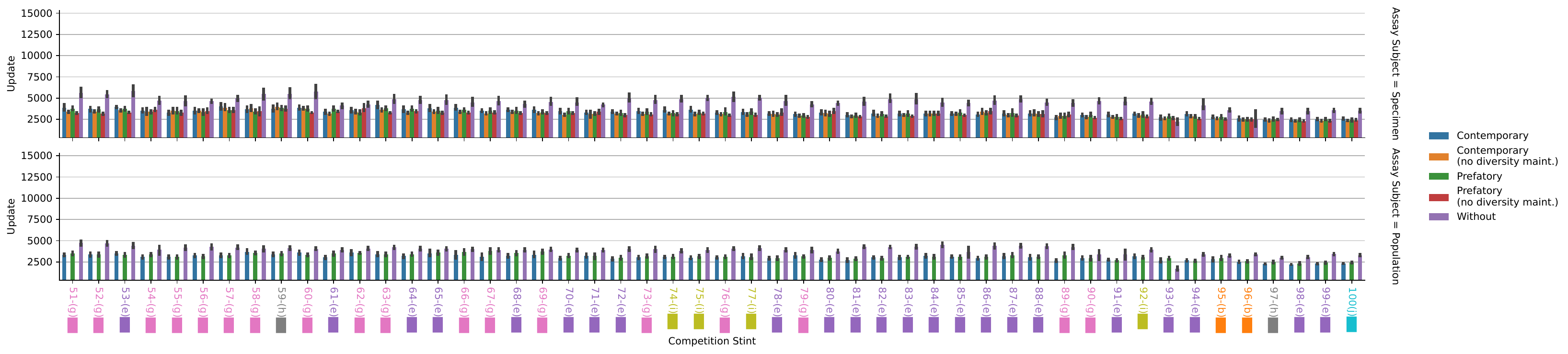}

\caption{
Number updates elapsed during fixed-duration adaptation assay competitions for sampled representative specimen (upper panels) population-level adaptation (lower panels).
Error bars are bootstrapped 95\% confidence intervals.
Figure is split into two rows due to layout considerations.
See Figure \ref{fig:adaptation_assay_cartoon} for explanation of competition biotic backgrounds.
}
\label{fig:num_updates_elapsed_barplot}
\end{sidewaysfigure*}

\begin{sidewaysfigure*}
\thisfloatpagestyle{mylandscape}%
\rotatesidewayslabel%
\centering
\includegraphics[width=\linewidth]{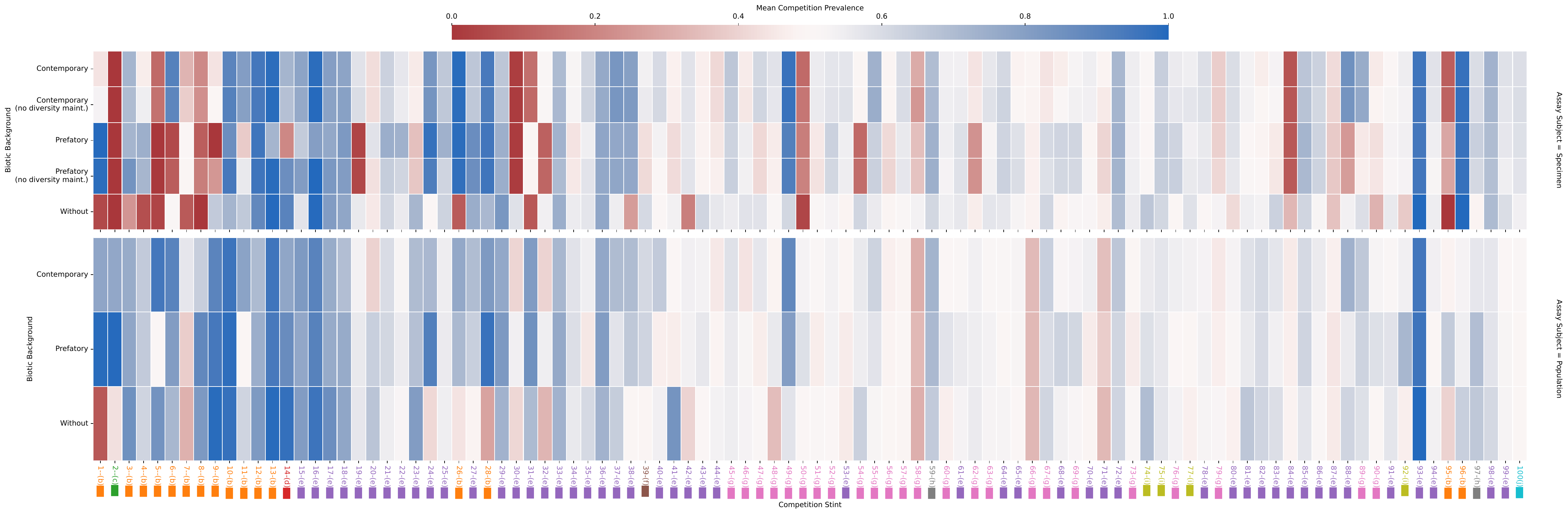}

\caption{
Mean end-state population composition of competition experiments.
Half (0.5) population composition corresponds to a neutral result, color mapped to white.
Blue indicates fitness gain compared to the previous stint and red indicates fitness loss.
Color coding and parentheticals of stint labels correspond to qualitative morph codes described in Table \ref{tab:morph_descriptions}.
Upper panel shows results for sampled focal strain genome, lower panel shows results for entire focal strain population.
See Figure \ref{fig:adaptation_assay_cartoon} for explanation of competition biotic backgrounds.
See Figure \ref{fig:mean_competition_prevalence_boxplot} for boxplot depiction of prevalence outcomes and Figure \ref{fig:mean_competition_prevalence_barplot} for bootstrapped confidence intervals on mean prevalence outcomes.
}
\label{fig:mean_competition_prevalence}
\end{sidewaysfigure*}

\begin{sidewaysfigure*}
\thisfloatpagestyle{mylandscape}%
\rotatesidewayslabel%
\centering
\includegraphics[width=\linewidth]{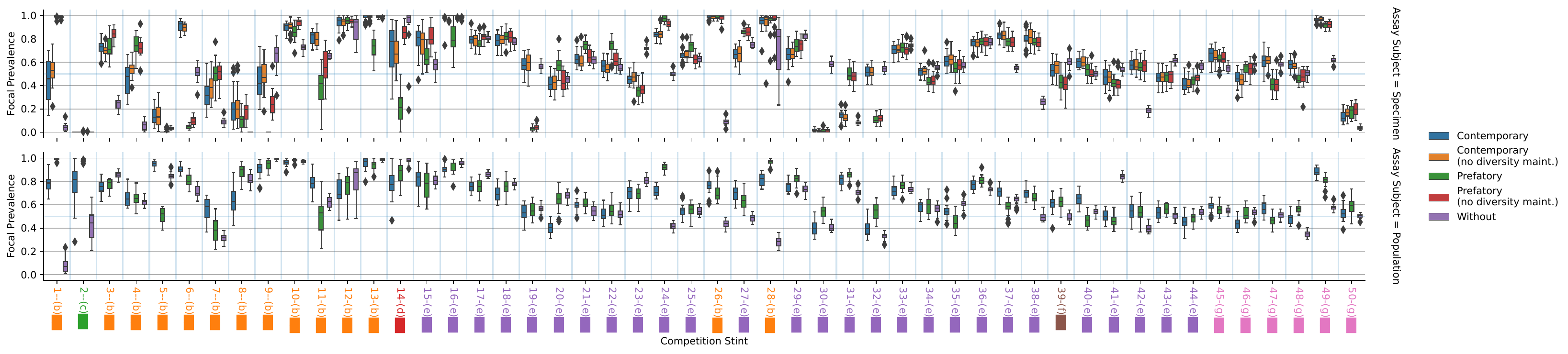}

\includegraphics[width=\linewidth]{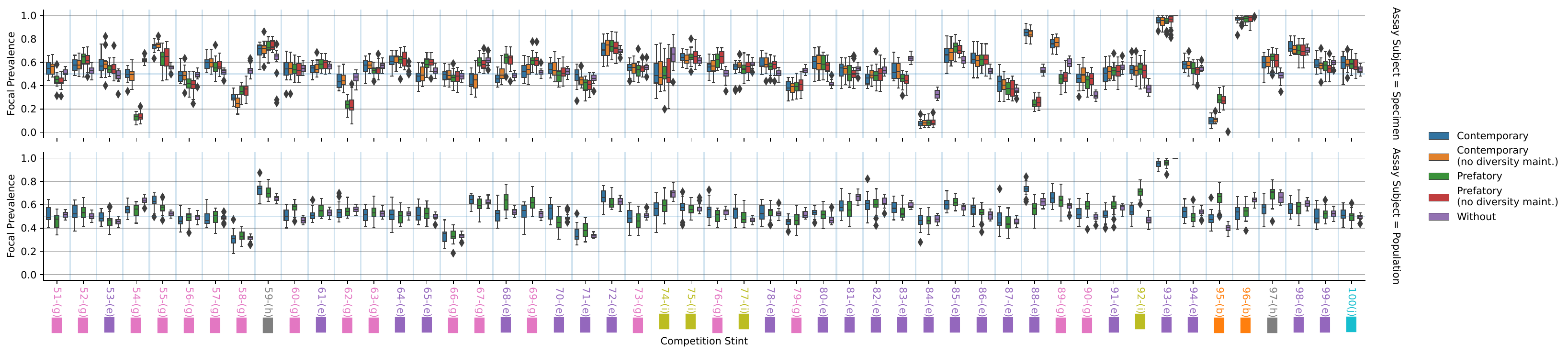}

\caption{
End-state population composition of competition experiments.
Half (0.5) population composition corresponds to a neutral result.
Zero population composition corresponds to extreme fitness loss compared to the previous stint.
Population composition of 1.0 corresponds to extreme fitness gain compared to the previous stint.
Color coding and parentheticals of stint labels correspond to qualitative morph codes described in Table \ref{tab:morph_descriptions}.
Upper panels show results for sampled focal strain genome, lower panels show results for entire focal strain population.
Figure is split into two rows due to layout considerations.
See Figure \ref{fig:adaptation_assay_cartoon} for explanation of competition biotic backgrounds.
See Figure \ref{fig:mean_competition_prevalence_barplot} for bootstrapped confidence intervals on mean prevalence outcomes.
}
\label{fig:mean_competition_prevalence_boxplot}
\end{sidewaysfigure*}

\begin{sidewaysfigure*}
\thisfloatpagestyle{mylandscape}%
\rotatesidewayslabel%
\centering

\includegraphics[width=\linewidth]{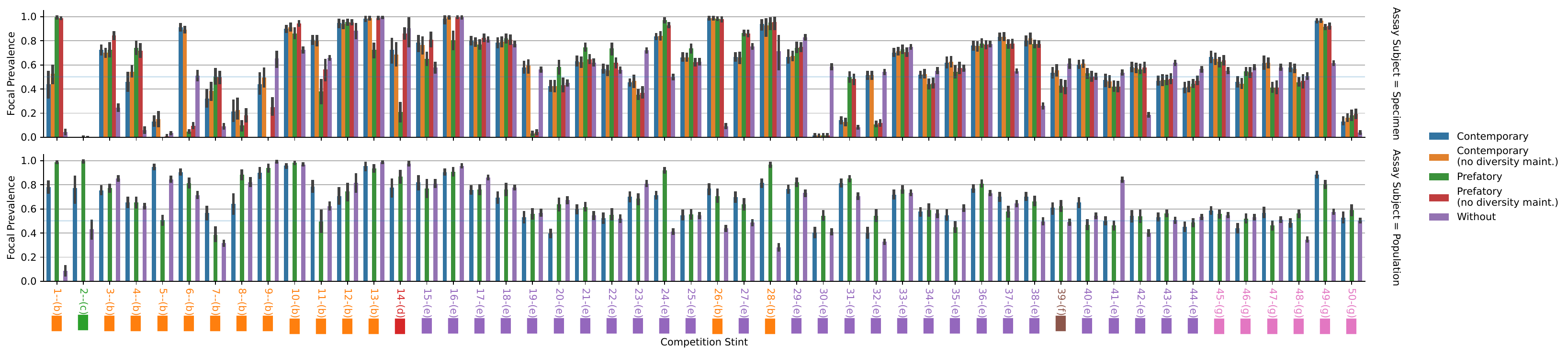}

\includegraphics[width=\linewidth]{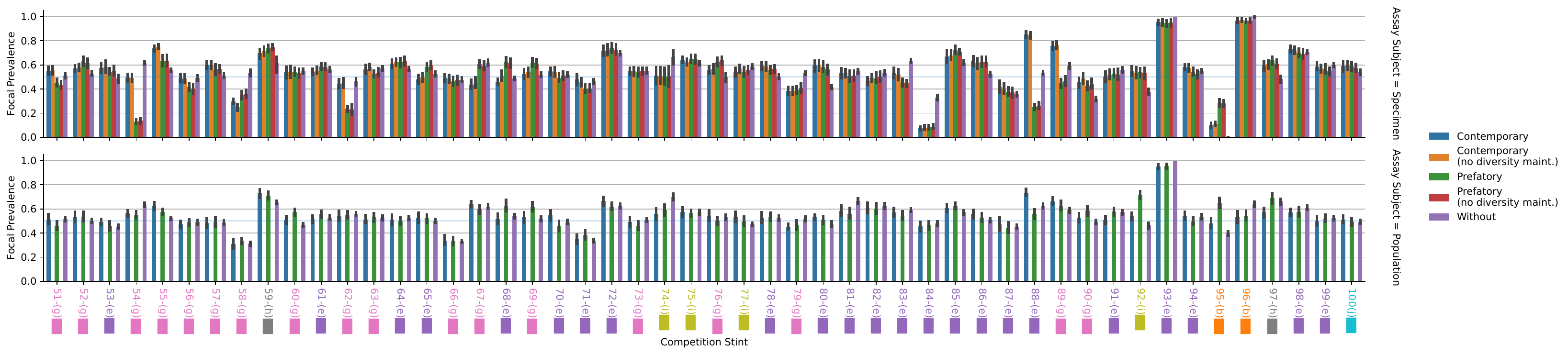}

\caption{
End-state population composition of competition experiments.
Half (0.5) population composition corresponds to a neutral result.
Zero population composition corresponds to extreme fitness loss compared to the previous stint.
Population composition of 1.0 corresponds to extreme fitness gain compared to the previous stint.
Error bars are bootstrapped 95\% confidence intervals.
Color coding and parentheticals of stint labels correspond to qualitative morph codes described in Table \ref{tab:morph_descriptions}.
Upper panels show results for sampled focal strain genome, lower panels show results for entire focal strain population.
Figure is split into two rows due to layout considerations.
See Figure \ref{fig:adaptation_assay_cartoon} for explanation of competition biotic backgrounds.
See Figure \ref{fig:mean_competition_prevalence_boxplot} for boxplot depiction of prevalence outcomes.
}
\label{fig:mean_competition_prevalence_barplot}
\end{sidewaysfigure*}

\begin{figure*}
\centering

\begin{subfigure}{\textwidth}

\includegraphics[width=\linewidth]{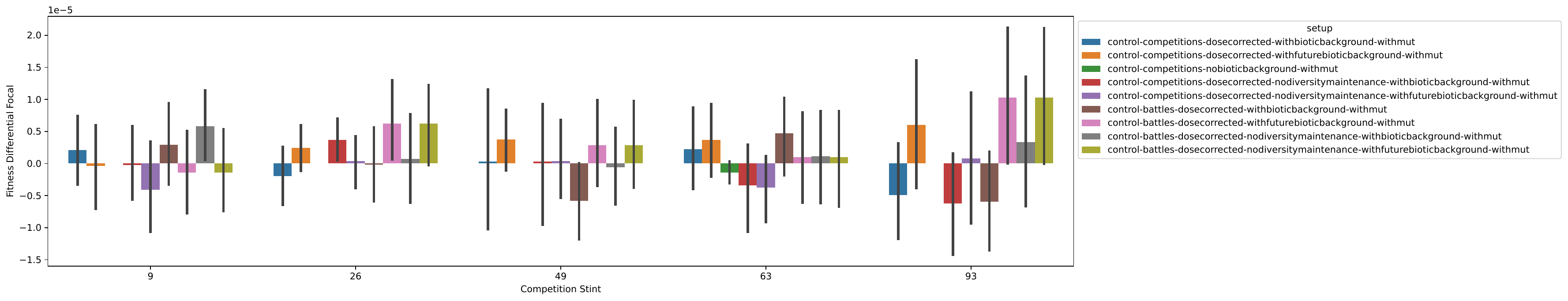}
\caption{
Calculated fitness differential between competing strains based on population composition at the end of competition experiments.
Zero is neutral.
Error bars are 95\% confidence intervals.
}
\end{subfigure}

\begin{subfigure}{\textwidth}
\includegraphics[width=\linewidth]{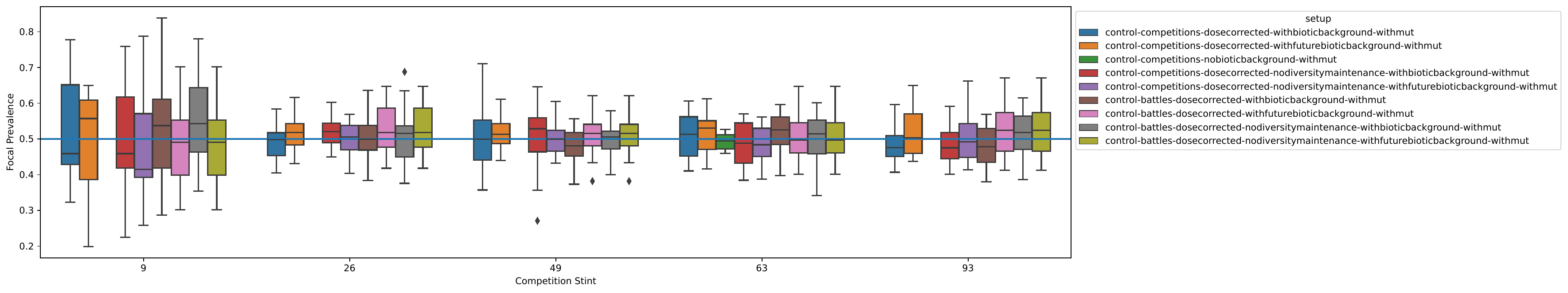}
\caption{
Fractional composition of focal population at the end of competition experiments.
A neutral outcome corresponds to even (0.5) composition, annotated with a horizontal line.
}
\end{subfigure}

\begin{subfigure}{\textwidth}
\includegraphics[width=\linewidth]{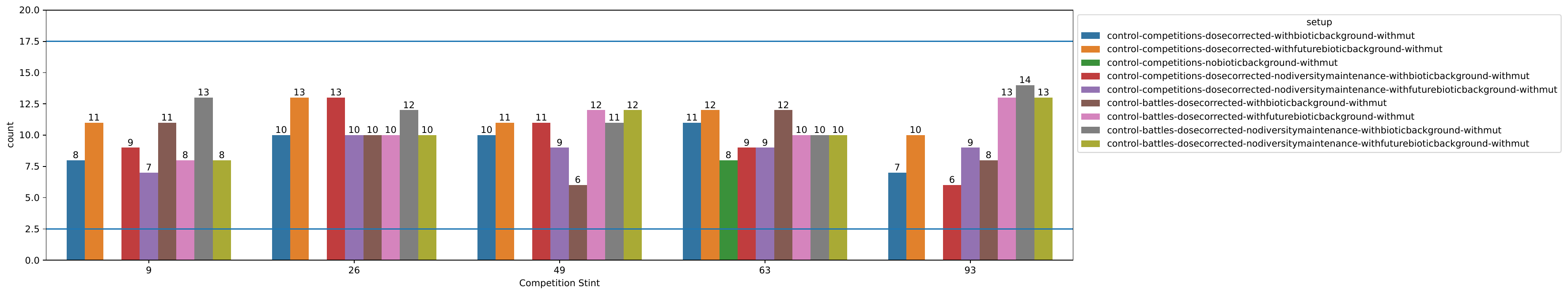}
\caption{
Number competitions out of 20 won by first strain.
Ten competitions won corresponds to a perfectly neutral outcome.
Eighteen and more or two or less competitions won were considered to indidicate a significant fitness difference between strains.
These thresholds for significance annotated with horizontal lines.
}
\end{subfigure}

\caption{
Control adaptation experiments for selected stints.
Control experiments were performed by competing two identical genomes or populations against each other with the contemporary biotic background, with the prefatory biotic background, or with no biotic background.
See Figure \ref{fig:adaptation_assay_cartoon} for summary of adaptation experiment design.
}
\label{fig:adaptation_control}
\end{figure*}

\begin{sidewaysfigure*}
    \thisfloatpagestyle{mylandscape}%
    \rotatesidewayslabel%
\centering
\includegraphics[width=\linewidth]{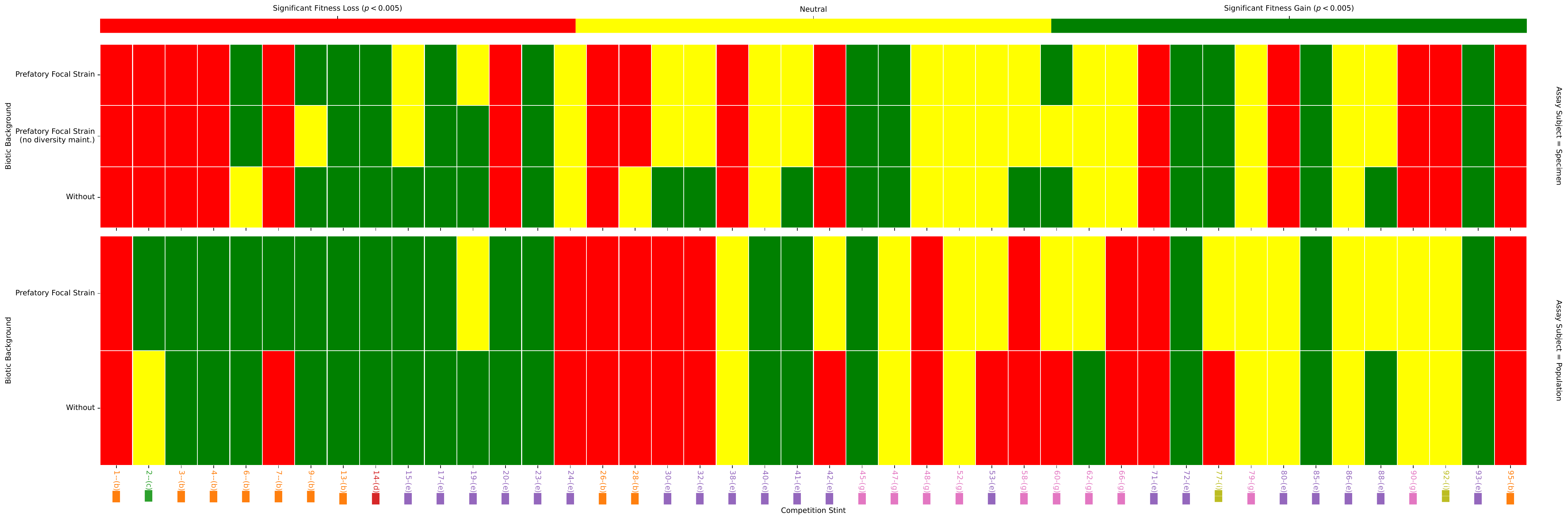}

\caption{
\textbf{Adaptation assay outcomes.}
\footnotesize
Summary of biotic background control adaptation assay outcomes for sampled representative specimen (top) and population-level adaptation (bottom).
Color coding and parentheticals of stint labels correspond to qualitative morph codes described in Table \ref{tab:morph_descriptions}.
See Figure \ref{fig:adaptation_assay_cartoon} for explanation of competition biotic backgrounds.
}
\label{fig:baseline_fitness_gain_or_loss}
\end{sidewaysfigure*}

\begin{figure*}
\centering

\begin{subfigure}{\textwidth}
\centering
\includegraphics[width=0.7\linewidth]{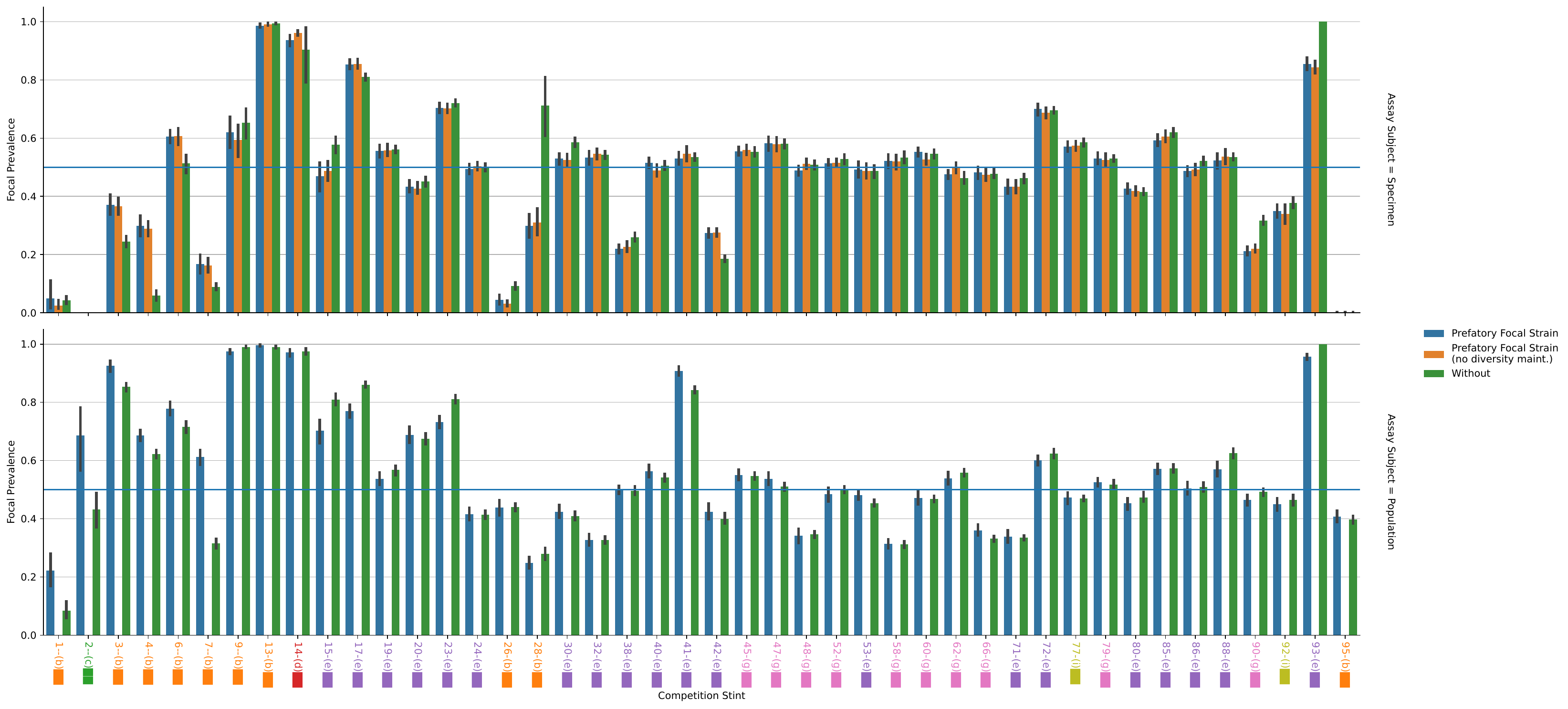}
\caption{
Fractional composition of focal population at the end of competition experiments.
Zero is neutral.
Error bars are 95\% confidence intervals.
}
\end{subfigure}

\begin{subfigure}{\textwidth}
\centering
\includegraphics[width=0.7\linewidth]{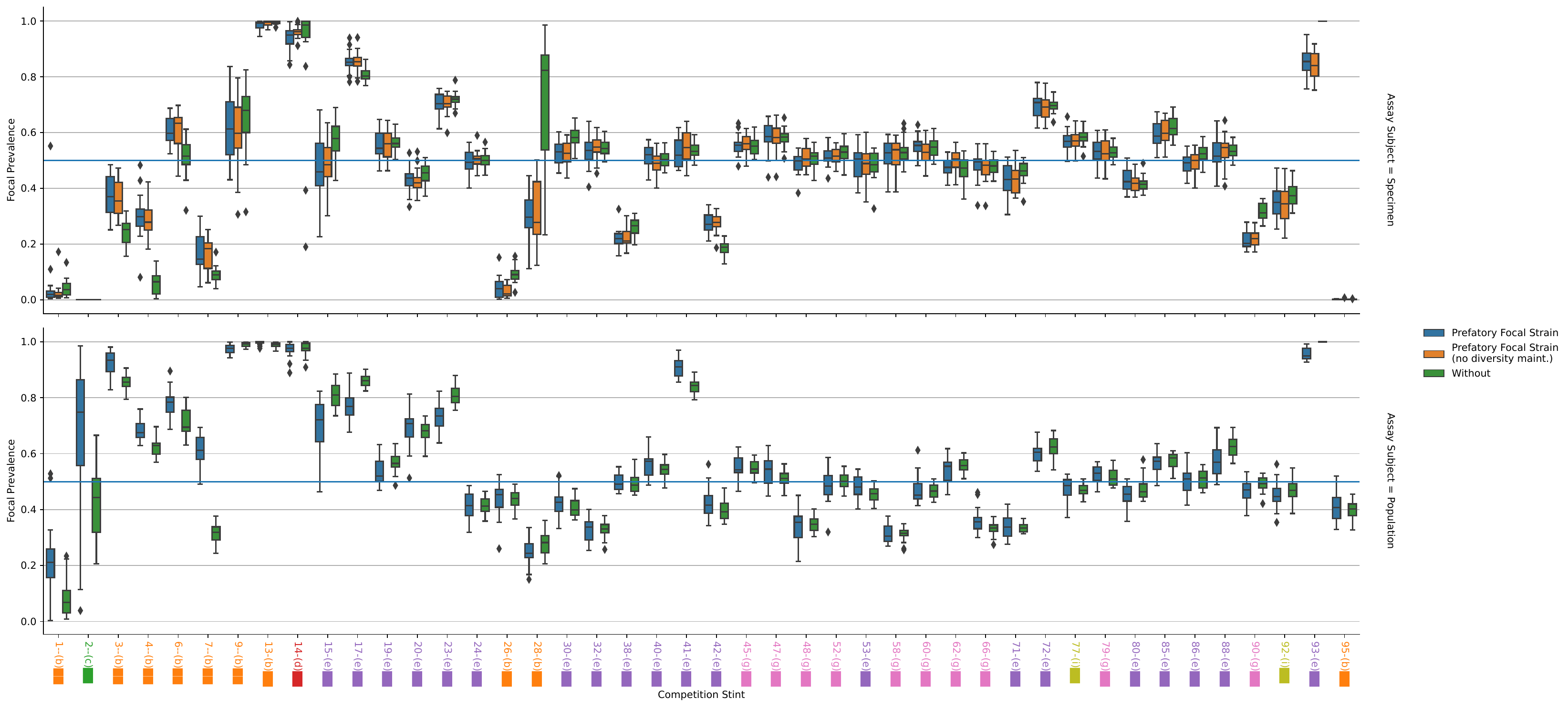}
\caption{
Fractional composition of focal population at the end of competition experiments.
A neutral outcome corresponds to even (0.5) composition, annotated with a horizontal line.
}
\end{subfigure}

\begin{subfigure}{\textwidth}
\centering
\includegraphics[width=0.7\linewidth]{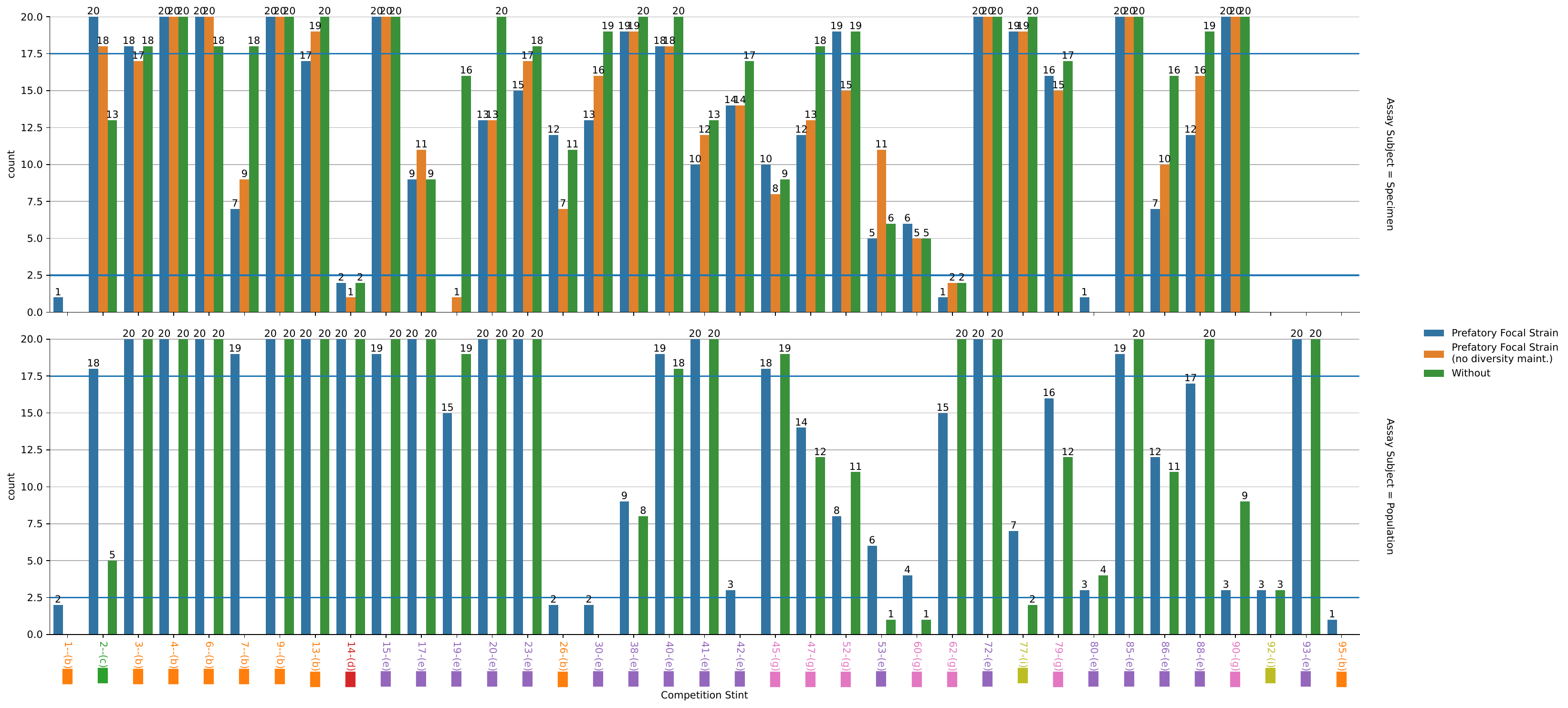}
\caption{
Number competitions out of 20 won by first strain.
Ten competitions won corresponds to a perfectly neutral outcome.
Eighteen and more or two or less competitions won were considered to indicate a significant fitness difference between strains.
These thresholds for significance annotated with horizontal lines.
}
\end{subfigure}

\caption{
Biotic background control adaptation experiments for selected stints.
Biotic background control experiments were performed by substituting the baseline competitor for the biotic background.
See Figure \ref{fig:adaptation_assay_cartoon} for summary of adaptation experiment design.
}
\label{fig:baseline_adaptation_control}
\end{figure*}

\end{document}